%% file: main.tex
\documentclass[10pt,twocolumn,letterpaper]{article}

\usepackage{adjustbox}
\usepackage{nicefrac}
\usepackage{amsmath}
\usepackage{framed}
\usepackage{multirow}
\usepackage{slashbox}
\usepackage{booktabs}
\usepackage{amsfonts}
\usepackage{microtype}
\usepackage{nicefrac}       %
\usepackage{microtype}      %
\usepackage[dvipsnames]{xcolor}         %
\usepackage[accsupp]{axessibility}
\usepackage{colortbl}
\usepackage{ulem}
\usepackage{balance}
\usepackage[strict]{changepage}
\definecolor{formalbar}{rgb}{0.290,0.325,0.337}
\definecolor{formalshade}{rgb}{0.925,0.941,0.976}

\newenvironment{formal}{%
  \MakeFramed{\advance\hsize-\width\FrameRestore}%
  \noindent\hspace{-4.55pt}%
  \begin{adjustwidth}{}{7pt}%
  \vspace{2pt}\vspace{2pt}%
}
{%
  \vspace{-3pt}\end{adjustwidth}\endMakeFramed%
}

\usepackage[pagenumbers]{cvpr} 


\definecolor{cvprblue}{rgb}{0.21,0.49,0.74}
\usepackage[pagebackref,breaklinks,colorlinks,allcolors=cvprblue]{hyperref}
\def\paperID{1787} 
\def\confName{CVPR}
\def\confYear{2025}
\title{ShadowHack: Hacking Shadows via Luminance-Color Divide and Conquer}

\author{Jin Hu$^{\ast}$  \quad  Mingjia Li\thanks{Equal Contribution.}  \quad  Xiaojie Guo\thanks{Corresponding author. This work was supported by National Natural Science Foundation of China under Grant nos. 62372251 and 62072327.}\\
College of Intelligence and Computing, Tianjin University\\
{\tt\small \{jinhu, mingjiali\}@tju.edu.cn, xj.max.guo@gmail.com}
}

\begin{document}
\maketitle
\input{sec/0_abstract}    
\input{sec/1_intro}
\input{sec/2_related}

\input{sec/3_method}
\input{sec/4_exp}
\input{sec/5_conclusion}

{\small
\bibliographystyle{ieeenat_fullname}
    \bibliography{main}
}



\end{document}

%% file: sec/0_abstract.tex
\begin{abstract}

Shadows introduce challenges such as reduced brightness, texture deterioration, and color distortion in images, complicating a holistic solution. This study presents \textbf{ShadowHack}, a divide-and-conquer strategy that tackles these complexities by decomposing the original task into luminance recovery and color remedy. To brighten shadow regions and repair the corrupted textures in the luminance space, we customize LRNet, a U-shaped network with a rectified outreach attention module, to enhance information interaction and recalibrate contaminated attention maps. With luminance recovered, CRNet then leverages cross-attention mechanisms to revive vibrant colors, producing visually compelling results. Extensive experiments on multiple datasets are conducted to demonstrate the superiority of ShadowHack over existing state-of-the-art solutions both quantitatively and qualitatively, highlighting the effectiveness of our design. Our code will be made publicly available at \href{https://github.com/lime-j/ShadowHack}{here}.

\end{abstract}

%% file: sec/1_intro.tex
\begin{formal}
\textit{Shadow is a color as light is, but less brilliant; light and shadow are only the relation of two tones.}

\hfill ~~~— Paul Cézanne
\end{formal}
\section{Introduction}
\label{sec:intro}

Shadows are the absence of light, an everyday phenomenon that often shapes the aesthetic beauty of nature. However, shadows obscure the actual shape of instances, diminishing their visibility. Very likely, they deteriorate the performance of various machine vision tasks, such as object detection~\cite{NadimiB04, HsiehHCC03}, face recognition~\cite{ZhangZMC19}, and semantic segmentation~\cite{ZhouBWN20}, to name just a few. Consequently, shadow removal has become a critical research topic. 
\begin{figure}[t]
    \centering

    \rotatebox{90}{\small{Luminance}}
    \includegraphics[width=0.23\linewidth]{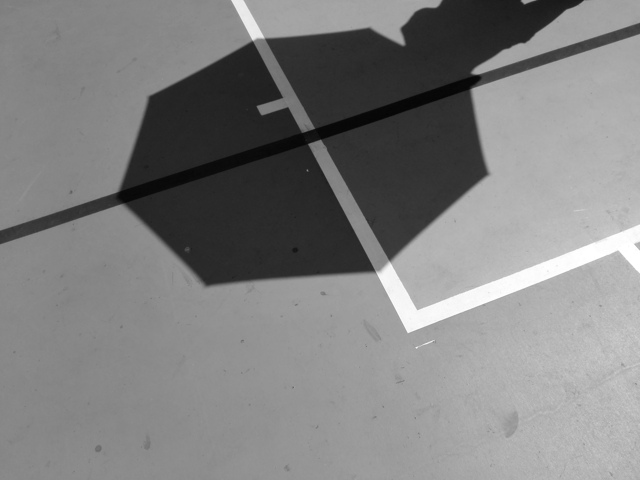}
    \includegraphics[width=0.23\linewidth]{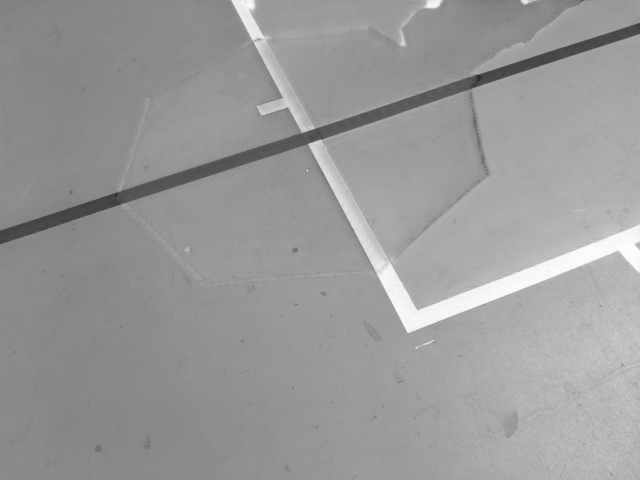}
    \includegraphics[width=0.23\linewidth]{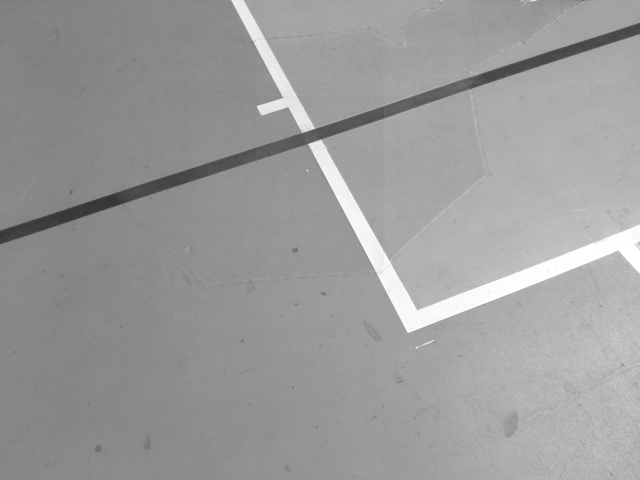}
    \includegraphics[width=0.23\linewidth]{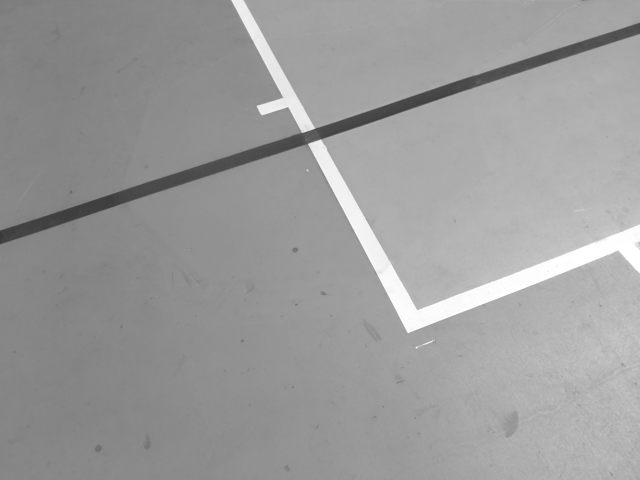}
    
    \rotatebox{90}{~~~~~\small{RGB}}
    \includegraphics[width=0.23\linewidth]{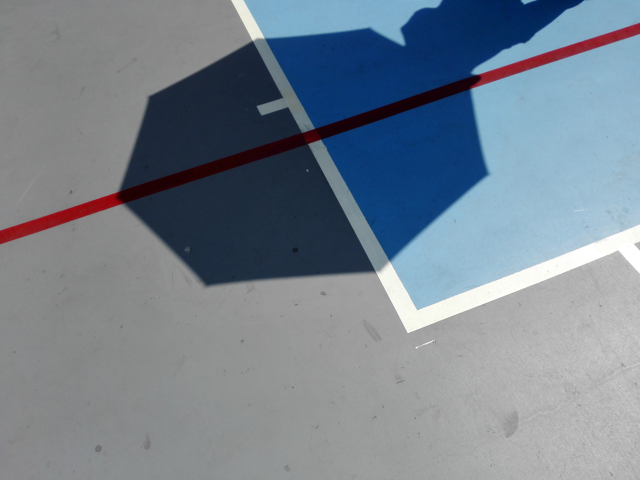}
    \includegraphics[width=0.23\linewidth]{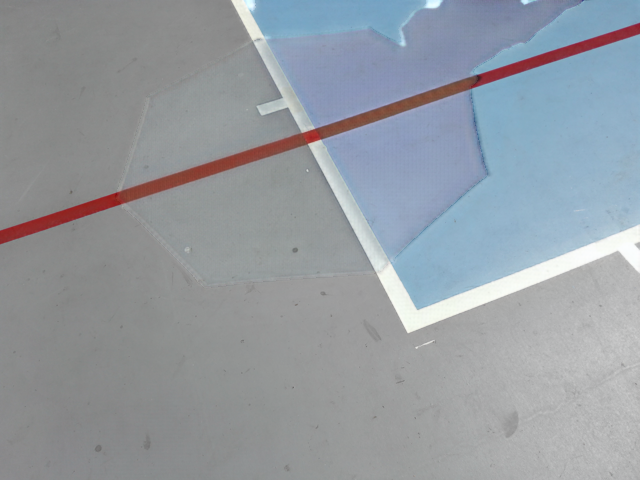}
    \includegraphics[width=0.23\linewidth]{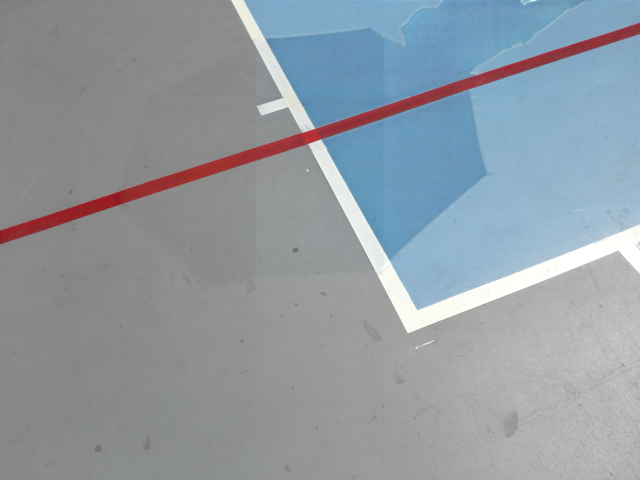}
    \includegraphics[width=0.23\linewidth]{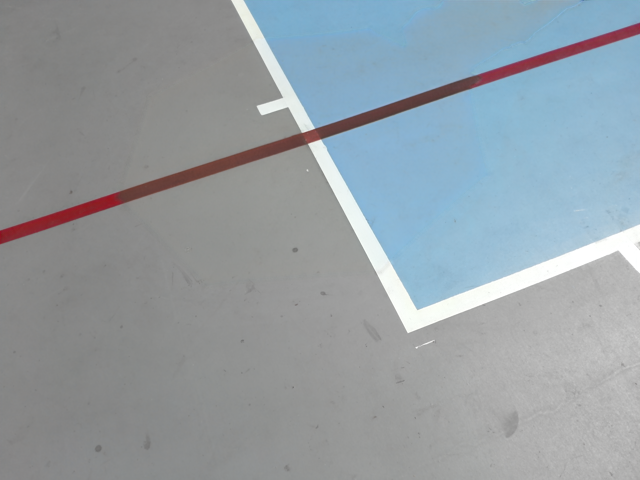}
    
    \rotatebox{90}{}
    \begin{subfigure}{0.23\linewidth}
        \subcaption{Input}
    \end{subfigure}
    \begin{subfigure}{0.23\linewidth}
        \subcaption{SG~\cite{WanYWWLW22}}
    \end{subfigure}
    \begin{subfigure}{0.23\linewidth}
        \subcaption{HF~\cite{0002FZL00Z24}}
    \end{subfigure}
    \begin{subfigure}{0.23\linewidth}
        \subcaption{Ours}
    \end{subfigure}
    \caption{Visual comparison between our method and state-of-the-art methods in both Luminance and RGB spaces.}
    \label{fig:intro1}
\vspace{-10pt}\end{figure}

Recent advancements in deep learning have greatly improved the efficiency and accuracy of shadow removal. Specifically, DeshadowNet~\cite{QuTHTL17} marked a pioneering step with its end-to-end Convolutional Neural Network (CNN) approach, while subsequent GAN-based methods~\cite{WangL018, HuJFH19, ZhangLZX20, LiuYWWM021} further advanced the field by adopting adversarial learning frameworks, enabling effective unpaired training. Additionally, Wan \textit{et al.}~\cite{WanYWWLW22} introduced style transfer, and Li \textit{et al.}~\cite{Li0A00T023} incorporated pre-trained inpainting priors to better restore textural information. Transformer-based architectures, such as ShadowFormer~\cite{GuoHLCW23}, HomoFormer~\cite{0002FZL00Z24}, and RASM~\cite{liu2024regional}, have since been developed to restore shadow images more effectively, while diffusion-based approaches~\cite{GuoWYHWPW23, Jin0YYT24} have also demonstrated promising results, particularly in complex scenarios where previous generative models often struggled.

\begin{figure}[t]
    \centering
     \includegraphics[width=\linewidth]{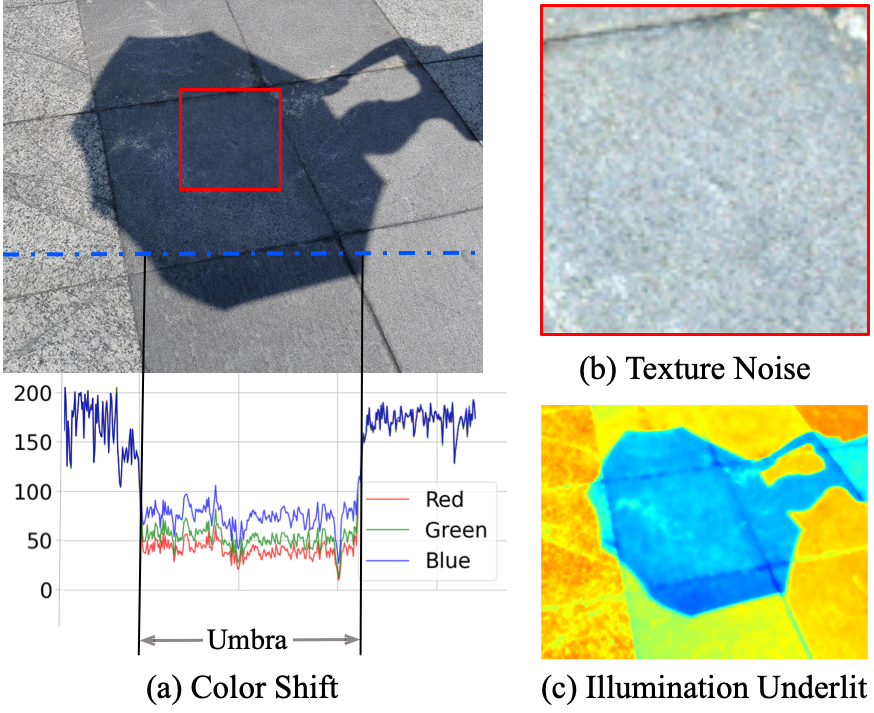}
      \caption{Example of the complex degradation in a shadow image.}
      \vspace{-5pt}
    \label{fig:deg}
\vspace{-10pt}\end{figure}

Despite these advancements, the odyssey of removing shadows is still far from its end. As evidenced in Fig.~\ref{fig:intro1}, even state-of-the-art models fail to recover color and underlying texture accurately. 
The difficulty stems from the complex and intertwined nature of shadow degradation, which involves not only \textit{reduced brightness} in shadow (umbra) regions~\cite{Lynch15}, but also \textit{texture deterioration} from the imaging process~\cite{Li0A00T023, LiuKXLWL24}, and \textit{color distortion} caused by surface material properties and environmental color influences~\cite{churma1994blue, LeS22, WanYWWLW22}, an example of which is in Fig.~\ref{fig:deg}. This complexity naturally suggests breaking down the original task, with its entangled interferences, into isolated and simpler sub-problems, each targeting a specific aspect and addressed independently.

Following the above line of thought, some works~\cite{GuoDH13, LeS19, LeS22} attempted to estimate light and color biases as scalars using linear models. Insights can also be drawn from the low-light enhancement literature, where the Retinex decomposition~\cite{land1977retinex, barrow1978recovering} is commonly employed~\cite{GuoLL17, GuoH23, DBLP:conf/iccv/CaiBLWTZ23, DBLP:journals/ijcv/XuZYM24}. Under the Retinex theory, an image can be decomposed into a structured illumination, and a colored texture-rich counterpart. However, these approaches encounter challenges due to \textit{the inherent inhomogeneity of shadows}. In other words, shadow removal extends beyond low-light enhancement and brings additional color biases in the decomposed reflectance map. This realization underscores the need for more sophisticated solutions to effectively disentangle the complex degradations introduced by shadows.  

This work invents ShadowHack, an effective shadow removal framework that individually tackles illumination and color components. ShadowHack starts with a simple yet highly effective hack for shadow decomposition. Unlike conventional methods that decouple illumination from color and texture, we alternatively separate color from illumination and texture. Once the input images are decomposed into luminance and color components, two dedicated networks are established, {\it i.e.}, the Luminance Restoration Network (LRNet) and the Color Regeneration Network (CRNet). LRNet, a lightweight transformer-based restoration model, focuses on recovering illumination and texture. To enable an effective reference for shadow region refinement, we introduce the Rectified Outreach Attention (ROA) Module, which uses dilated outreached window partitions and differential attention rectification to support high-quality luminance restoration.
Afterwards, the processed luminance serves as a guide for querying accurate color information in shadow regions from the non-shadow areas. Inspired by exemplar-based colorization methods, we inject color information via an ImageNet-pre-trained color encoder, which integrates color data into the intermediate features of the skip connections of our CRNet. This is facilitated by cross-attention modules that align the color in the shadow region with patches of similar luminance. Through this sequential processing, ShadowHack achieves effective shadow removal, supported by extensive experimental results. 
The primary contributions of this study are summarized as follows:
\begin{itemize}
    \item Based on an in-depth analysis of shadow properties and the limitations in current end-to-end decomposition methods, we introduce ShadowHack, a novel shadow removal framework that breaks shadow removal into two distinct subtasks: luminance recovery and color regeneration.
    \item We propose rectified outreach attention, an innovative mechanism that expands the receptive field in window-based attention modules and re-calibrates noise with marginal computational overhead, enabling precise restoration of contaminated texture and irregular illumination.
    \item We design two specialized networks, LRNet and CRNet, tailored for luminance refinement and color regeneration. Together, these networks allow our model to achieve accurate and robust results, setting new state-of-the-art performance across benchmarking datasets.

\end{itemize}

%% file: sec/2_related.tex
\section{Related Work}
\label{sec:formatting}

\textbf{Shadow Removal.}
Traditional methods, such as~\cite{YangTA12} and~\cite{ShorL08}, leveraged principles of physics and mathematical statistics on factors like illumination~\cite{ZhangZX15, FinlaysonHD02}, color~\cite{WuT05, KhanBST16}, region~\cite{GuoDH13}, and gradient~\cite{FinlaysonDL09, FinlaysonHLD06}. However, these methods often struggle with complex real-world scenes, where shadows interact in nuanced ways with surrounding colors and textures.
Deep learning has recently made significant strides in the field~\cite{QuTHTL17, WangL018, LeS19}.  For instance,
Qu {\it et al.}~\cite{QuTHTL17} proposed to extracts both global and semantic features from images, marking an early attempt to integrate deep learning into shadow removal. Later, Wang {\it et al.}~\cite{WangL018} proposed a multi-task framework that performs shadow detection and removal concurrently, enhancing overall accuracy. Generative Adversarial Networks (GANs) have also been applied to this task. As a representative, RIS-GAN~\cite{ZhangLZX20} incorporates residual and illumination features to refine shadow regions, while other works~\cite{HuJFH19, LiuYWWM021} use unsupervised GANs to generate shadows for achieving cycle consistency. Li {\it et al.}~\cite{Li0A00T023} adopted the concept of inpainting, simultaneously using a shadow image encoder and a generative-pretrained image encoder to extract features. Zhu {\it et al.}~\cite{ZhuHFZSZ22} employed normalizing flow for a reversible link between shadow generation and removal. Additionally, Le {\it et al.}~\cite{LeS22} proposed a linear model that adjusts the global lightness of shadow regions through physical modeling. 
More recently, diffusion models~\cite{Sohl-DicksteinW15} have shown even greater generative capabilities than GANs. Guo {\it et al.}~\cite{GuoWYHWPW23} pioneered the use of diffusion models for shadow removal, refining shadow masks throughout the process. Liu {\it et al.}~\cite{LiuKXLWL24} applied Retinex decomposition~\cite{land1977retinex, GuoLL17} and used diffusion models to refine textures under the guidance by the obtained illumination. However, a primary drawback of these diffusion models is their intensive computational cost.

\noindent\textbf{ViT-based Restoration.}
Vision Transformers (ViTs) have demonstrated impressive performance across vision tasks, owing to their powerful attention mechanisms that can effectively capture long-range dependencies~\cite{DosovitskiyB0WZ21, LiuL00W0LG21, CaronTMJMBJ21, XiaPSLH22, 000100LS23}. In image restoration, IPT~\cite{Chen000DLMX0021} was among the first ViT-based solutions. Unfortunately, its dependence on large-scale paired datasets for training restricts its applicability when data is scarce. Subsequent models have sought to mitigate this limitation with more efficient architectures. Uformer~\cite{WangCBZLL22} combined UNet with a window attention, yielding strong results, while Restormer~\cite{ZamirA0HK022} introduced transposed attention for efficient channel-wise processing. Further, X-Restormer~\cite{chen2023comparative} extended Restormer by integrating the OCA~\cite{ChenWZ0D23} module, providing a generalized framework for image restoration through rigorous analysis.
Specific to the domain of shadow removal, ShadowFormer~\cite{GuoHLCW23} introduced an attention mechanism that facilitates interaction between shadow and non-shadow regions, while HomoFormer~\cite{0002FZL00Z24} improved the transformer by integrating a shuffling mechanism to its transformer-based encoder. While these ViT-based models have achieved promising results with lower computational demands than diffusion models, they employ fixed-size windows for information interaction, which may limit their ability to capture the contextual information necessary for shadows. Most recently, RASM~\cite{liu2024regional} refreshed the window-based attention with a sliding-window mechanism, achieving more accurate results. Even though, as discussed in Sec.~\ref{sec:intro}, there remains significant room for the community to improve the effectiveness of shadow removal.

%% file: sec/3_method.tex
\section{Problem Analysis}

Degradation in umbra primarily encompasses reduced brightness, color distortion, and texture degradation (\textit{e.g.}, noise and blurriness), as illustrated in Fig.~\ref{fig:deg}. To better understand the types of involved degradation, we begin by formulating shadow effects via a modified image formation model as proposed in \cite{barrow1978recovering, ShorL08, LeS22}. In this model, the observed pixel intensity \( I(x, \lambda) \) at a given spatial coordinate \( x \) and wavelength \( \lambda \) is expressed as the product of shading \( S(x, \lambda) \) and albedo \( A(x, \lambda) \) as follows:
\begin{equation}
    I(x, \lambda) = S(x, \lambda)\cdot A(x, \lambda).
\end{equation}
In scenes with a single dominant light source (\textit{e.g.}, sunlight in outdoor settings), the shading for a non-shadow (lit) pixel \( x \) can be decomposed into direct \( S_d(x, \lambda) \) and ambient \( S_a(x, \lambda) \) components:
\begin{equation}
    S(x, \lambda) = S_d(x, \lambda) + a(x)\cdot S_a(x, \lambda),
\end{equation}
where $a(x)\in[0,1]$ designates a spatially varying attenuation function.
As a consequence, the intensity in lit regions \( I_{\text{lit}}(x, \lambda) \) can be simply written as:
\begin{equation}
    I_{\text{lit}}(x, \lambda) = (S_d(x, \lambda) + S_a(x, \lambda)) \cdot A(x, \lambda).
    \label{eq:lit}
\end{equation}
Alternatively, when an object occludes the primary light source, creating shadow regions, the reflected intensity, \( I_{\text{shadow}}(x, \lambda) \), can be calculated through:
\begin{equation}
    I_{\text{shadow}}(x, \lambda) = a(x) \cdot S_a(x, \lambda) \cdot A(x, \lambda).
    \label{eq:shadow}
\end{equation}
By comparing Eq. \eqref{eq:shadow} and \eqref{eq:lit}, we can identify the origin of reduced illumination. An example of this effect is depicted in Fig.~\ref{fig:deg}(c). 
Moreover, the observed color intensity at each pixel can be integrated across wavelengths as follows:
\begin{equation}
    I_k(x) = \int_{\lambda_{\text{min}}}^{\lambda_{\text{max}}} S(x, \lambda)\cdot A(x, \lambda) \cdot P_k(\lambda) \, d\lambda,
    \label{eq:light}
\end{equation}
where \( I_k(x) \) stands for the intensity of the \(k\)-th color channel (\textit{e.g.}, red, green, or blue) at pixel \(x\), and \( P_k(\lambda) \) is the spectral sensitivity function of the camera for the \(k\)-th channel. In addition, \( [ \lambda_{\text{min}}, \lambda_{\text{max}} ] \) defines the visible spectrum range. This formulation yields the final RGB value at each pixel by summing the contributions across the visible wavelengths. 

\begin{figure}[t]
    \centering
    \begin{subfigure}{0.49\linewidth}
        \centering
        \includegraphics[width=\linewidth]{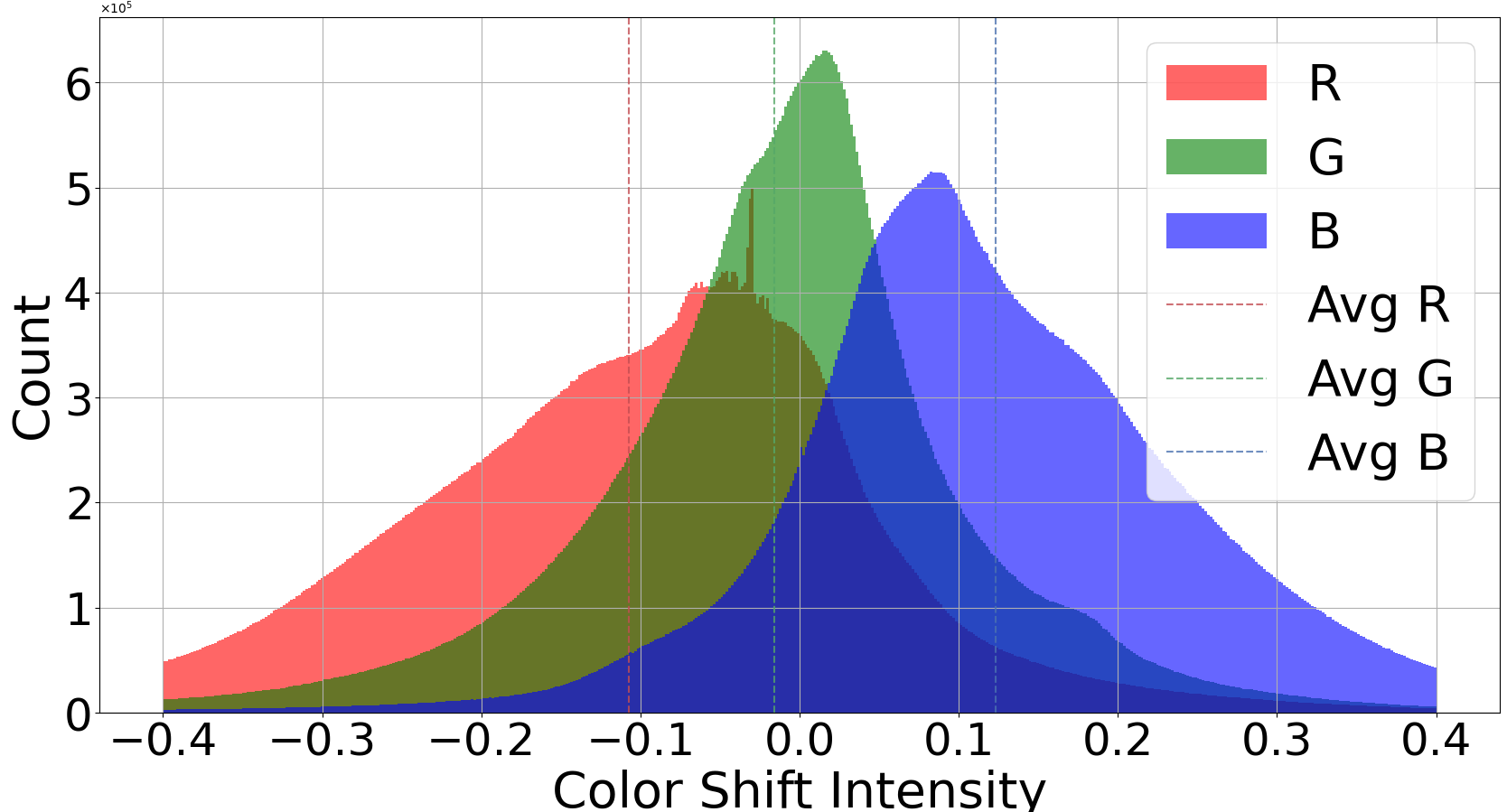}
        \caption{ISTD+ training set}
    \end{subfigure}
    \hfill
    \begin{subfigure}{0.49\linewidth}
        \centering
        \includegraphics[width=\linewidth]{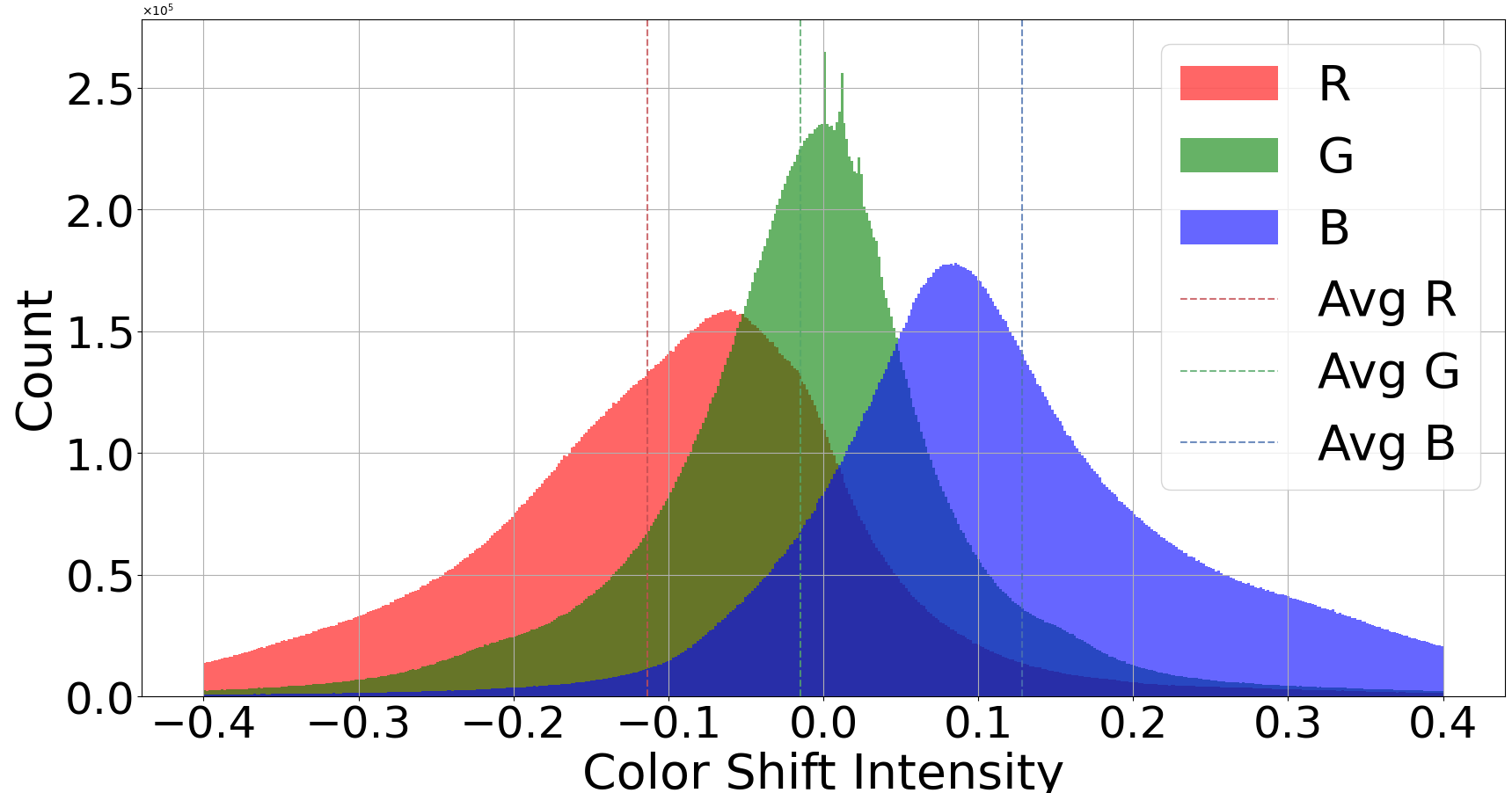}
        \caption{ISTD+ testing set}
    \end{subfigure}

    \vspace{0.3cm}
    \begin{subfigure}{0.49\linewidth}
        \centering
        \includegraphics[width=\linewidth]{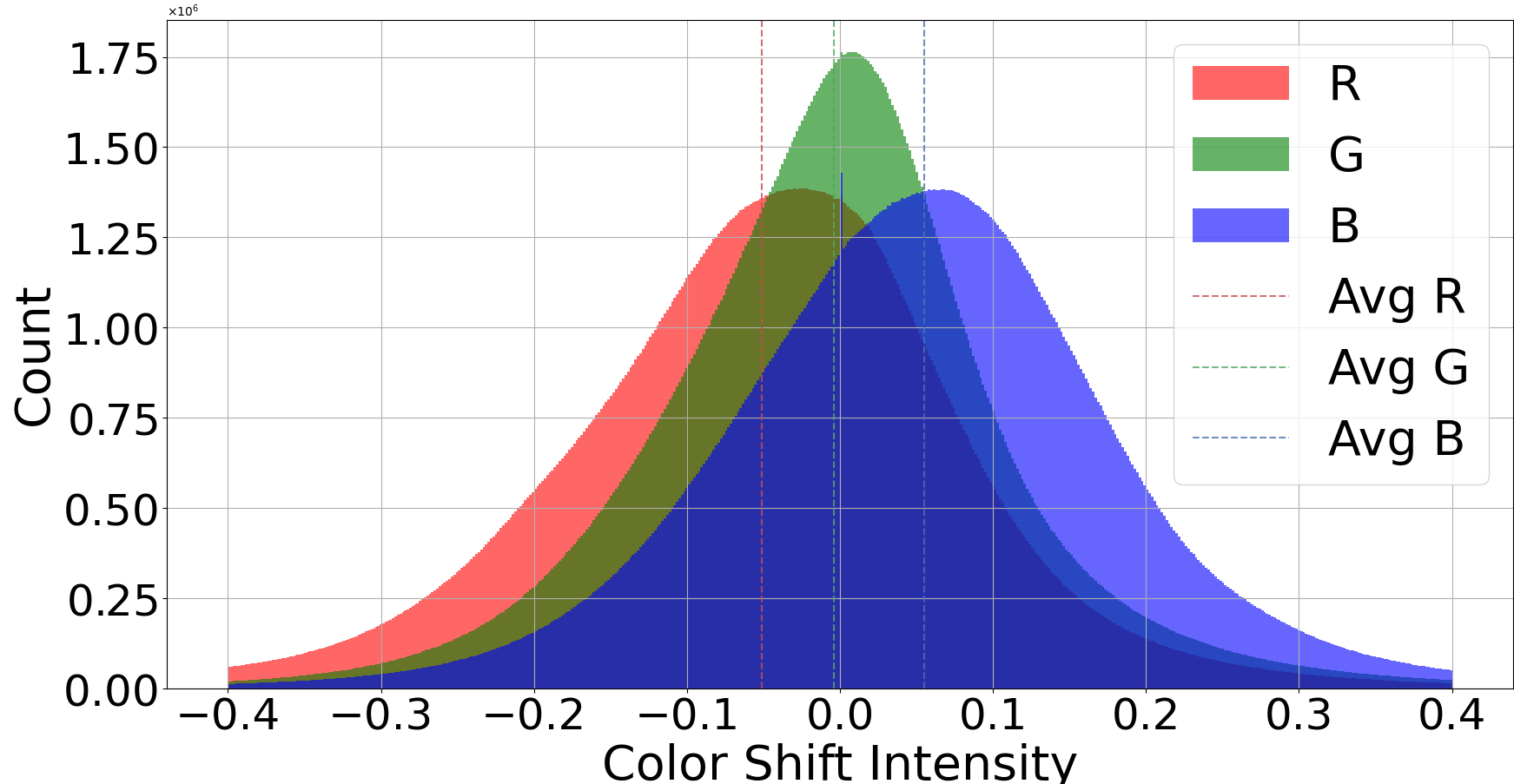}
        \caption{SRD training set}
    \end{subfigure}
    \hfill
    \begin{subfigure}{0.49\linewidth}
        \centering
        \includegraphics[width=\linewidth]{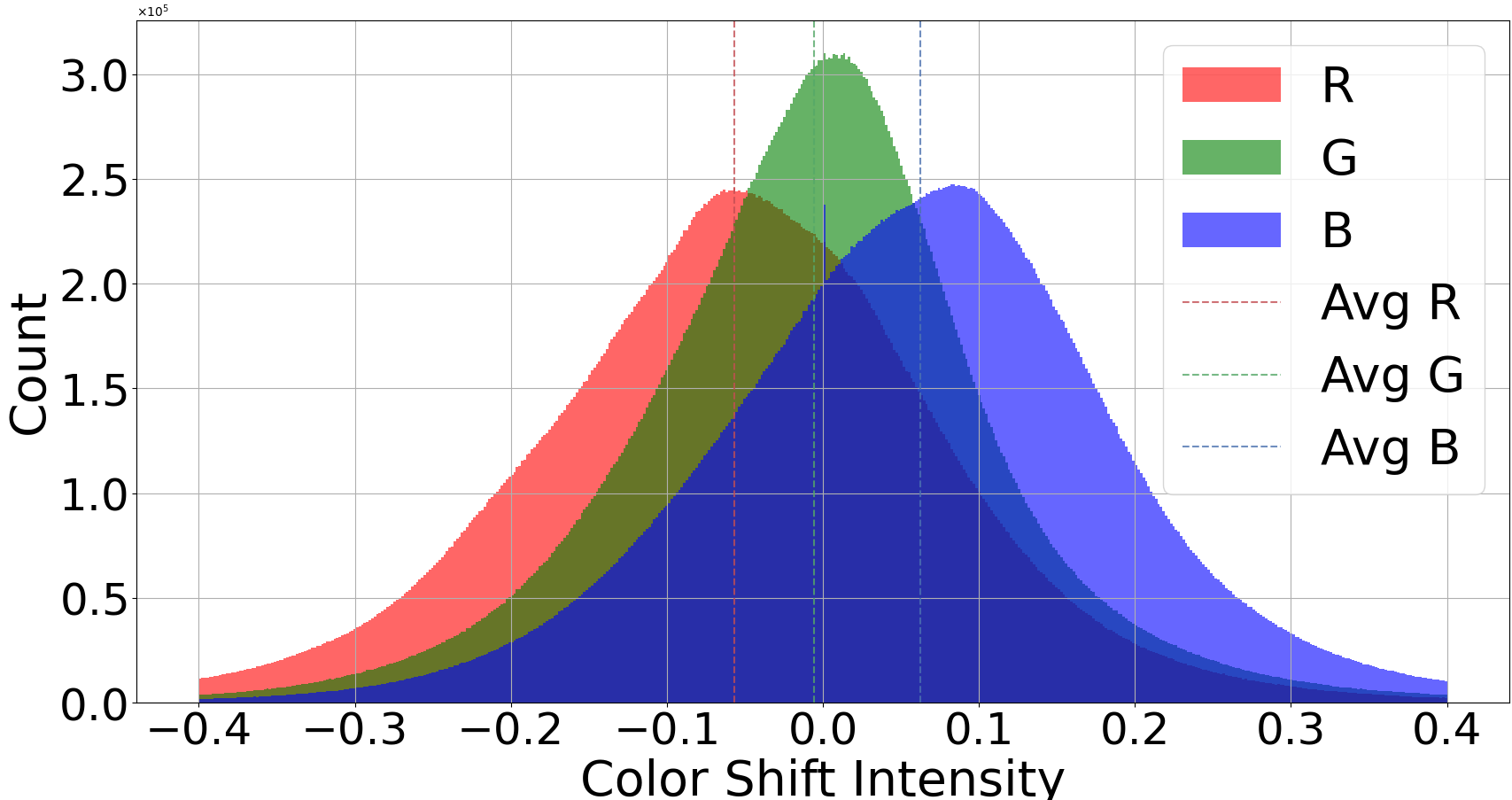}
        \caption{SRD testing set}
    \end{subfigure}

    \caption{Color bias in shadow regions between the reflectances of shadow and shadow-free images by Retinex decomposition~\cite{GuoLL17}. }
    \label{fig:intro2}
\vspace{-10pt}\end{figure}

\begin{figure*}[t]
    \centering
    \includegraphics[width=1\linewidth]{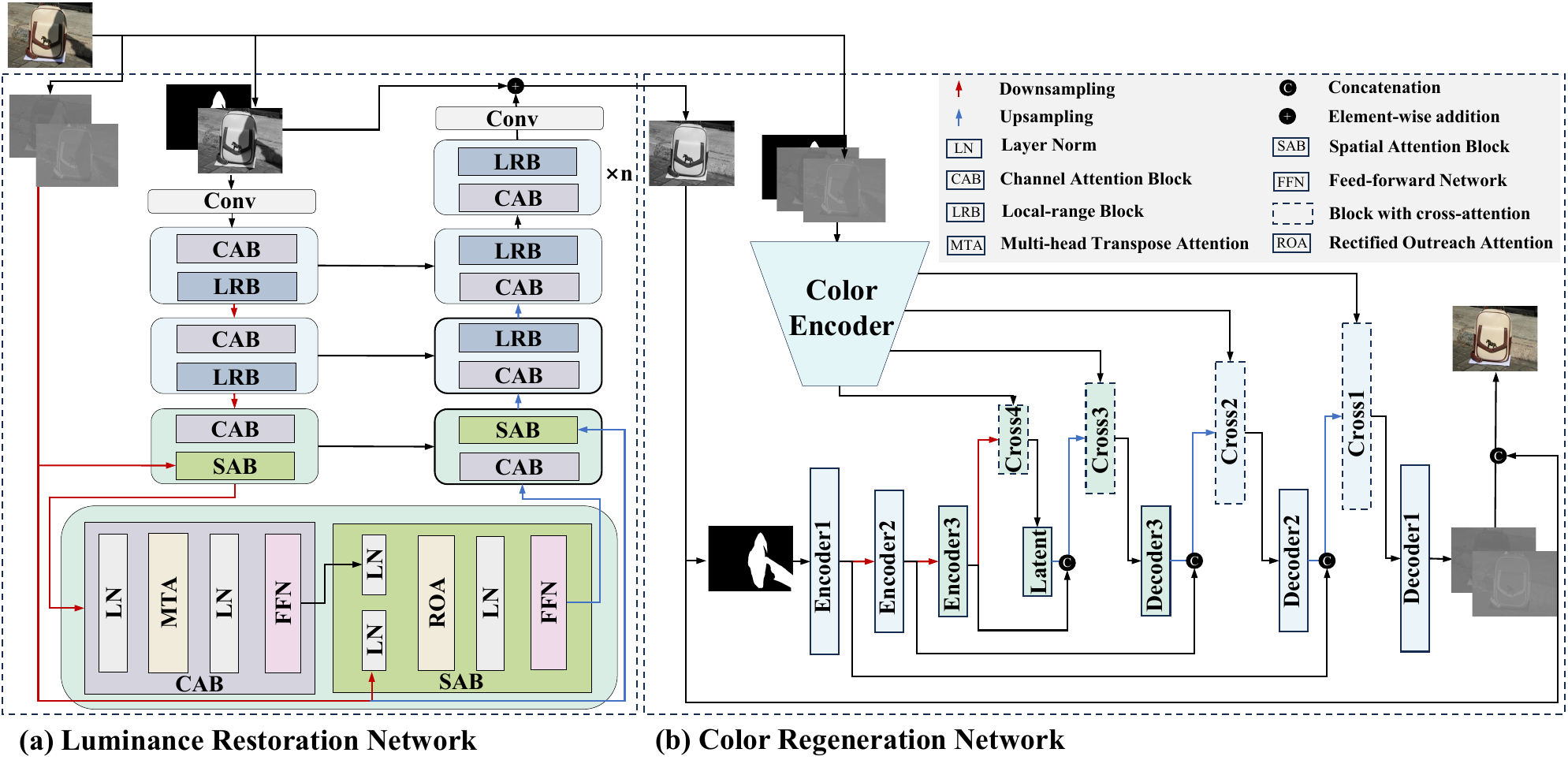}
    \vspace{-15pt}
    \caption{The overall architecture of ShadowHack comprises two core components: (a) a luminance restoration network, which focuses on recovering illumination and texture, and (b) a color regeneration network dedicated to recalibrating harmonized colors for natural outputs.}
    \label{fig:method1_overview}
    \vspace{-10pt}
\end{figure*}

To kindle a poorly lit image, Retinex-based methods assume a uniform illumination \( L(x) \) across all wavelengths \( \lambda \). As a result, Eq.~\eqref{eq:light} simplifies to:
\begin{equation}
    I_k(x) = L(x)\cdot R_k(x) \approx S(x) \int_{\lambda_{\text{min}}}^{\lambda_{\text{max}}} A(x, \lambda) \cdot P_k(\lambda) \, d\lambda,
\end{equation}
where \( R_k(x) \) corresponds to \( \int_{\lambda_{\text{min}}}^{\lambda_{\text{max}}} A(x, \lambda) \, P_k(\lambda) \, d\lambda\), representing the target reflectance component. However, as shown in Fig.~\ref{fig:deg}(a), the color in the umbra deviates significantly from that in the lit area, suggesting that uniform illumination is insufficient to model shadow-induced degradation. To offer a comprehensive view, we plot the per-channel color bias in the reflectance map derived from Retinex decomposition in Fig.~\ref{fig:intro2}. It is evident that the reflectance extracted from the Retinex decomposition exhibits substantial color bias, which violates the illumination-invariant assumption of the Retinex theory. This color shift has to be carefully addressed in a robust shadow removal framework.

During the imaging process, additional degradation may be introduced due to various factors. For example, sensor noise, due to inherent sensor limitations, and quantization errors from analog-to-digital conversion can both erode textural details, particularly in shadowed (umbra) regions. Compression adds further artifacts, disproportionately affecting high-frequency detail areas and introducing additional distortions. For simplicity, we aggregate all these types of degradation under the term $\eta_k(x)$, leading to the final observation:
\begin{equation}
    {I}_k(x) =  \int_{\lambda_{\text{min}}}^{\lambda_{\text{max}}} S(x, \lambda) \cdot A(x, \lambda) \cdot P_k(\lambda)  \, d\lambda +\eta_k(x).
\end{equation}
As illustrated in Fig.~\ref{fig:deg}(b), textural degradation notably impacts visual quality when the shadow area is simply brightened, underscoring the compounded effects of noise, quantization, and compression on image fidelity. 

To sum up, complex shadow degradation poses a substantial challenge to achieving high-fidelity shadow removal and image restoration. Therefore, an effective shadow removal solution requires a comprehensive model capable of addressing all these degradation effects simultaneously.

\section{Methodology}
A viable solution involves separating an image into texture \( I_t \) and color \( I_c \) components for texture restoration and color correction, respectively. This separation is achieved through an invertible decoupling function \( \mathcal{D}(\cdot) \), defined as:
\begin{equation}
    (I_t, I_c) = \mathcal{D}(I).
\end{equation}
As shown in Fig.~\ref{fig:intro2}, the blue and red channels exhibit the most significant deviations, aligning with our shadow color formation analysis. The Cb and Cr channels in the YCbCr color space represent the blue and red offsets, respectively, which correspond well with the observed color bias trends in most shadows. Thus, we choose luminance to represent texture. We adjust the Y channel for brightness and texture details, then colorize to produce a shadow-free image. In this process, the color shift is isolated within the color component, while brightness and texture distortion remain in the texture component. This manner enables a divide-and-conquer strategy for shadow removal:
\begin{equation}
    \hat{I} = \mathcal{D}^{-1} \mathcal{C} (\mathcal{R} (I_t), I_c),
\end{equation}
where \( \mathcal{R}(\cdot) \) means the texture restoration function, \( \mathcal{C}(\cdot) \) is the color regeneration function, $\hat{I}$ is the shadow-free output, and \( I_c \) serves as a conditioning variable for color regeneration. Additionally, \(\mathcal{D}^{-1}(\cdot)\) is the inverse procedure of \(\mathcal{D}(\cdot)\).
The overall framework of ShadowHack is schematically depicted in Fig.~\ref{fig:method1_overview}, consisting of two main components: a luminance restoration network (LRNet $\mathcal{R}(\cdot)$) and a color regeneration network (CRNet $\mathcal{C}(\cdot)$). The processing pipeline starts by decoupling the RGB image via \(\mathcal{D}(\cdot)\), which separates it into luminance \( I_t \) and color \( I_c \) components. The decomposed luminance is then passed through LRNet to restore high-quality luminance information. Afterwards, the restored luminance \( \hat{I}_t=\mathcal{R}(I_t) \) along with the original color component \( I_c \) is fed into CRNet for color regeneration, which harmonizes the colors in the image. Finally, the estimated luminance and color components are recombined using \(\mathcal{D}^{-1}(\cdot)\) to produce the final shadow-free RGB result. In what follows, we will detail the two networks.

\begin{figure}[t]
    \centering
    \includegraphics[width=1\linewidth]{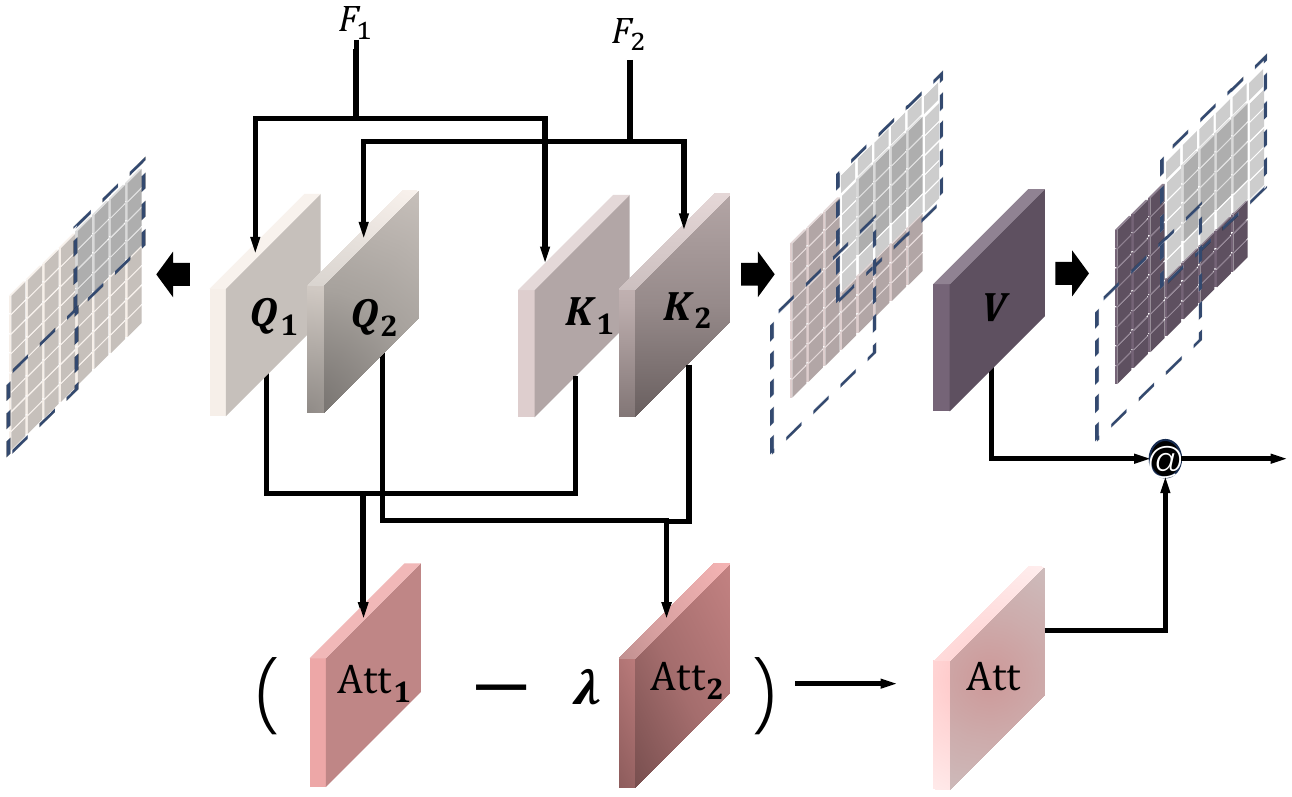}
    \caption{The rectified outreach attention module with two different window partitioning schemes.} %
    \vspace{-7pt}
    \label{fig:method_SDAM}
\vspace{-10pt}\end{figure}

\subsection{Luminance Restoration Network}

To eliminate shadows in $I_t$, enhancing the brightness of the shadow areas and restoring their textures is essential. As illustrated in Fig.~\ref{fig:method1_overview}(a), the shallow two stages and the refinement module in the luminance restoration network (LRNet) are constructed using a Local Range block (LRB), a Multi-head Transpose Attention (MTA) module, and a Feed-Forward Network (FFN), while a Rectified Outreach Attention (ROA) module replaces LRB in the deeper stages. The LRB is designed to extract local texture information through a depth-wise convolution block, capturing fine-grained details. The MTA, following previous works~\cite{ZamirA0HK022, chen2023comparative}, reallocates the weights of channels to enhance the relevant features. The FFN further processes these extracted features to improve texture restoration. Finally, the ROA module, introduced in the deeper layers, captures more global information by focusing on regions close to shadow boundaries, enabling the network to refine luminance restoration in complex shadow regions.

\noindent\textbf{Rectified Outreach Attention Module.}
In the shadow removal process, achieving robust enhancement in shadow regions requires restoration efforts to reference better surrounding, related non-shadow areas. Therefore, we replace the Local Range Block (LRB) with a Transformer module in the deeper layers to capture more distant information. 

Given the strong dependency of shadow regions on adjacent non-shadow areas, we apply overlapping cross-attention~\cite{ChenWZ0D23} to outreach spatial information interaction within this module. As depicted in the top part of Fig.~\ref{fig:method_SDAM}, indicated by the arrows, our query ($Q$) is derived from regular window partitions, while key ($K$) and value ($V$) are generated by partitioning the extracted features into outreach windows. Furthermore, we incorporate dilation in calculating $K$ and $V$ to expand the receptive field, effectively capturing more relevant non-shadow details.

\begin{figure}[t]
    \centering
    \includegraphics[width=1\linewidth]{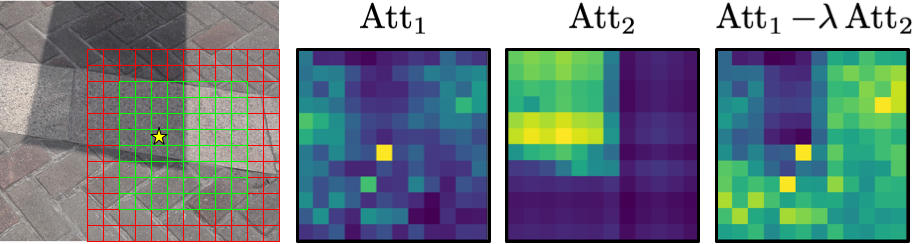}
    \caption{An example of our rectified outreach attention. The query window is in green, while the outreached key/values are in red. The final attention (c) is obtained by subtracting $\mathop{\text{Att}}_2$ from $\mathop{\text{Att}}_1$. The $\mathop{\text{Att}}_1$ and $\mathop{\text{Att}}_2$ are plotted in [0, 1], while their differential result is plotted in [-1, 1]. } 
    \label{fig:method_atten}
\vspace{-10pt}\end{figure}

\begin{table*}[t]
\centering
\footnotesize

\adjustbox{width=1\linewidth}{
    \begin{tabular}{c|l|ccc| ccc| ccc}
        \toprule
                \multirow{2}{*}{Datasets} & \multirow{2}{*}{~~~~~~~~~~~~~~~~~~~~~~Method}  & \multicolumn{3}{c|}{Shadow Region (S)}  &
                 \multicolumn{3}{c|}{Non-Shadow Region (NS)}  &
                 \multicolumn{3}{c}{All Image (ALL)} \\
                & & PSNR$\uparrow$ & SSIM$\uparrow$ & RMSE$\downarrow$ & PSNR$\uparrow$ & SSIM$\uparrow$ & RMSE$\downarrow$ & PSNR$\uparrow$ & SSIM$\uparrow$ & RMSE$\downarrow$ \\
                 \midrule
        \multirow{13}{*}{ISTD+}
        & DHAN~\cite{CunPS20} (AAAI'20)  & 33.08 & 0.988 & 9.49 & 27.28 & 0.972 & 7.39 & 25.78 & 0.958 & 7.74\\
        & G2R~\cite{LiuYWWM021} (CVPR'21) & 33.88 & 0.978 & 8.71 & 35.94 & 0.977 & 2.81 & 30.85 & 0.946 & 3.78\\
        & BMNet~\cite{ZhuHFZSZ22} (CVPR'22) & 38.17 & 0.991 & 5.72 & 37.95 & 0.986 & 2.42 & 34.34 & 0.974 & 2.93\\
        & SG-ShadowNet~\cite{WanYWWLW22} (ECCV'22) & 36.80 & 0.990 & 5.93 & 35.57 & 0.978 & 2.92 & 32.46 & 0.962 & 3.41 \\
        & ShadowFormer~\cite{GuoHLCW23} (AAAI'23) & 39.67 & 0.992 & 5.21 & 38.82 & 0.983 & 2.30 & 35.46 & 0.973 & 2.80\\
        & ShadowDiffusion~\cite{GuoWYHWPW23} (CVPR'23) & 39.82 & - & 4.90 & 38.90 & - & 2.30 & 35.72 & - & 2.70 \\
        & Li {\it et al.}~\cite{Li0A00T023} (ICCV'23) & 38.46 & 0.989 & 5.93 & 37.27 & 0.977 & 2.90 & 34.14 & 0.960 & 3.39 \\
        & Liu {\it et al.}~\cite{LiuKXLWL24} (AAAI'24) & 38.04 & 0.990 & 5.69 & 39.15 & 0.984 & 2.31 & 34.96 & 0.968 & 2.87 \\
        & Homoformer~\cite{0002FZL00Z24} (CVPR'24) & 39.47 & \textbf{0.993} & 4.72 & 38.73 & 0.984 & 2.23 & 35.34 & 0.975 & 2.64 \\
            & RASM~\cite{liu2024regional} (MM'24) & \textbf{40.73} & \textbf{0.993} & \textbf{4.41} & \underline{39.23} & \underline{0.985} & \underline{2.17} & \underline{36.16} & \underline{0.976} & \underline{2.53} \\
        & \cellcolor{gray!20}ShadowHack (Ours) & \cellcolor{gray!20} \underline{40.56} & \cellcolor{gray!20} \textbf{0.993} & \cellcolor{gray!20} \underline{4.46} & \cellcolor{gray!20} \textbf{39.66}& \cellcolor{gray!20} \textbf{0.987} & \cellcolor{gray!20} \textbf{2.09}& \cellcolor{gray!20} \textbf{36.31} & \cellcolor{gray!20} \textbf{0.977} & \cellcolor{gray!20} \textbf{2.48}\\

        \midrule
        \multirow{10}{*}{SRD}
        & DHAN~\cite{CunPS20} (AAAI'20) & 33.67 & 0.978 & 8.94 & 34.79 & 0.979 & 4.80 & 30.51 & 0.949 & 5.67\\
        & BMNet~\cite{ZhuHFZSZ22} (CVPR'22)  & 35.05 & 0.981 & 6.61 & 36.02 & 0.982 & 3.61  & 31.69 & 0.956 & 4.46\\
        & SG-ShadowNet~\cite{WanYWWLW22} (ECCV'22) & - & - & 7.53 & - & - & 2.97 & - & - & 4.23 \\
        & ShadowFormer~\cite{GuoHLCW23} (AAAI'23) & 36.91 & \underline{0.989} & 5.90 & 36.22 & 0.989 & 3.44 & 32.90 & 0.958 & 4.04\\
        & ShadowDiffusion~\cite{GuoWYHWPW23} (CVPR'23) & 38.72 & 0.987 &  4.98 &  37.78 & 0.985 & 3.44 & 34.73 & 0.970 & 3.63 \\
        & Li {\it et al.}~\cite{Li0A00T023} (ICCV'23) &  \underline{39.33} & 0.985 & 6.09 & 35.61 & 0.967 & 2.97 & 33.17 & 0.940 & 3.83 \\
        & DeS3~\cite{Jin0YYT24} (AAAI'24) & 37.45 & 0.984 & 5.88 & 38.12 & 0.988 & 2.83 & 34.11 & 0.968 & 3.72 \\
        & Homoformer~\cite{0002FZL00Z24} (CVPR'24) & 38.81 & 0.987 & \underline{4.25} & \underline{39.45} & 0.988 & 2.85 & \underline{35.37} & 0.972 & \underline{3.33} \\
        & RASM~\cite{liu2024regional} (MM'24) & 37.91 & 0.988 & 5.02 & 38.70 &  \underline{0.992} & \underline{2.72}   & 34.46 & \underline{0.976}  &  3.37\\
        & \cellcolor{gray!20}ShadowHack (Ours)  & \cellcolor{gray!20} \textbf{39.47} & \cellcolor{gray!20} \textbf{0.991} & \cellcolor{gray!20} \textbf{4.18} & \cellcolor{gray!20} \textbf{40.06} & \cellcolor{gray!20} \textbf{0.994}& \cellcolor{gray!20} \textbf{2.38}   & \cellcolor{gray!20} \textbf{35.94} & \cellcolor{gray!20} \textbf{0.982}  & \cellcolor{gray!20} \textbf{2.90}\\

\bottomrule
    \end{tabular}
}
\vspace{-5pt}
\caption{Quantitative results on the ISTD+ and SRD datasets. The best results are in \textbf{bold}, while the second-best ones are \underline{underlined}. For RASM, we re-evaluate their method with released output images from their official repository.}
    \vspace{-15pt}
\label{tab:aistd_res}
\end{table*}

However, because $I_t$ primarily contains brightness and texture, and lacks rich color information, it becomes challenging for LRNet to accurately identify suitable non-shadow areas to reference in the feature $F_t$, especially when surrounding regions exhibit similar brightness. To improve reference accuracy, we incorporate color channels into the model's attention mechanism by integrating them into $Q$ and $K$ calculations, which enables LRNet to use more informative cues in determining the regions for shadow removal. Inspired by Ye {\it et al.}~\cite{ye2024differential}, we incorporate a differential mechanism for rectification. To be specific, we additionally obtain the features $F_c$ corresponding to color component $I_c$ through convolution with downsampling. Since the luminance $I_t$ contains a color bias that may cause the model to focus on incorrect content information, we compute the softmax attention map for $F_t$ and $F_c$ and calculate a second set with only $F_c$, as shown in Fig.~\ref{fig:method_SDAM}. By subtracting these two sets, we expect to reduce potential distraction.
The query, key, and value matrices are formed via:
\begin{equation}
    \begin{aligned}
        Q_1 &= [F_t;F_c]W_1^Q, \  K_1 = [F_t;F_c]W_1^K, \\
        Q_2 &= F_cW_2^Q, \ K_2 = F_cW_2^K, \ V = F_t W^V, 
    \end{aligned}
\end{equation}
where $W$ matrices are learnable projection parameters.
Next, the above two sets of queries and keys are used to compute their respective attentions simply by: 
\begin{equation}
    \mathop{\text{Att}_i} = \mathop{\text{Softmax}}\left(\frac{Q_i K_i}{\sqrt{d}} + B\right),\ i \in \{1,2\},
\end{equation}
where $B$ denotes the learnable relative position bias~\cite{WangCBZLL22}. This allows us to apply the rectified outreach attention mechanism as follows:
\begin{equation}
    \mathop{\text{ROA}}(F_t, F_c) = \left(\mathop{\text{Att}_1} - \lambda \cdot \mathop{\text{Att}_2}\right) V.
    \label{eq:roa}
\end{equation}
In Eq.~\eqref{eq:roa}, $\lambda$ can be re-parameterize by:
\begin{equation}
    \lambda = \text{exp}(\lambda_1^1 \cdot \lambda_1^2) - \text{exp}(\lambda_2^1 \cdot \lambda_2^2) + \lambda_0,
\end{equation}
where $\lambda_i^j, i,j \in \{1,2\}$ and $\lambda_0$ are learnable and predefined parameters, respectively. As illustrated in Fig.~\ref{fig:method_atten}, such a rectified attention mechanism creates room for negative correlation in the shadow area while strong affinity with well-lit reference, urging the model to focus on relevant content. 

\subsection{Color Regeneration Network}
After restoring texture and lighting information in the luminance channel, the next task is to accurately regenerate color. Inspired by exemplar-based colorization models~\cite{LuYPZW20, LeeKLKCC20, HuangZL22}, we leverage contextual cues and condition-based guidance to improve color fidelity and ensure alignment with the semantic content of the images. Concretely, we propose a dual-encoder network, termed CRNet, which integrates a multi-scale color feature extractor into the U-shaped architecture of LRNet. The color encoder is a ConvNext-v2 atto model~\cite{0003MWFDX22} pretrained on ImageNet-21k with 2M parameters, which takes the Cb and Cr channels as input and extracts color features across multiple scales, as illustrated in Fig.~\ref{fig:method1_overview}(b).
To achieve precise color restoration, the network shall reference regions with similar/identical material properties and accurate color, using them as reliable guides for color regeneration. The outputs generated by LRNet provide qualified indexing information to facilitate this purpose. Therefore, we introduce an attention-based mechanism where luminance features serve as the query and key to compute similarity, while color features act as the value aggregating essential color information. This approach injects color features into the skip connections of the U-shaped backbone, remarkably enriching the network’s color regeneration capability.

\noindent\textbf{Checkpoint Ensemble.} We observe that the luminance network often yields better performance on the training set, which suggests the color regeneration network faces challenges in adapting to imperfect luminance outputs encountered during testing. To address this issue, we employ checkpoint ensembling during color network training, where multiple early-stage checkpoints from the luminance network training are randomly selected as candidates. This technique applied only during training, strengthens the robustness of the color network without incurring additional computational costs at inference time.

%% file: sec/4_exp.tex
\section{Experimental Validation}

\begin{figure*}[t]
    \centering
    
    \includegraphics[width=.118\linewidth]{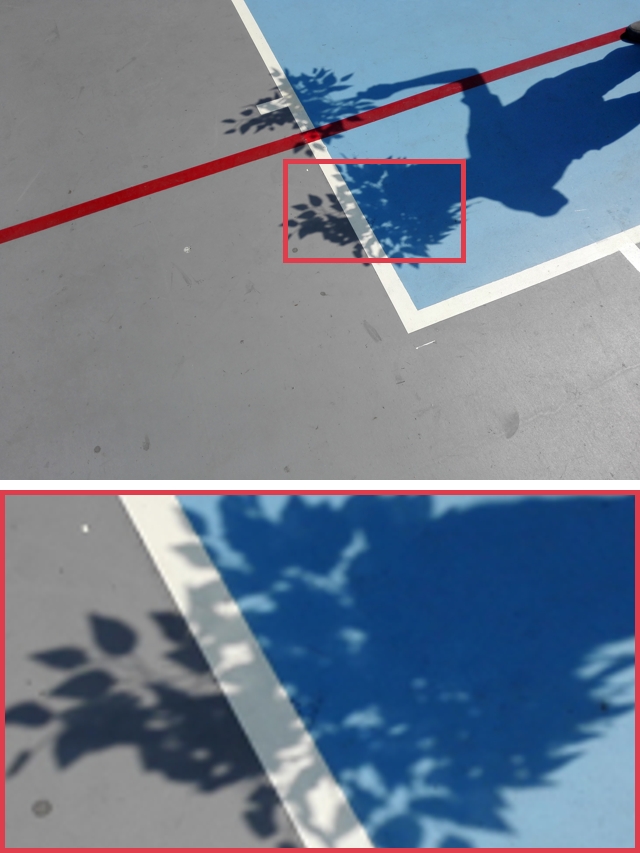}
    \includegraphics[width=.118\linewidth]{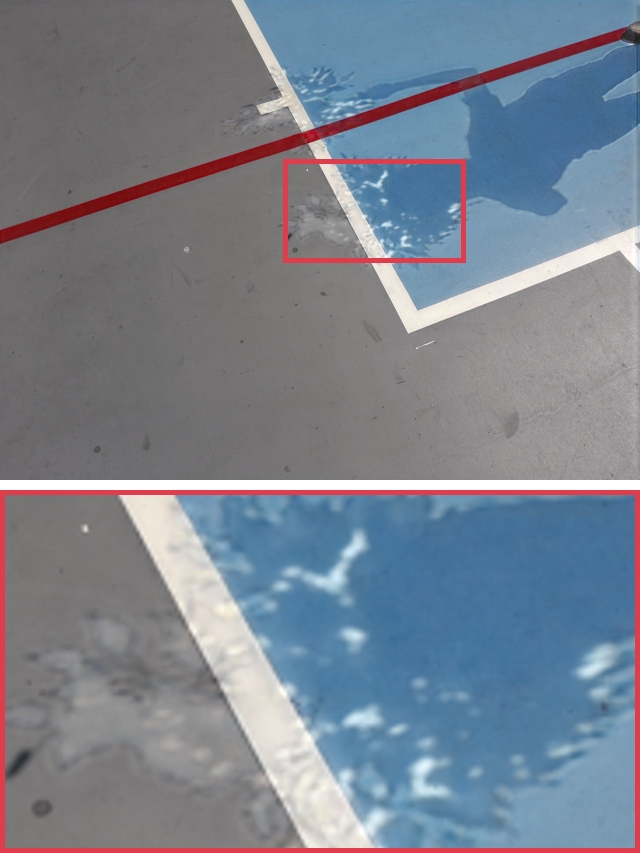}
    \includegraphics[width=.118\linewidth]{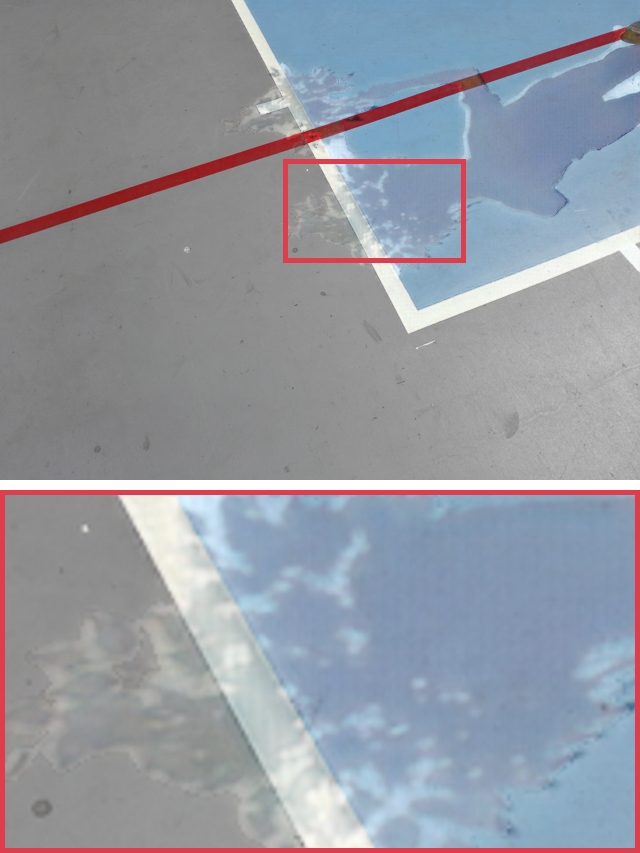}
    \includegraphics[width=.118\linewidth]{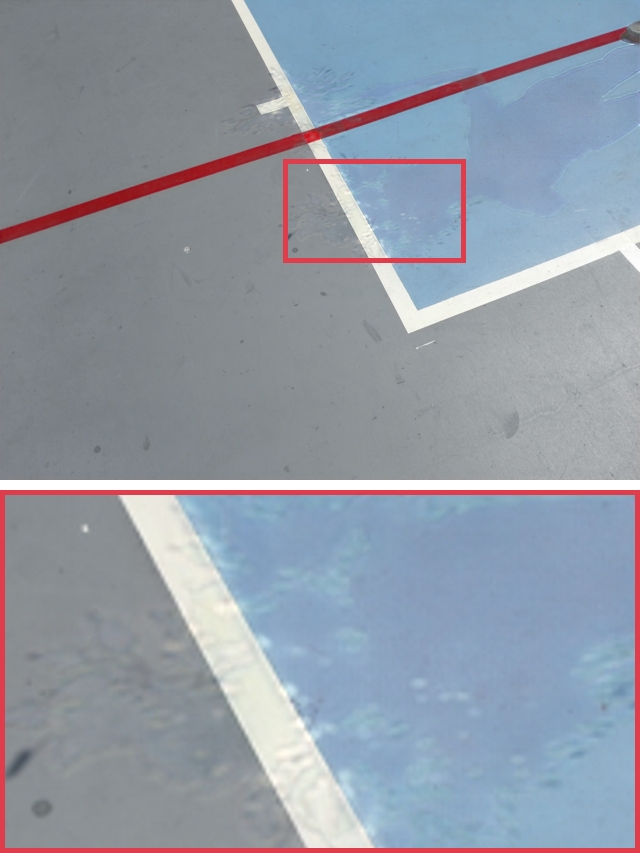}
    \includegraphics[width=.118\linewidth]{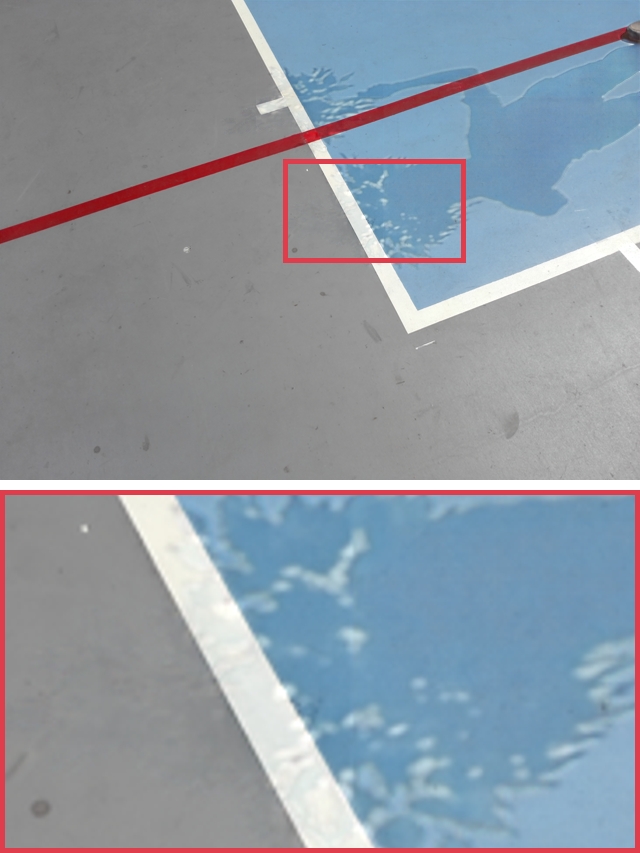}
    \includegraphics[width=.118\linewidth]{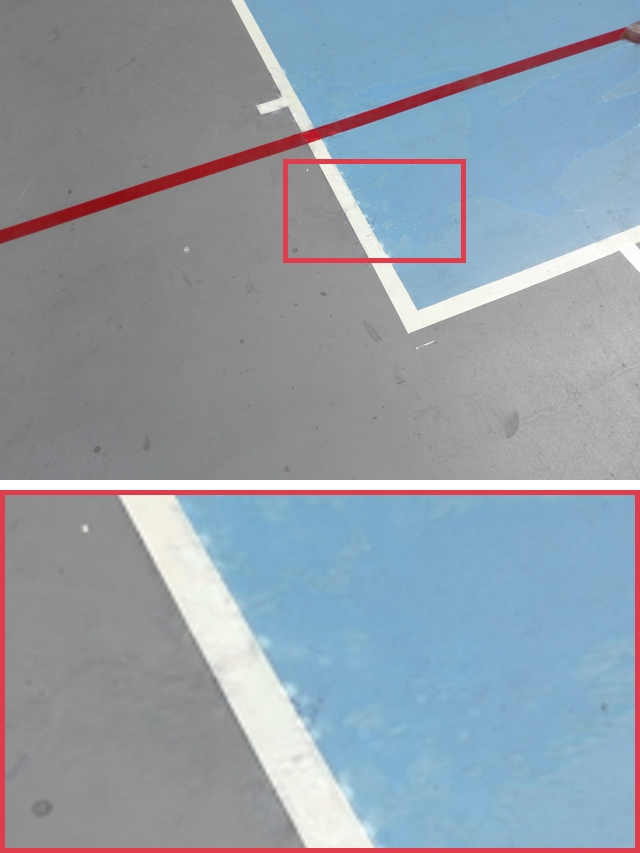}
    \includegraphics[width=.118\linewidth]{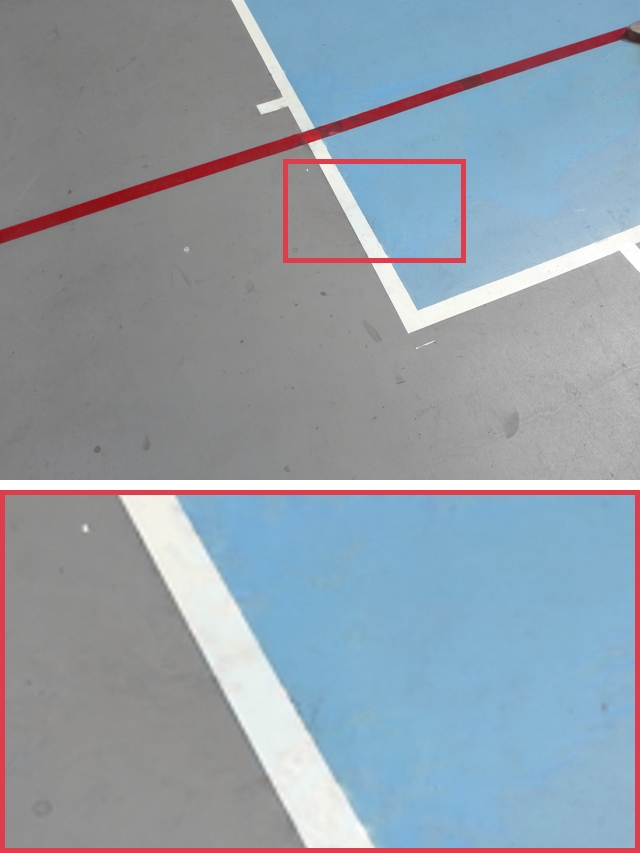}
    \includegraphics[width=.118\linewidth]{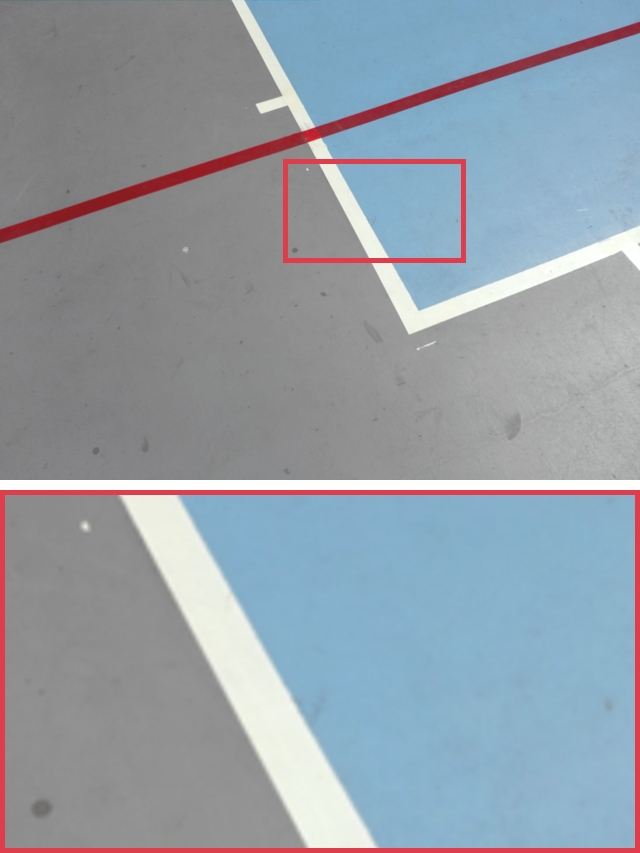}

    \includegraphics[width=.118\linewidth]{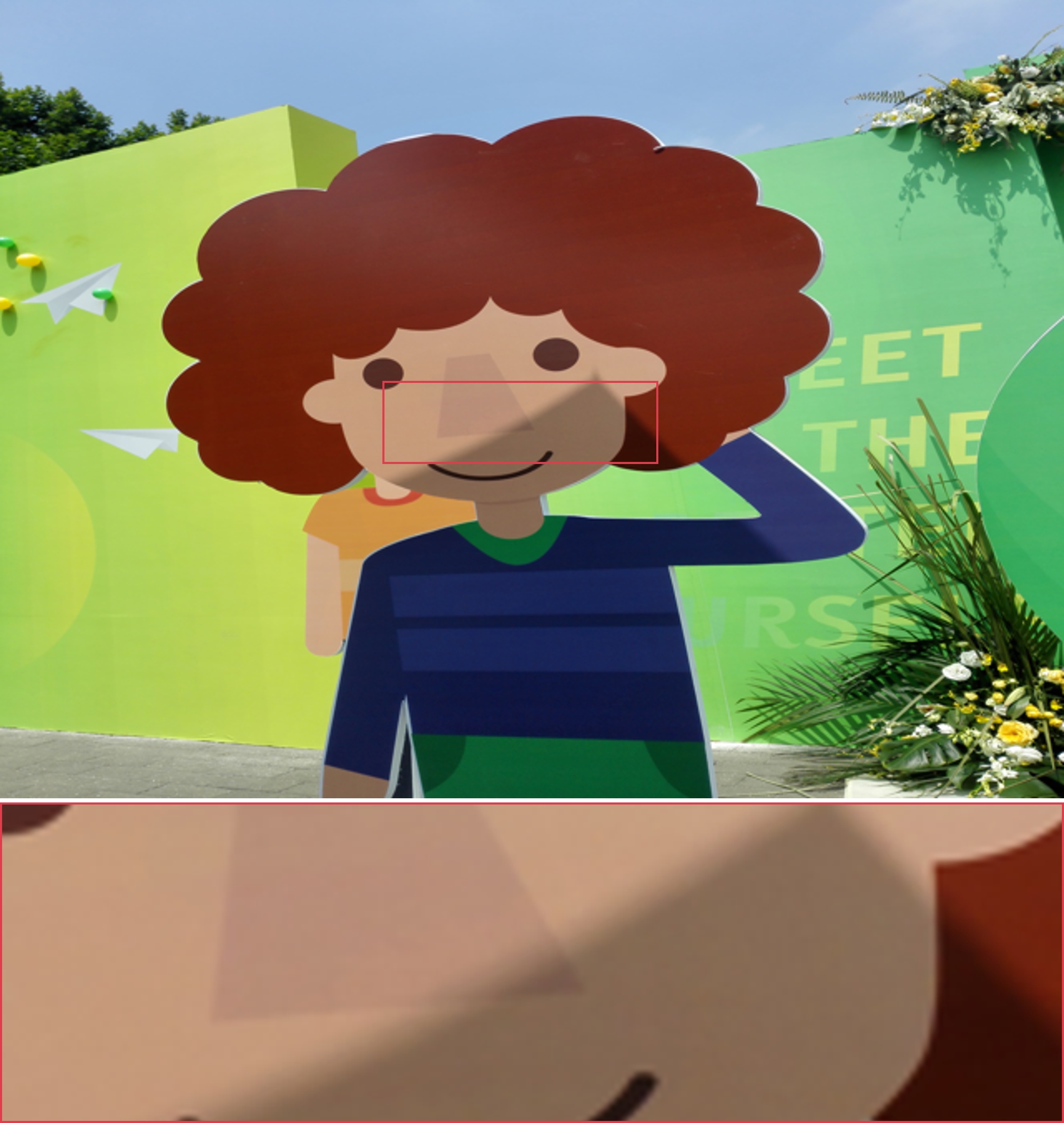}
    \includegraphics[width=.118\linewidth]{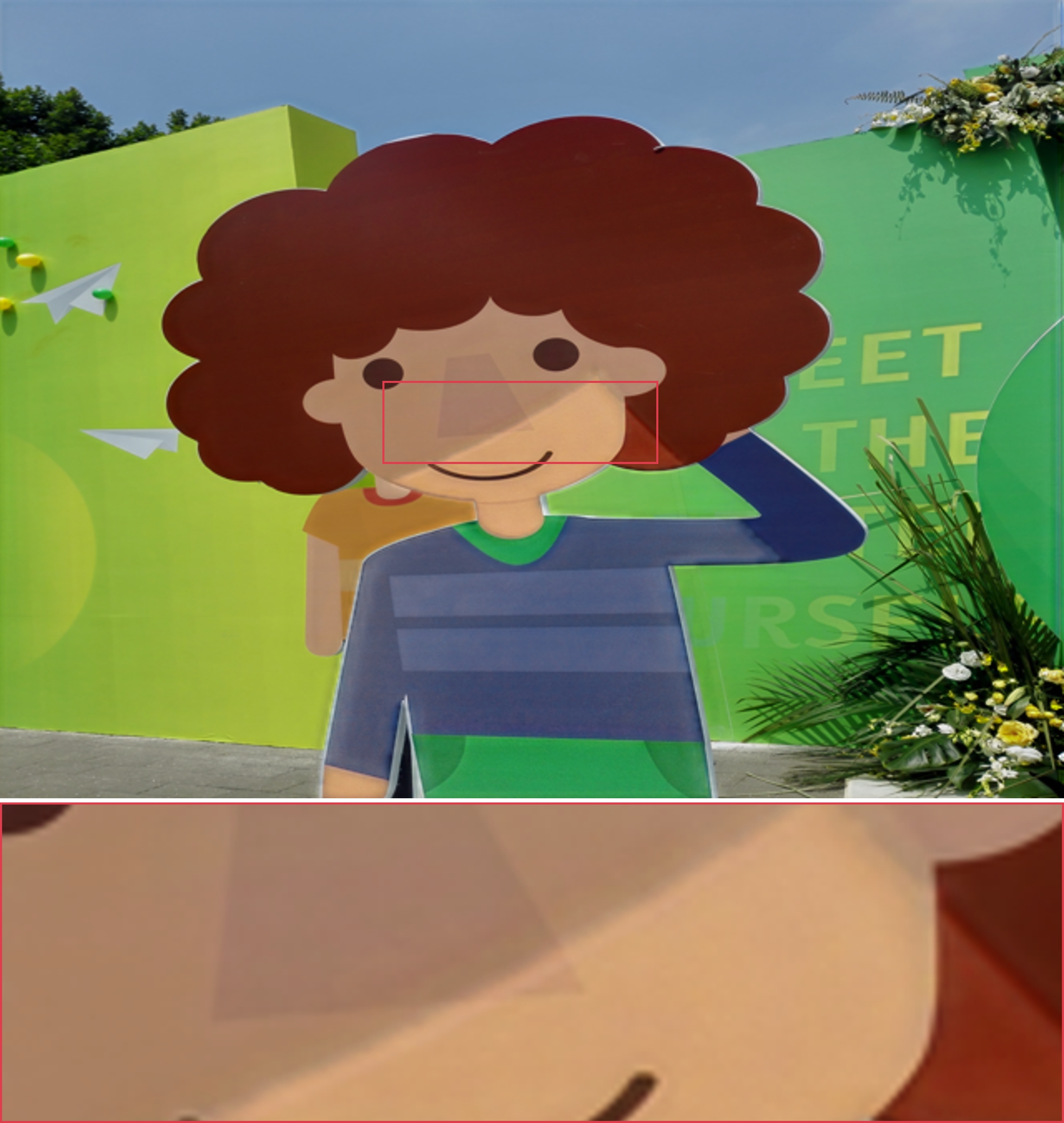}
    \includegraphics[width=.118\linewidth]{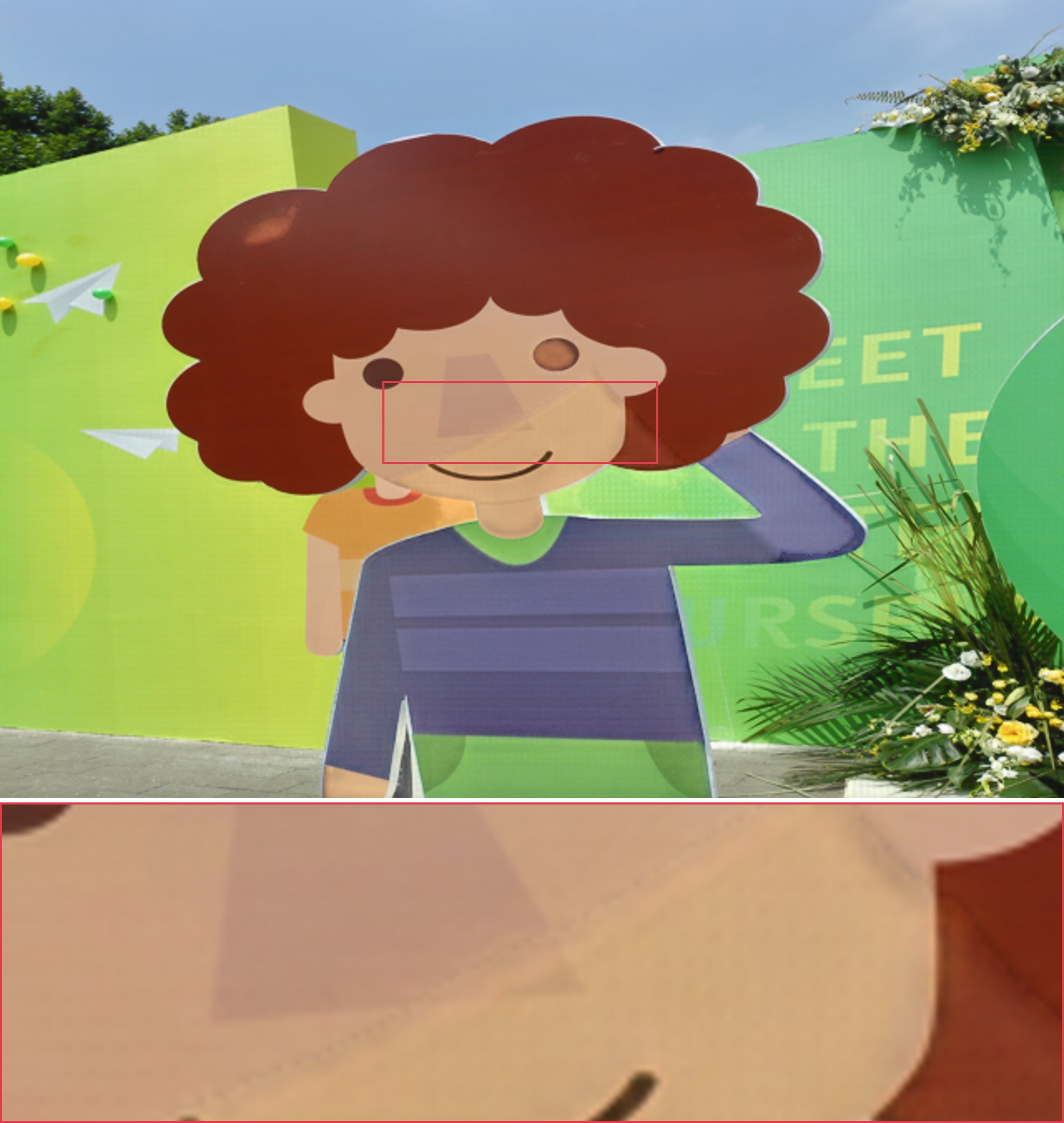}
    \includegraphics[width=.118\linewidth]{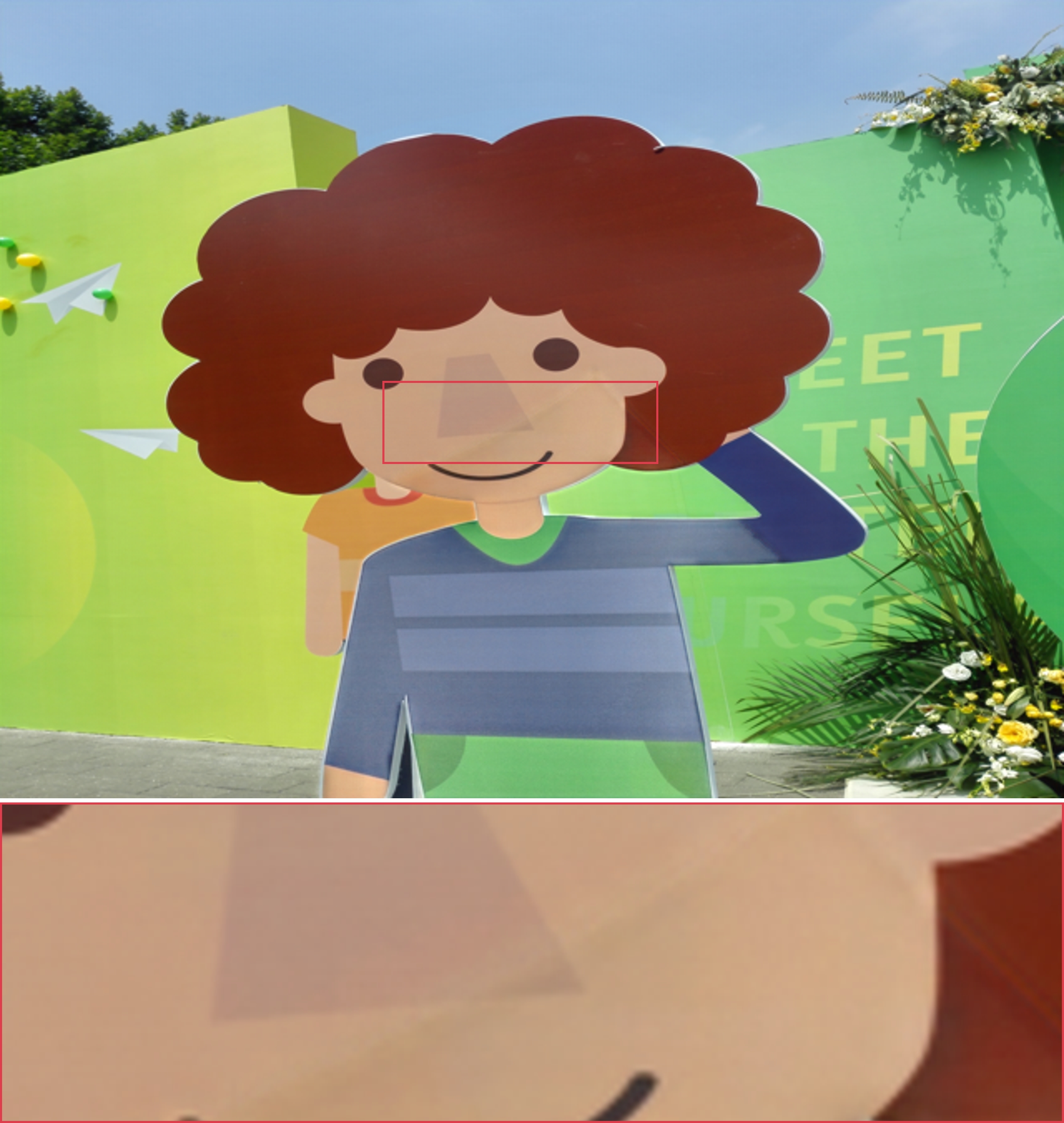}
    \includegraphics[width=.118\linewidth]{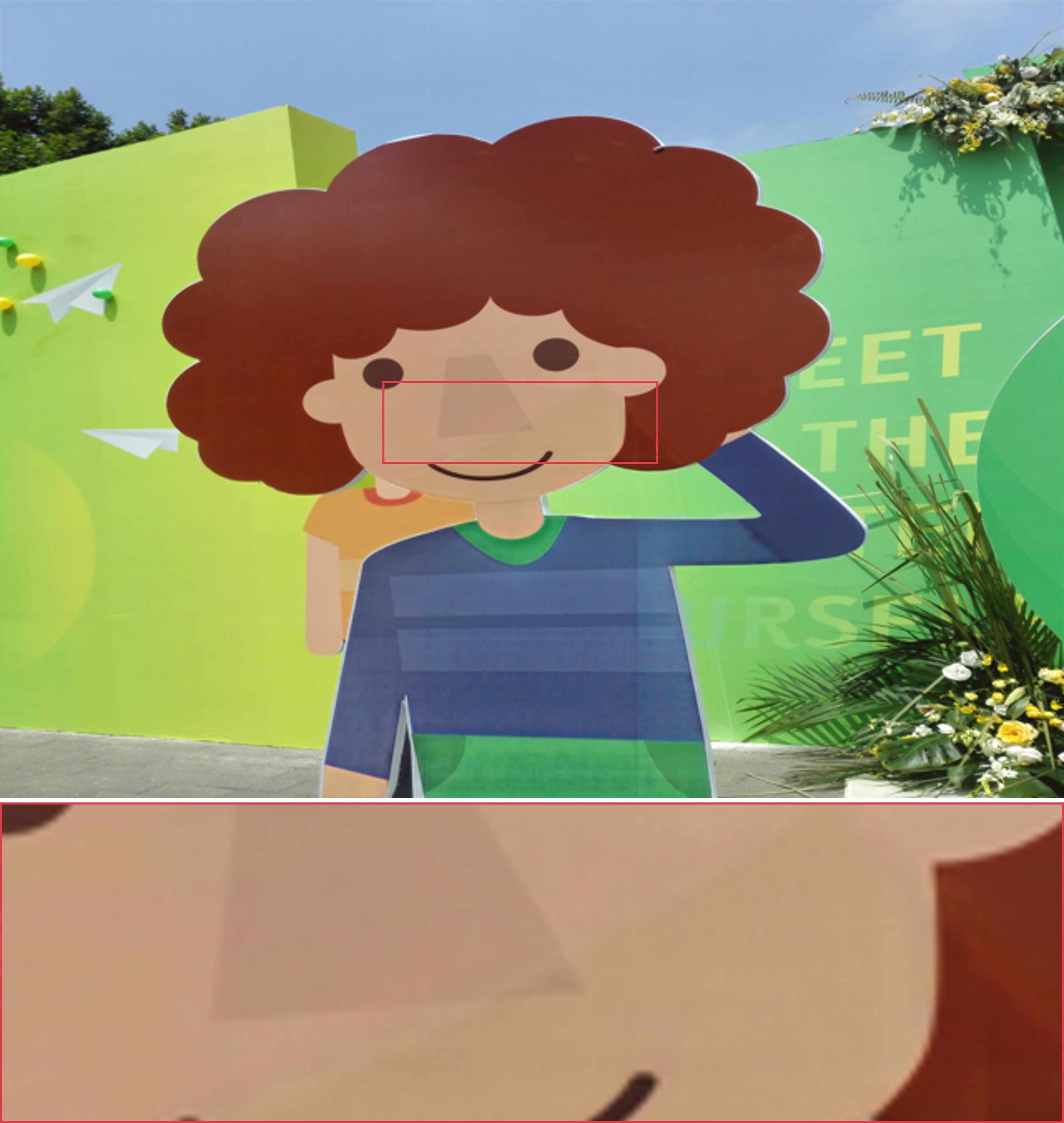}
    \includegraphics[width=.118\linewidth]{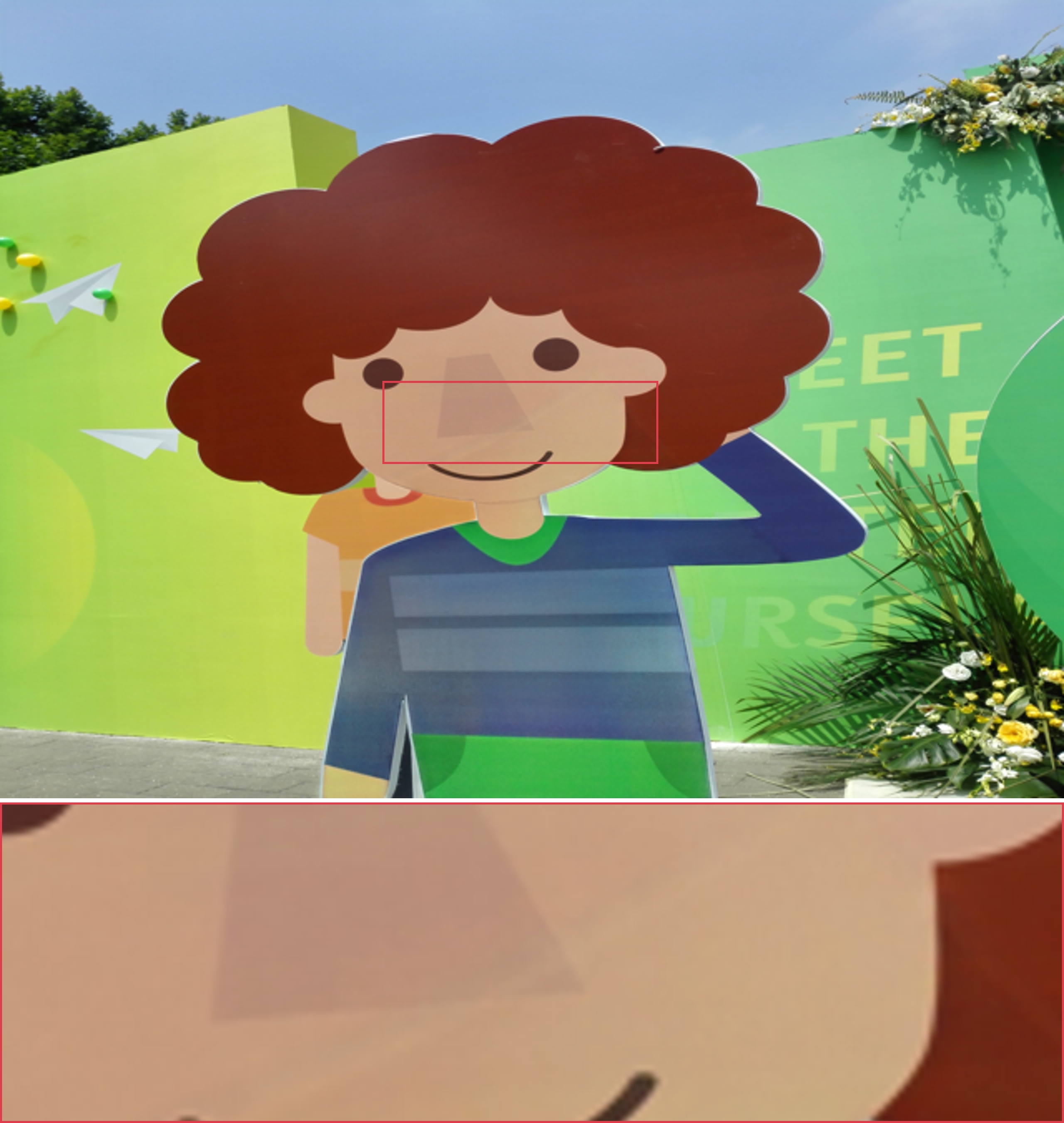}
    \includegraphics[width=.118\linewidth]{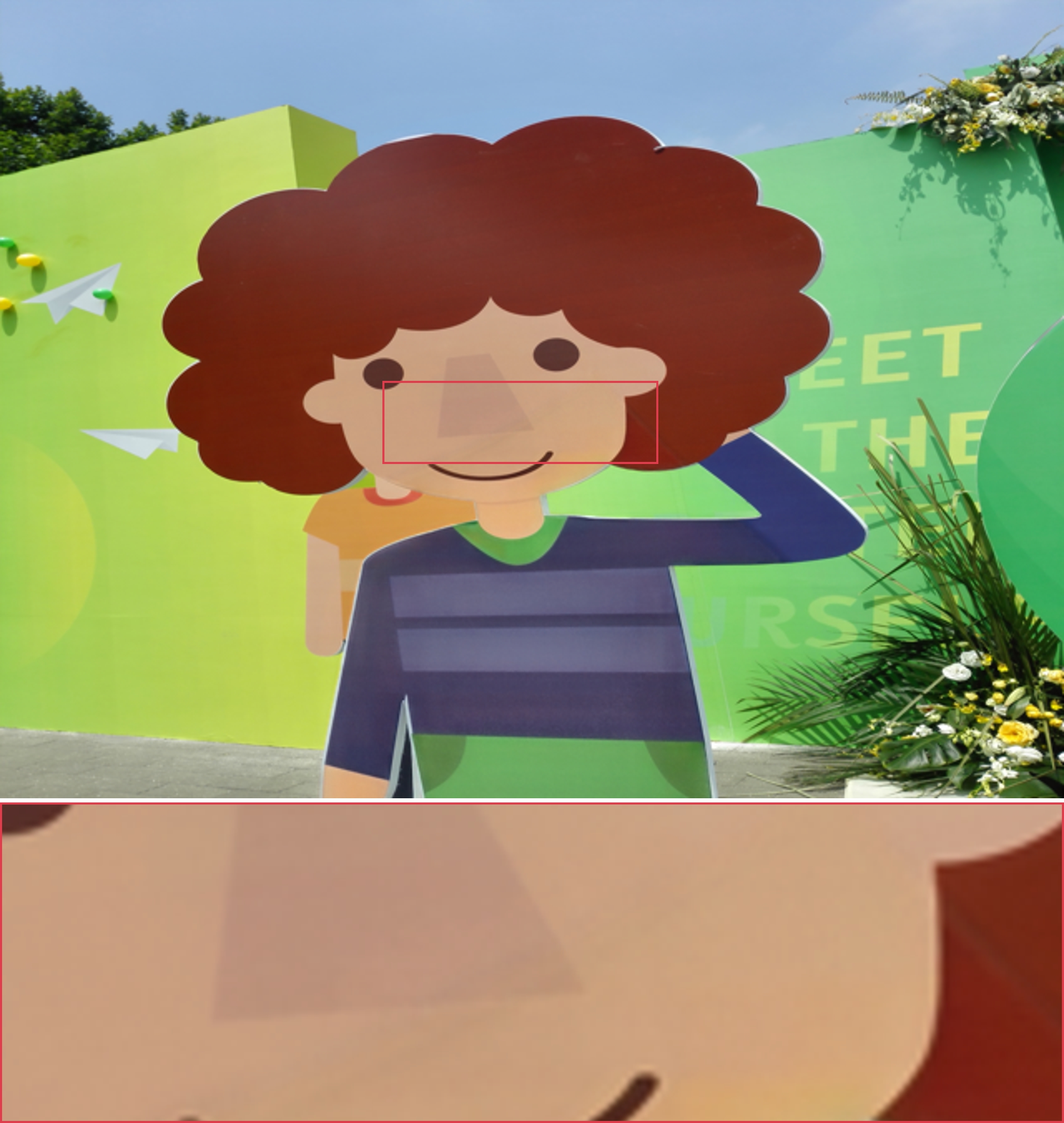}
    \includegraphics[width=.118\linewidth]{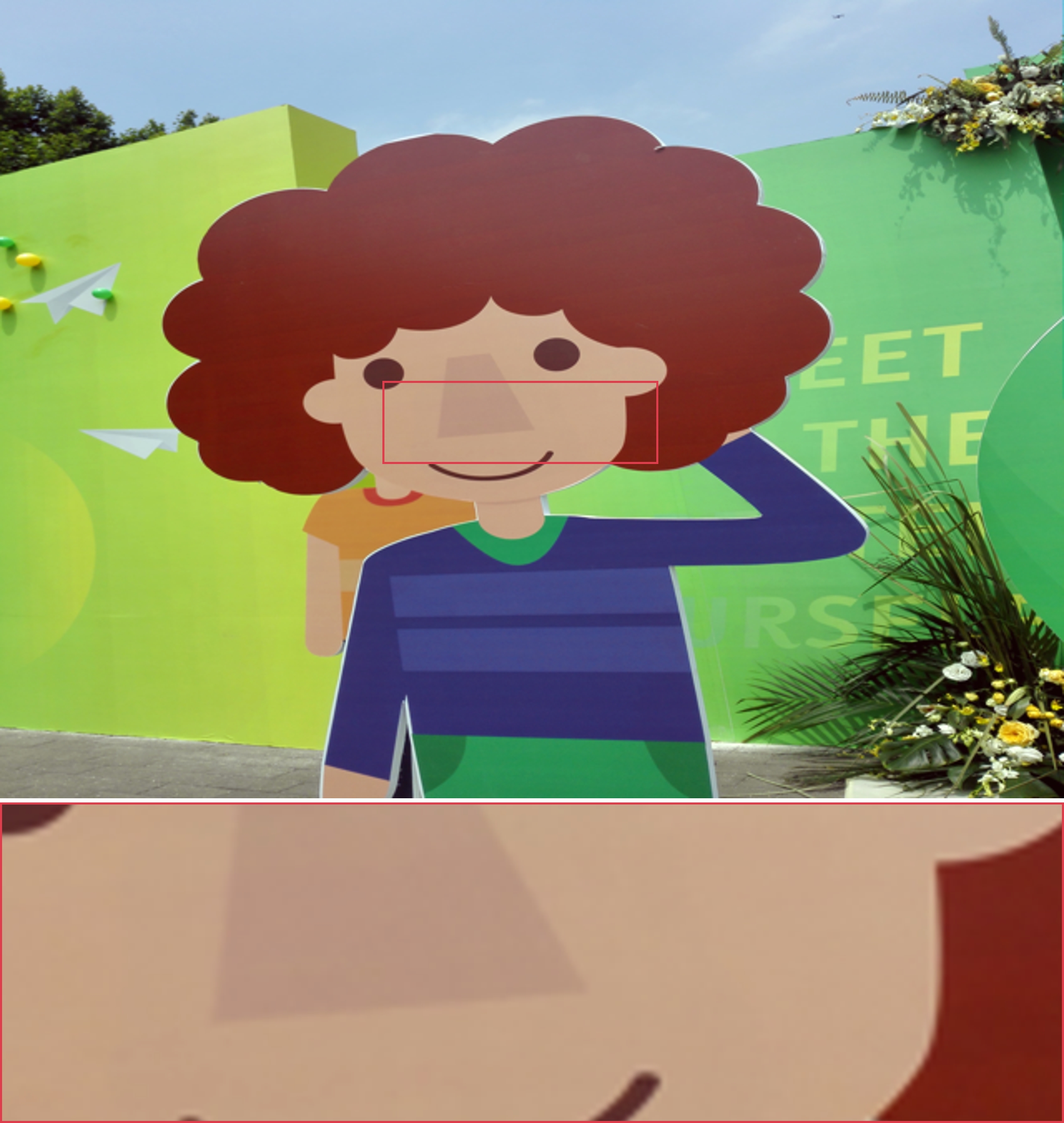}

    \begin{subfigure}{0.118\linewidth}
        \centering
        \subcaption{Input}
    \end{subfigure}
    \begin{subfigure}{0.118\linewidth}
        \centering
        \subcaption{BMNet~\cite{ZhuHFZSZ22}}
    \end{subfigure}
    \begin{subfigure}{0.118\linewidth}
        \centering
        \subcaption{SG~\cite{WanYWWLW22}}
    \end{subfigure}
    \begin{subfigure}{0.118\linewidth}
        \centering
        \subcaption{SF~\cite{GuoHLCW23}}
    \end{subfigure}
    \begin{subfigure}{0.118\linewidth}
        \centering
        \subcaption{HF~\cite{0002FZL00Z24}}
    \end{subfigure}
    \begin{subfigure}{0.118\linewidth}
        \centering
        \subcaption{RASM~\cite{liu2024regional}}
    \end{subfigure}
    \begin{subfigure}{0.118\linewidth}
        \centering
        \subcaption{Ours}
    \end{subfigure}
    \begin{subfigure}{0.118\linewidth}
        \centering
        \subcaption{GT}
    \end{subfigure}
    \vspace{-5pt}\caption{Visual comparison on a sample from the ISTD+ dataset. }
    \label{fig:exp.istd}
        \vspace{-5pt}
\end{figure*}

\begin{figure*}[t]
    \centering

    \includegraphics[width=.118\linewidth]{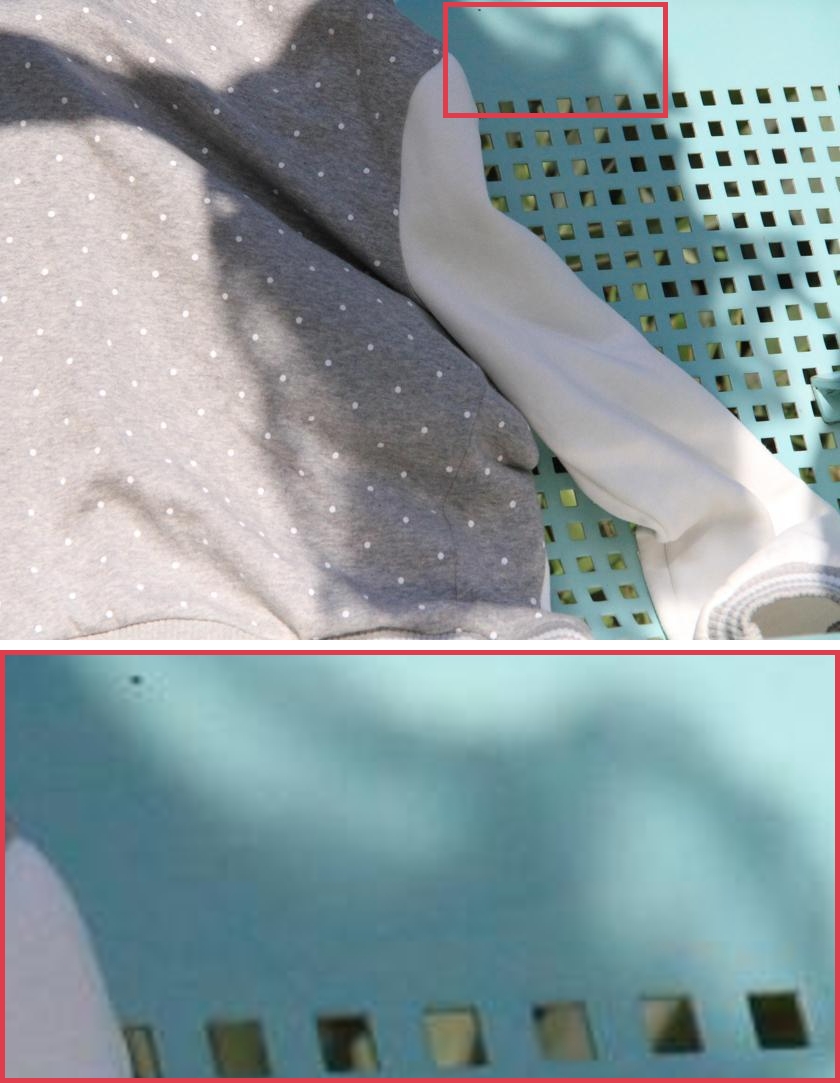}
    \includegraphics[width=.118\linewidth]{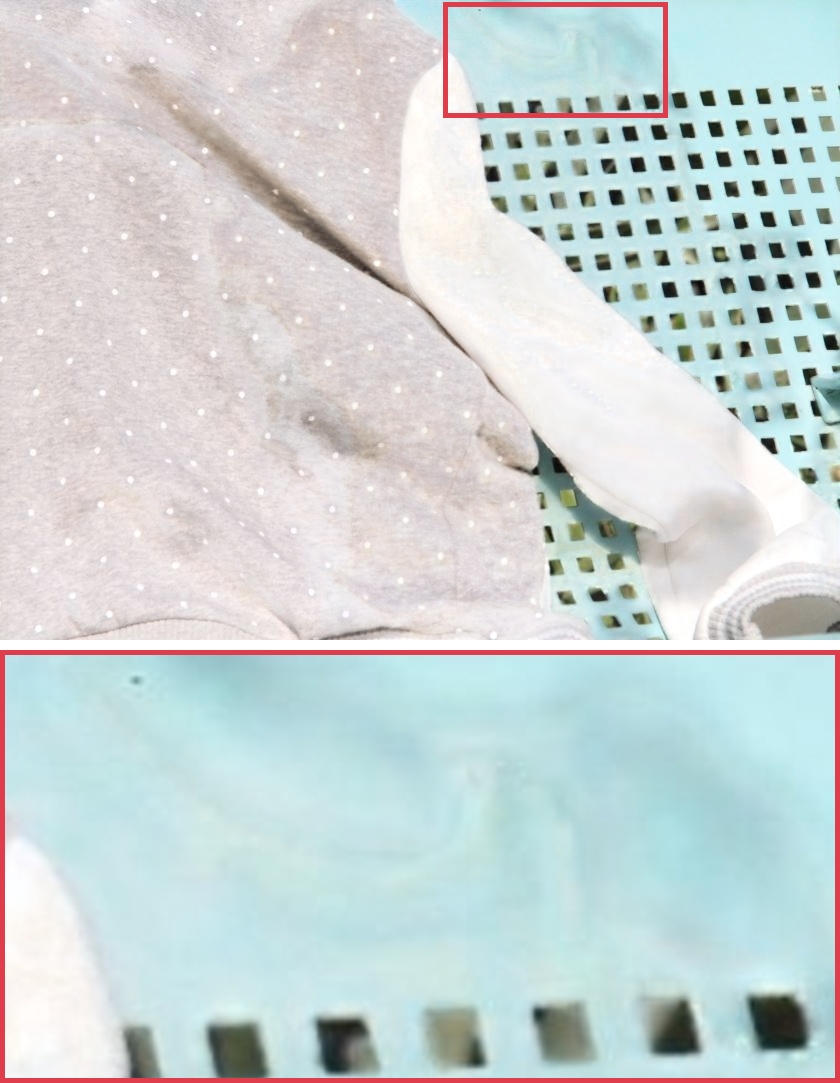}
    \includegraphics[width=.118\linewidth]{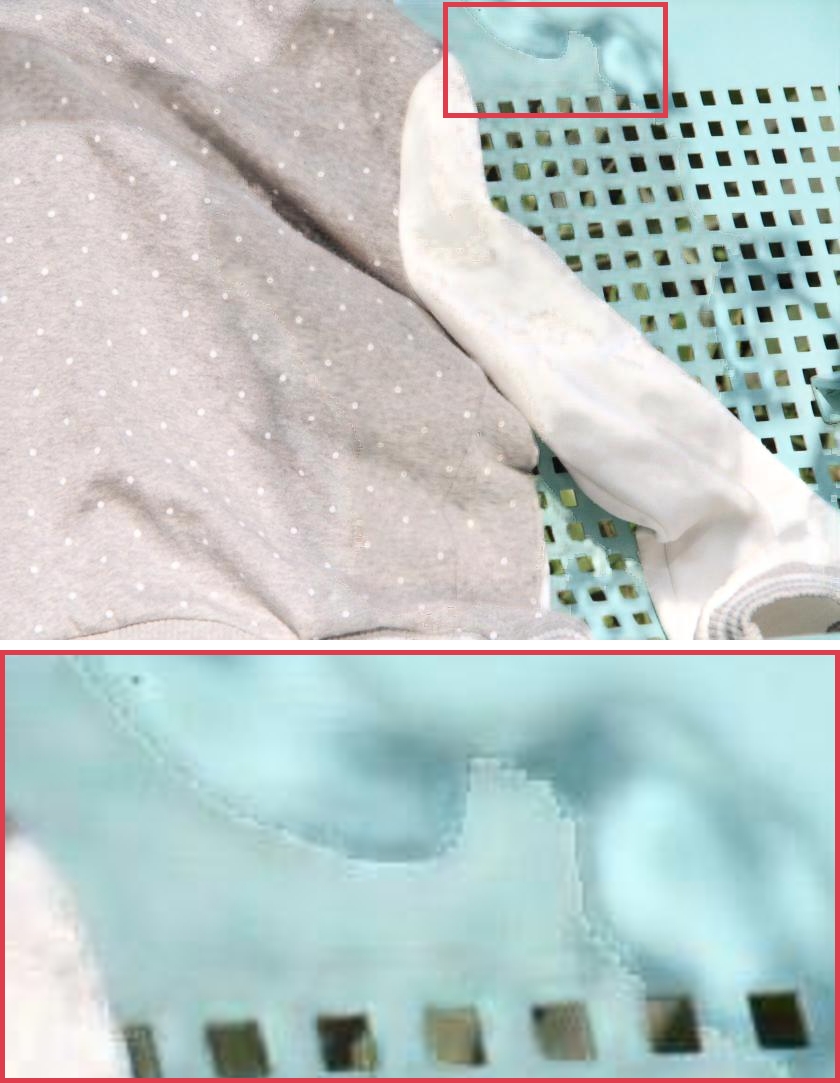}
    \includegraphics[width=.118\linewidth]{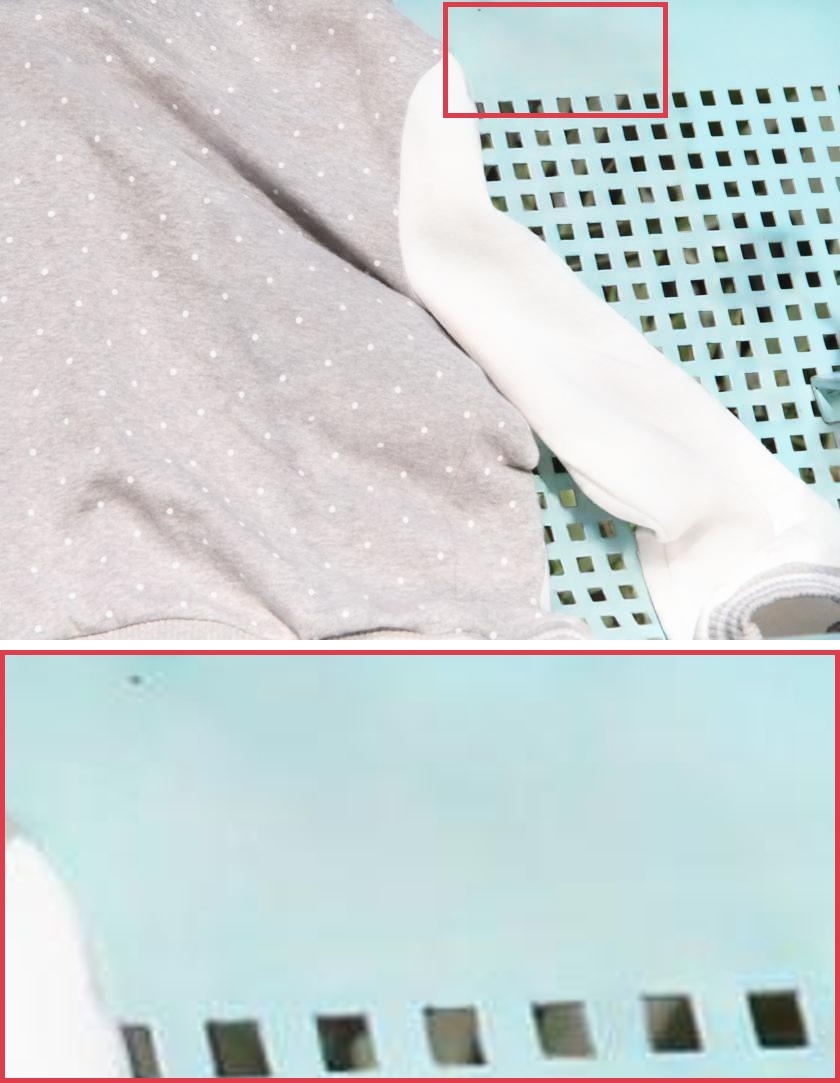}
    \includegraphics[width=.118\linewidth]{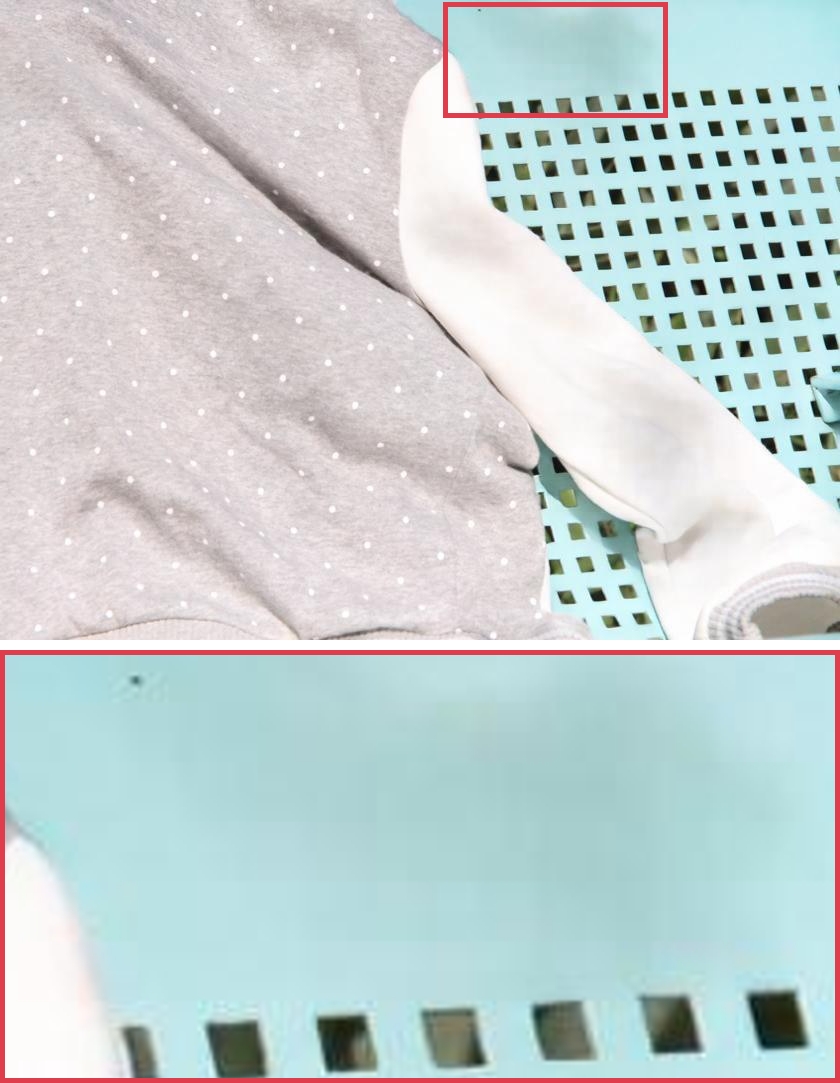}
    \includegraphics[width=.118\linewidth]{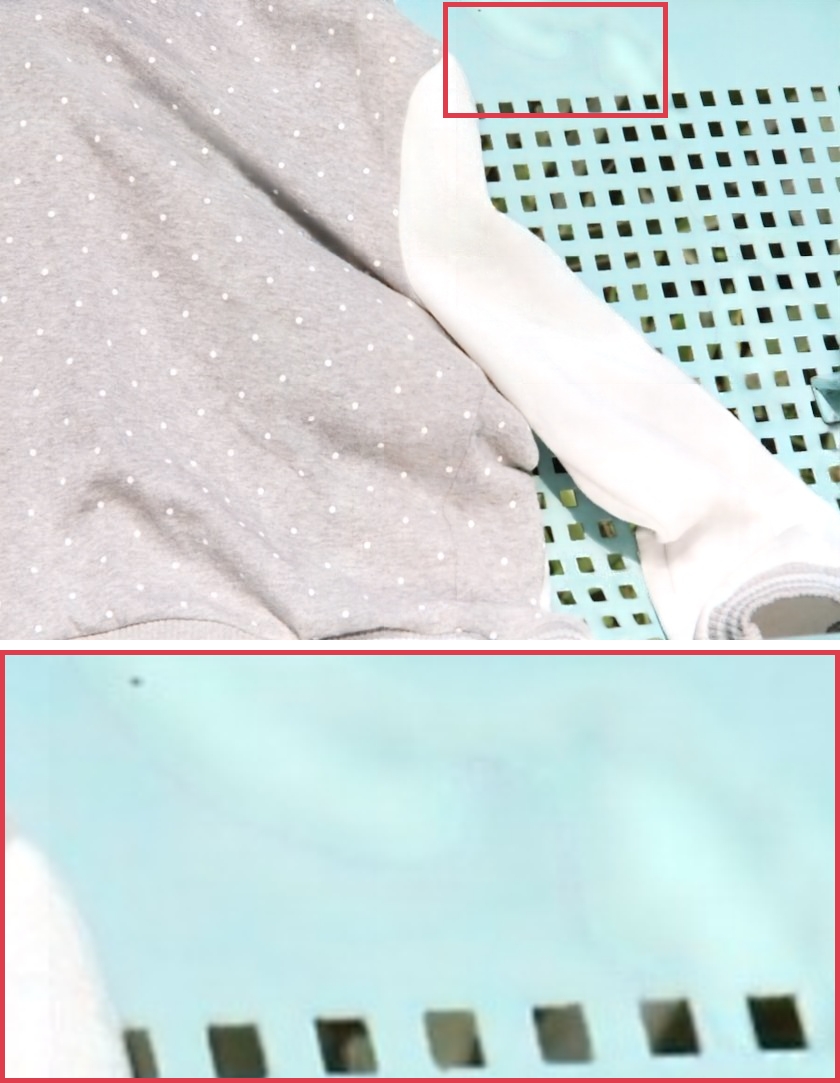}
    \includegraphics[width=.118\linewidth]{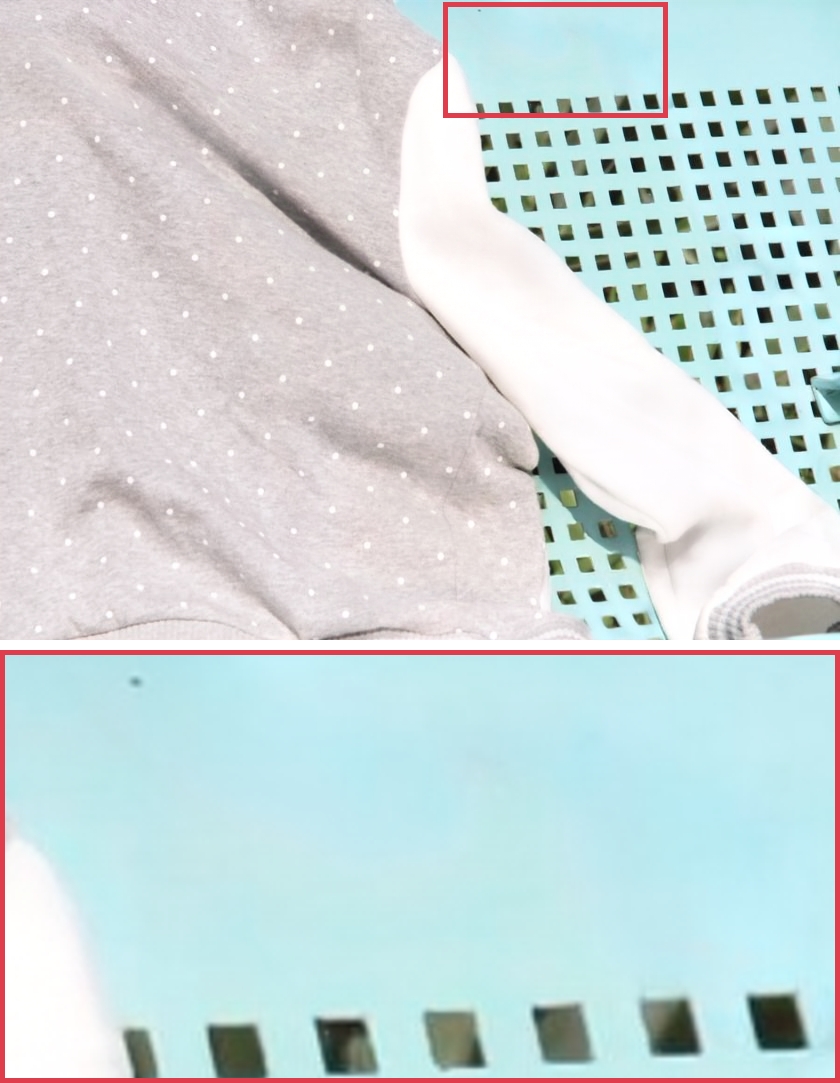}
    \includegraphics[width=.118\linewidth]{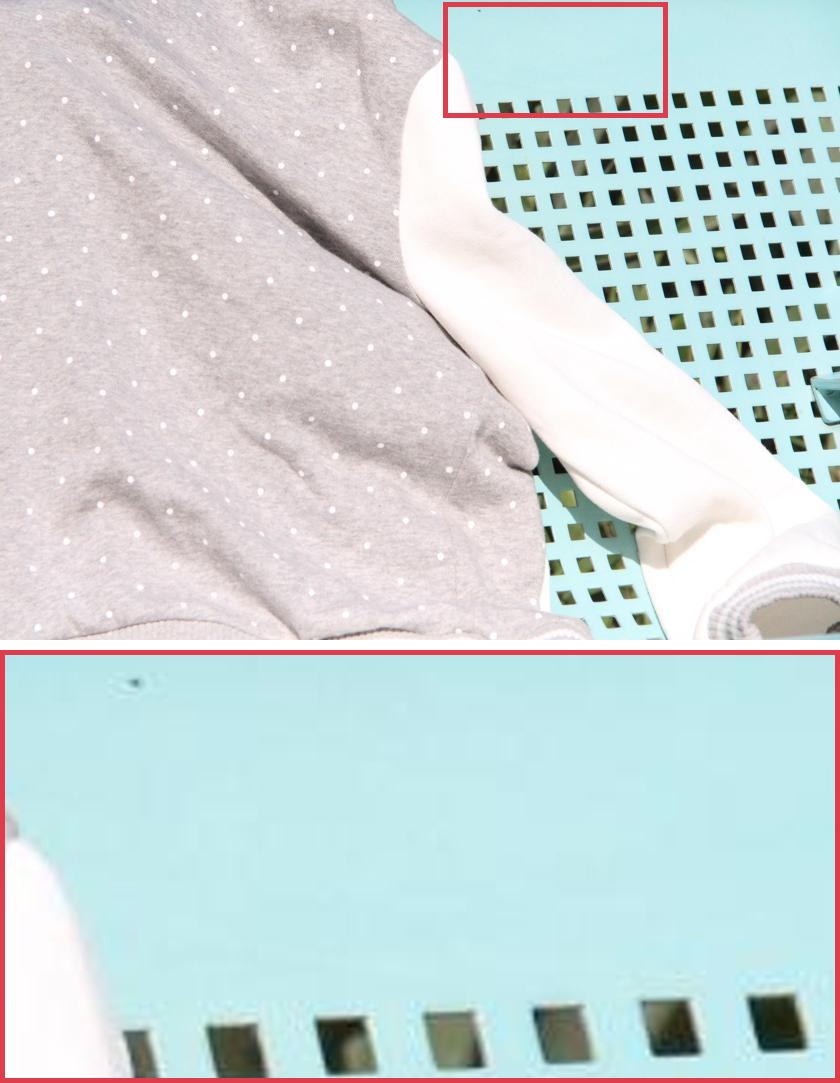}

    \includegraphics[width=.118\linewidth]{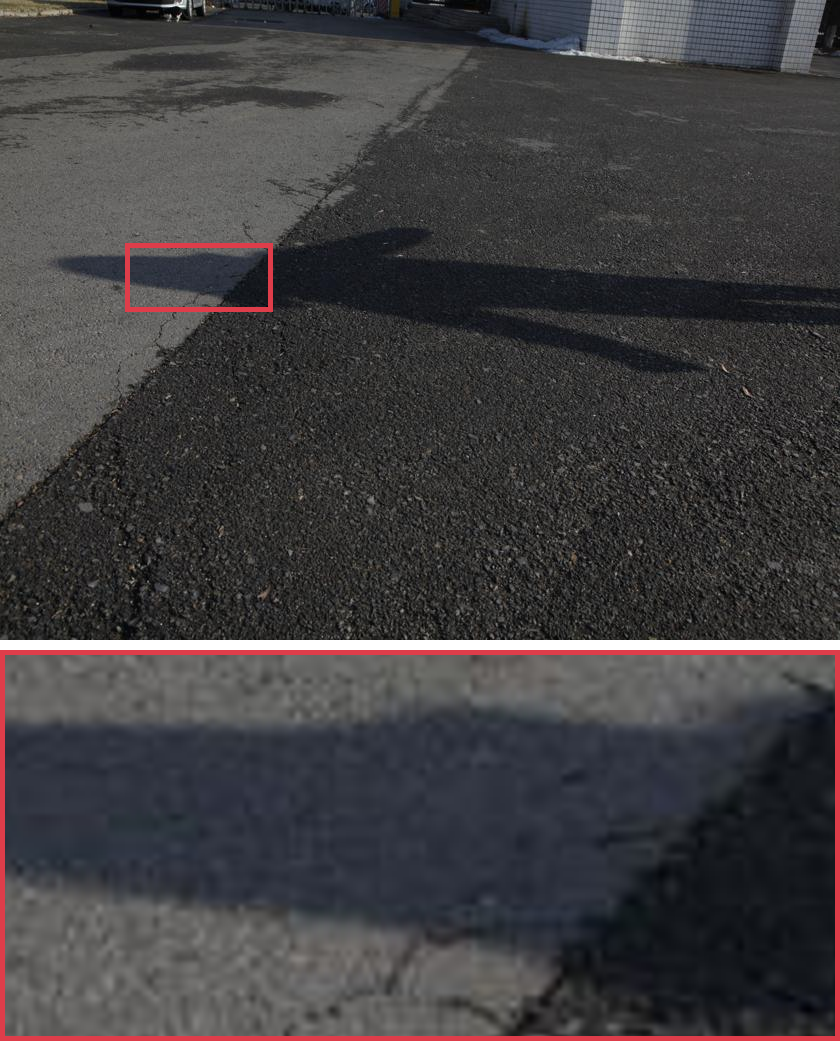}
    \includegraphics[width=.118\linewidth]{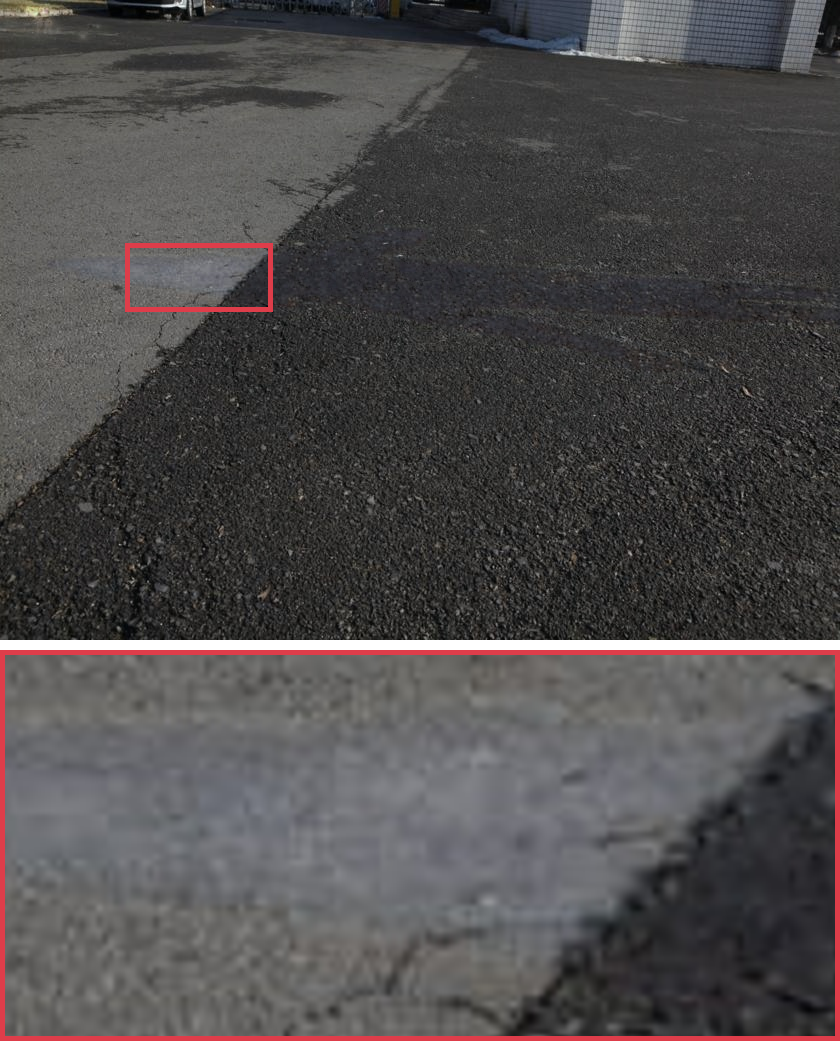}
    \includegraphics[width=.118\linewidth]{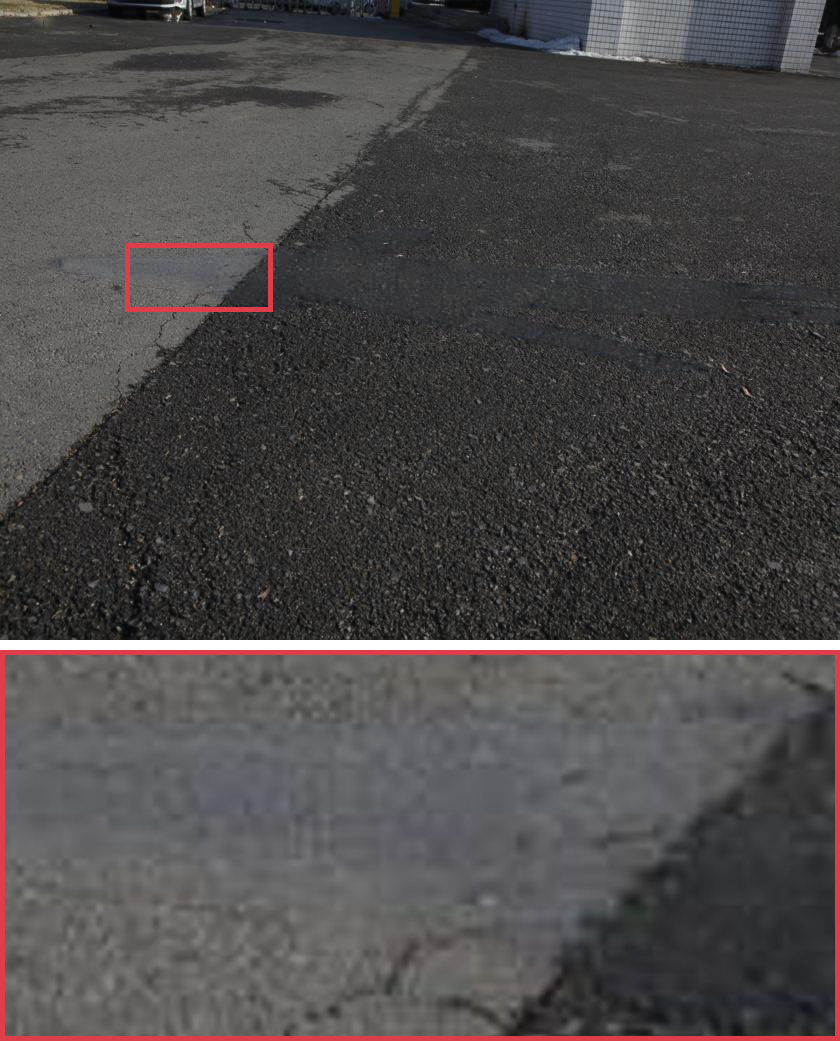}
    \includegraphics[width=.118\linewidth]{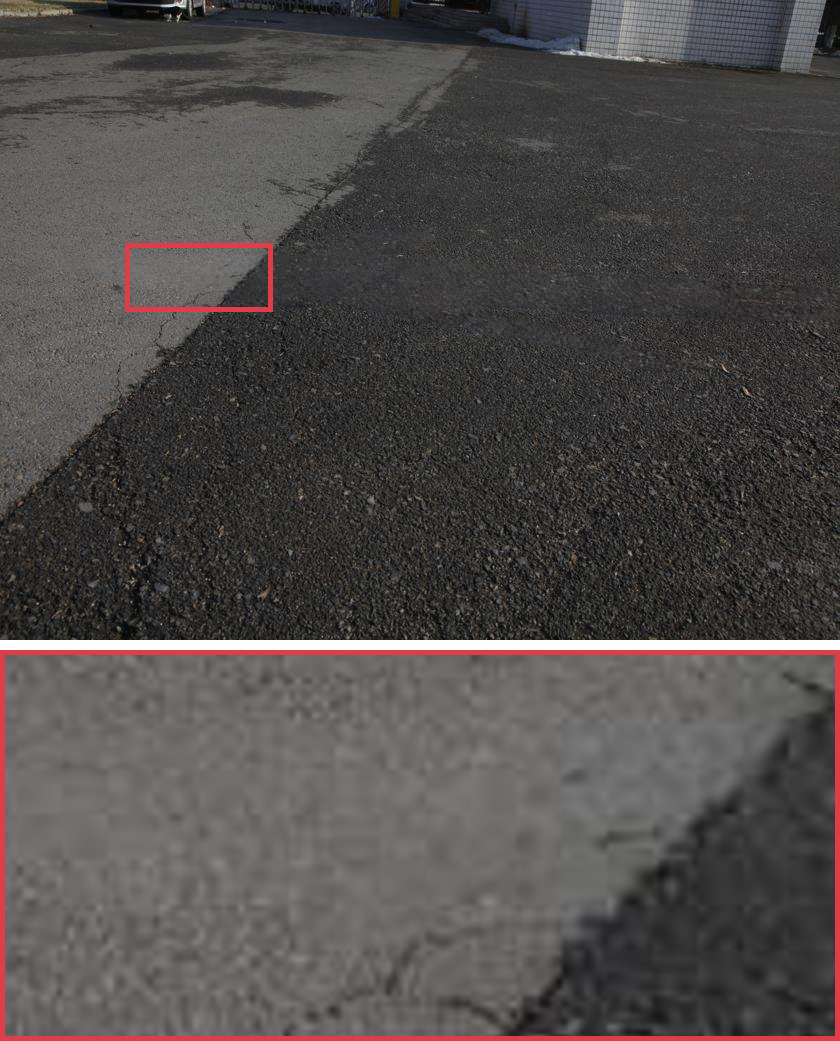}
    \includegraphics[width=.118\linewidth]{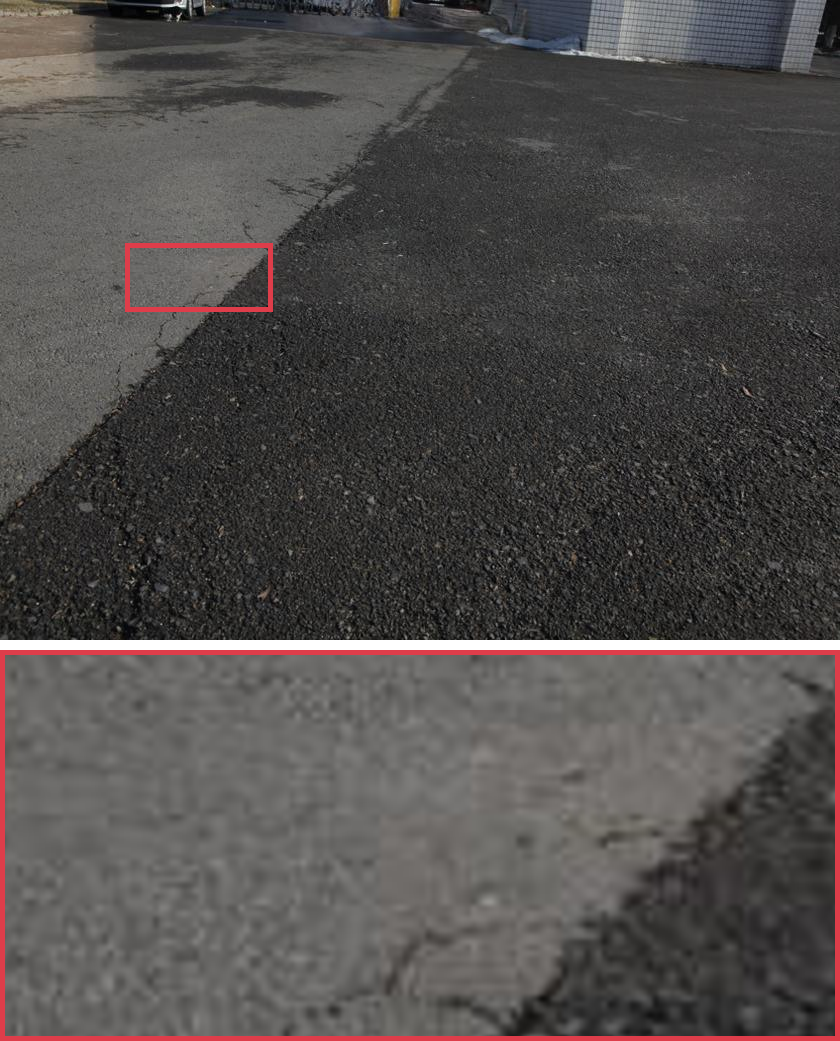}
    \includegraphics[width=.118\linewidth]{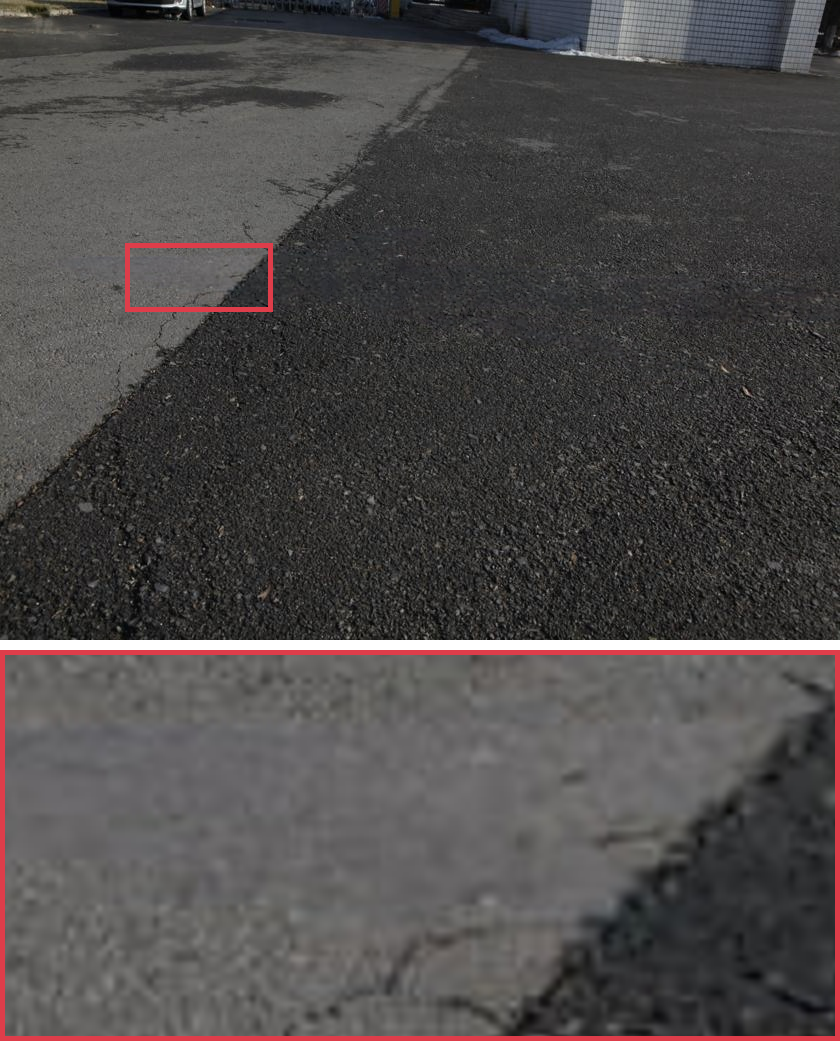}
    \includegraphics[width=.118\linewidth]{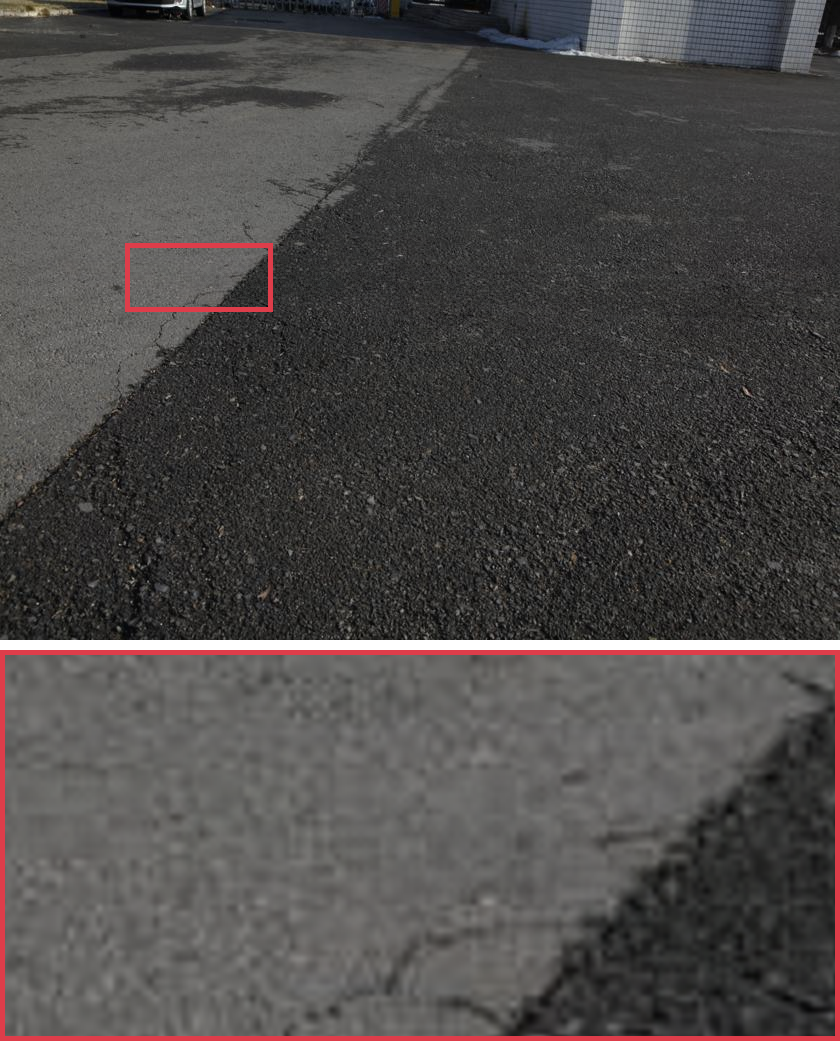}
    \includegraphics[width=.118\linewidth]{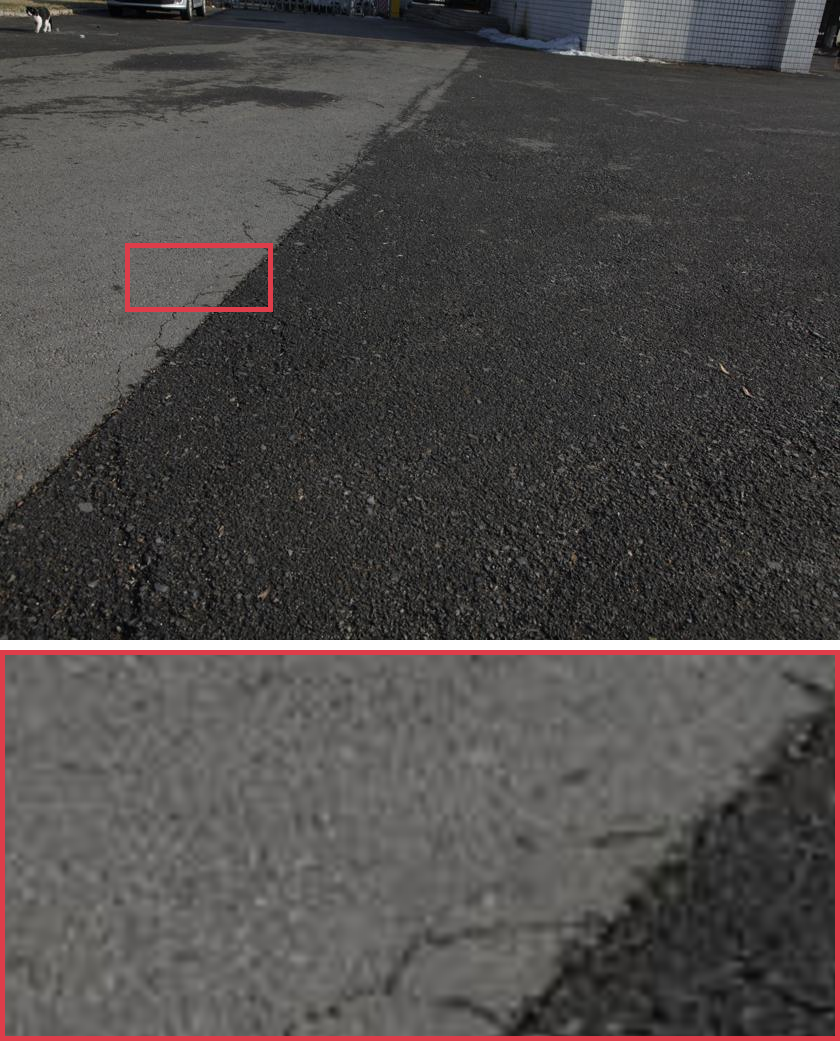}

    \begin{subfigure}{0.118\linewidth}
        \centering
        \subcaption{Input}
    \end{subfigure}
    \begin{subfigure}{0.118\linewidth}
        \centering
        \subcaption{BMNet~\cite{ZhuHFZSZ22}}
    \end{subfigure}
    \begin{subfigure}{0.118\linewidth}
        \centering
        \subcaption{SG~\cite{WanYWWLW22}}
    \end{subfigure}
        \begin{subfigure}{0.118\linewidth}
        \centering
        \subcaption{DMTN~\cite{LiuWFLQT23}}
    \end{subfigure}
    \begin{subfigure}{0.118\linewidth}
        \centering
        \subcaption{DeS3~\cite{Jin0YYT24}}
    \end{subfigure}
    \begin{subfigure}{0.118\linewidth}
        \centering
        \subcaption{HF~\cite{0002FZL00Z24}}
    \end{subfigure}
    \begin{subfigure}{0.118\linewidth}
        \centering
        \subcaption{Ours}
    \end{subfigure}
    \begin{subfigure}{0.118\linewidth}
        \centering
        \subcaption{GT}
    \end{subfigure}
    \vspace{-5pt}\caption{Visual comparison on a sample from the SRD dataset. }
    \label{fig:exp.srd}
        \vspace{-10pt}
\end{figure*}

\subsection{Experimental Setup}
\noindent\textbf{Implementation details.}
Our ShadowHack is implemented in Python with the PyTorch framework and is trained using NVIDIA RTX4090 GPUs. We utilize the AdamW optimizer~\cite{LoshchilovH19} with momentum set to (0.9, 0.999) and a weight decay of $10^{-2}$. The initial learning rate is set to $2\times10^{-4}$ and decays to $10^{-6}$ at the schedule's end in a cosine annealing manner. During training, we randomly crop the training data into patches of size 384$\times$384, and apply data augmentation techniques, including rotation, flipping, mixup, and color jittering as in previous works~\cite{GuoHLCW23, liu2024regional}. We use L1 and VGG loss terms as employed in previous works~\cite{LiuKXLWL24, liu2024regional}.
For the base network of each stage, we use a four-layer encoder-decoder structure (from L1 to L4), consisting of three downsampling and three upsampling operations. The dimension of the first layer is 32. In the LRNet, our ROA modules are integrated into L3 and L4, with an overlapping ratio set to 0.5 and a rectified mechanism $\lambda_0$ value of 0.7. For the CRNet, we bring a color encoder on the top of the U-shape network. The color encoder is an FC-MAE pretrained atto variant of ConvNext-v2, with merely 2M parameters.

\begin{table}[t]
\centering
    \begin{tabular}{l|ccc}
    \toprule
        Method & Params (M) & PSNR$\uparrow$ & RMSE$\downarrow$ \\
        \midrule
        ShadowFormer~\cite{GuoHLCW23} & 11.4 & 32.90 & 4.04 \\
        ShadowDiffusion~\cite{GuoWYHWPW23} & 55.2 & 34.73 & 3.63\\
        Li {\it et al.}~\cite{Li0A00T023} & 23.9 & 33.17 & 3.83 \\
        DeS3~\cite{Jin0YYT24} & 113.7 & 34.11 & 3.72 \\
        RASM~\cite{liu2024regional} & 5.2 & 34.46 & 3.37 \\
        Homoformer~\cite{0002FZL00Z24} & 17.8 & 35.37 & 3.33 \\
        Ours & 23.3 & 35.94 & 2.90 \\
    \bottomrule
    \end{tabular}
    \vspace{-5pt} \caption{Comparisons of model size and performance on the SRD dataset with state-of-the-art methods.}
    \vspace{-15pt}
\label{tab:complexity}
\end{table}

\noindent\textbf{Datasets.}
We conduct our experiments on the following benchmark datasets for shadow removal: 1) Adjusted ISTD (ISTD+) dataset~\cite{LeS19}, which refines the ISTD dataset~\cite{WangL018} by correcting the illumination discrepancies between the ground-truth and input images, includes 1330 training and 540 testing triplets (shadow image, shadow-free image, and shadow mask); 2) SRD dataset~\cite{QuTHTL17} consists of 2680 training and 408 testing pairs. We adopt the predicted masks provided by DHAN~\cite{CunPS20} for training and testing following previous works~\cite{ZhuHFZSZ22, 0002FZL00Z24, liu2024regional}; 3) UIUC dataset~\cite{GuoDH13} provides 76 images with shadow mask for testing; and 4) UCF dataset~\cite{ZhuSMT10} provided by Jin {\it et al.}~\cite{Jin0YYT24} contains shadow images without shadow-free references.

\noindent\textbf{Evaluation metrics.}
To evaluate the performance of our proposed framework, we utilize the Peak Signal-to-Noise Ratio (PSNR), Structural Similarity Index Measure (SSIM), and Root Mean Square Error (RMSE). PSNR and SSIM are both calculated in the RGB space, with higher values indicating better results. For RMSE, which is calculated in the LAB color space, the lower values indicate better results. To be consistent with previous works~\cite{ZhuHFZSZ22, 0002FZL00Z24, liu2024regional}, we downsample the images to 256$\times$256 during evaluation.

\begin{table}[t]
\centering

    \begin{tabular}{l|ccc}
    \toprule
        Metric & RGB Input &Retinex-based &Ours \\
        \midrule
        PSNR$\uparrow$ & 36.16 & 35.8 &  \textbf{36.31} \\
        RMSE$\downarrow$ & 2.54 & 2.63 &\textbf{2.46}\\
    \bottomrule
    \end{tabular}
\vspace{-5pt}\caption{Ablation study of the proposed decoupling framework on the ISTD+ dataset.}
\label{tab:Ablation_1}
\vspace{-20pt}\end{table}

\begin{figure}[!t]
    \centering
    
    \includegraphics[width=.24\linewidth, height=.24\linewidth]{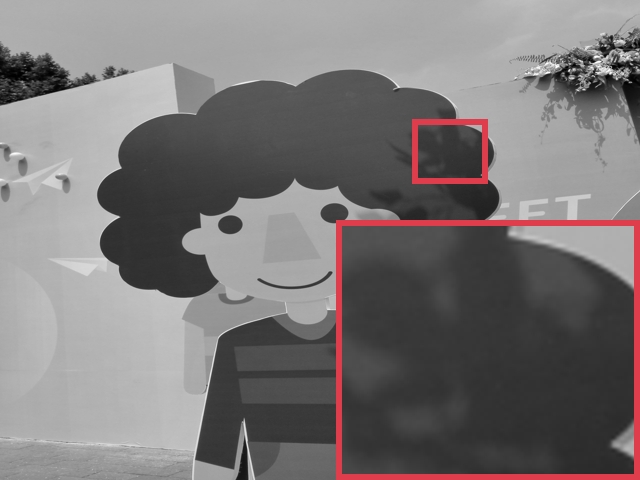}
    \includegraphics[width=.24\linewidth, height=.24\linewidth]{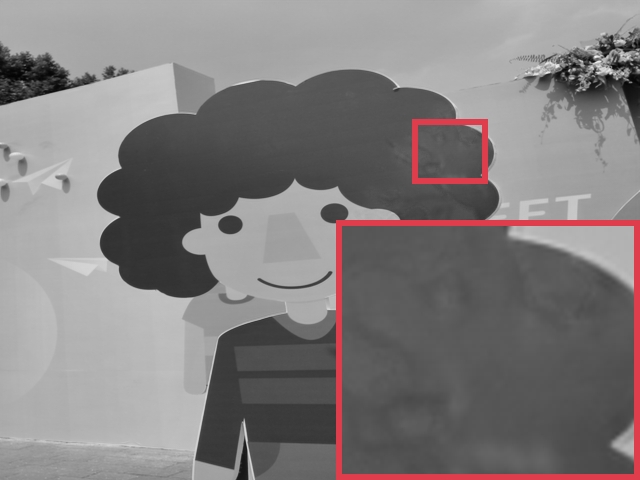}
    \includegraphics[width=.24\linewidth, height=.24\linewidth]{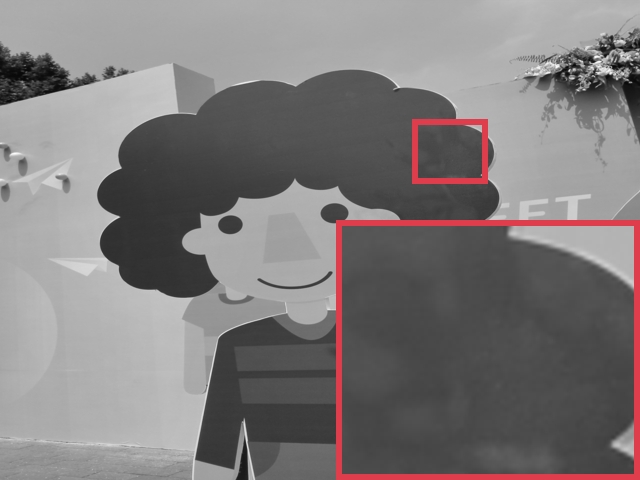}
    \includegraphics[width=.24\linewidth, height=.24\linewidth]{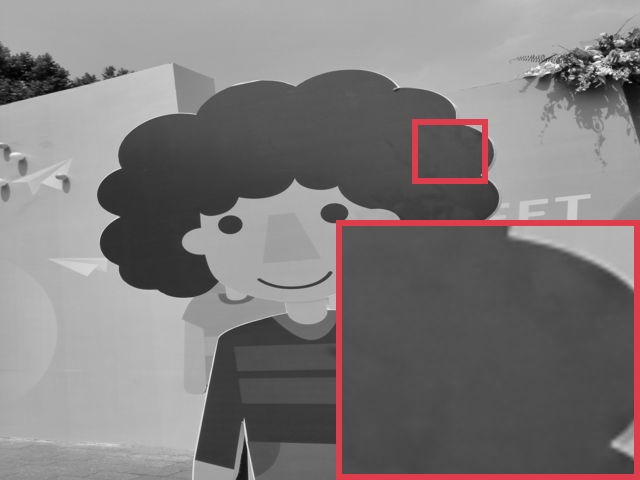}
    
    \begin{subfigure}{0.24\linewidth}
        \centering
        \subcaption{Input}
    \end{subfigure}
    \begin{subfigure}{0.24\linewidth}
        \centering
        \subcaption{w/o outreach}
    \end{subfigure}
    \begin{subfigure}{0.24\linewidth}
        \centering
        \subcaption{w/o rectify}
    \end{subfigure}
    \begin{subfigure}{0.24\linewidth}
        \centering
        \subcaption{Ours}
    \end{subfigure}

    \vspace{-5pt}\caption{Ablation study on rectified outreach attention. The results are displayed in the luminance space.}
    \label{fig:ablation_ROA}
\end{figure}

\begin{figure}[t]
    \centering
    
    \includegraphics[width=.19\linewidth, height=.15\linewidth]{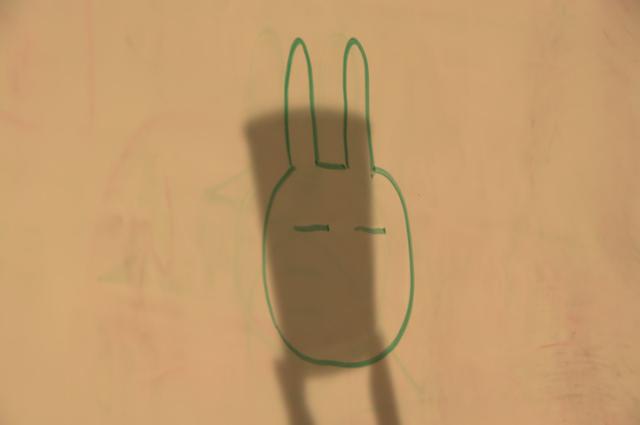}
    \includegraphics[width=.19\linewidth, height=.15\linewidth]{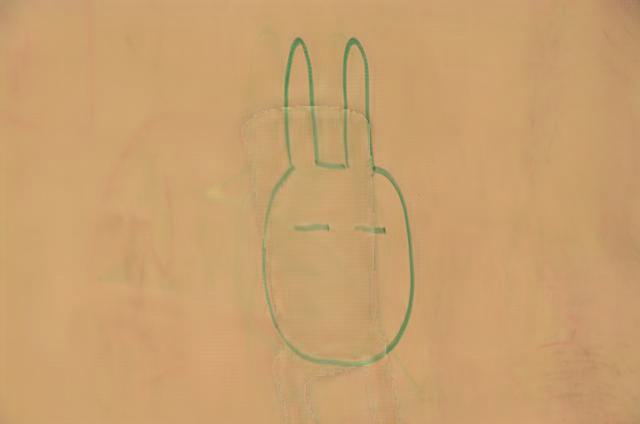}
    \includegraphics[width=.19\linewidth, height=.15\linewidth]{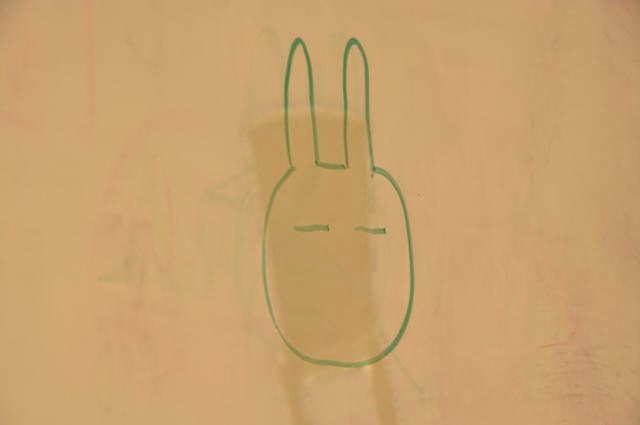}
    \includegraphics[width=.19\linewidth, height=.15\linewidth]{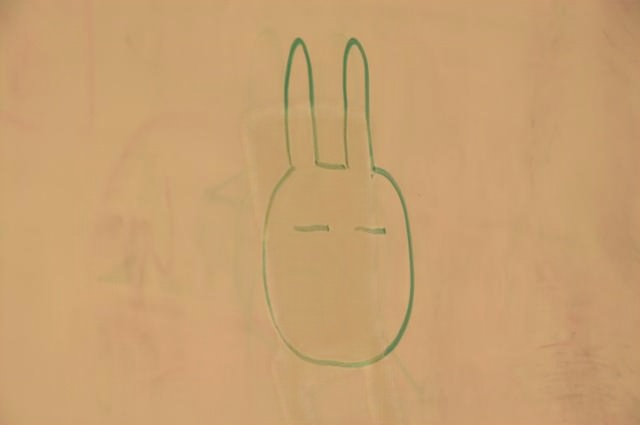}
    \includegraphics[width=.19\linewidth, height=.15\linewidth]{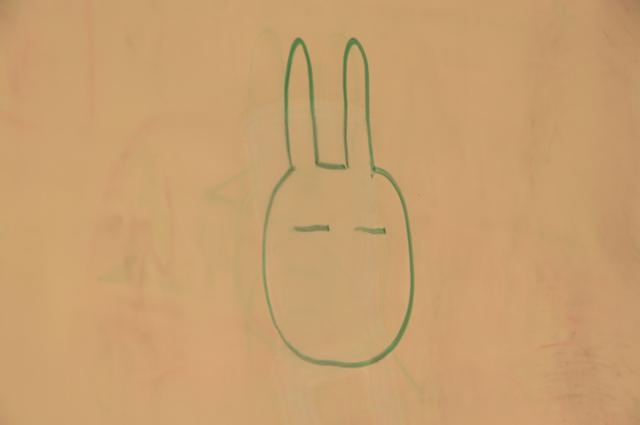}

    \includegraphics[width=.19\linewidth, height=.15\linewidth]{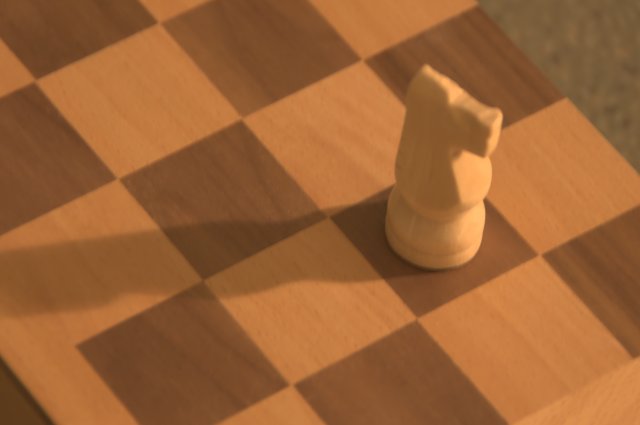}
    \includegraphics[width=.19\linewidth, height=.15\linewidth]{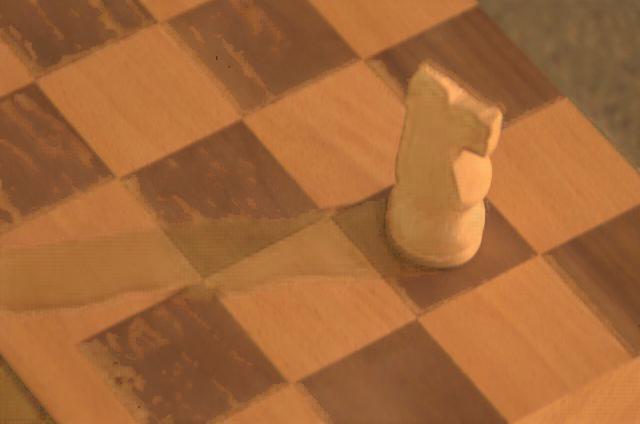}
    \includegraphics[width=.19\linewidth, height=.15\linewidth]{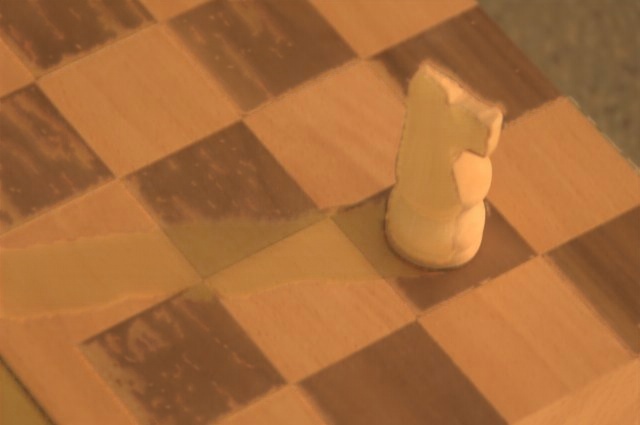}
    \includegraphics[width=.19\linewidth, height=.15\linewidth]{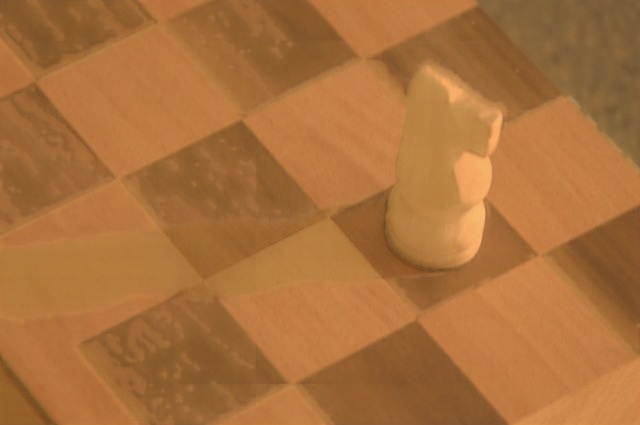}
    \includegraphics[width=.19\linewidth, height=.15\linewidth]{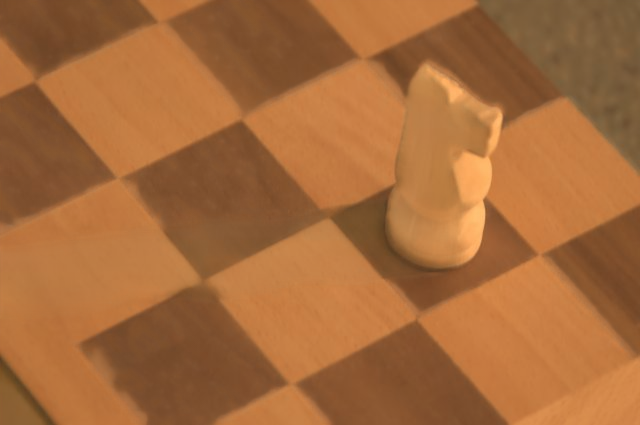}

    \includegraphics[width=.19\linewidth, height=.15\linewidth]{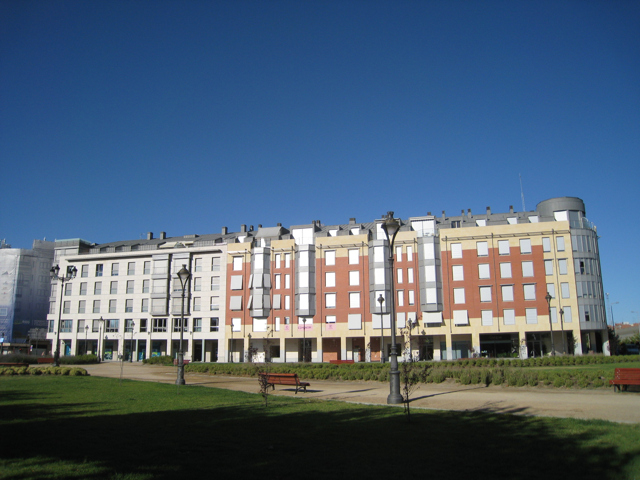}
    \includegraphics[width=.19\linewidth, height=.15\linewidth]{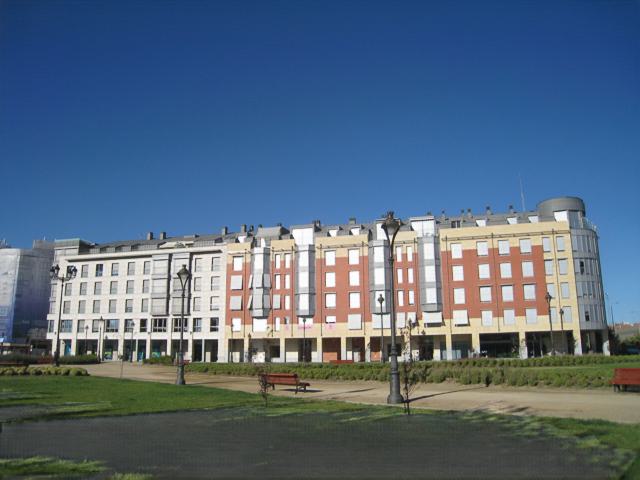}
    \includegraphics[width=.19\linewidth, height=.15\linewidth]{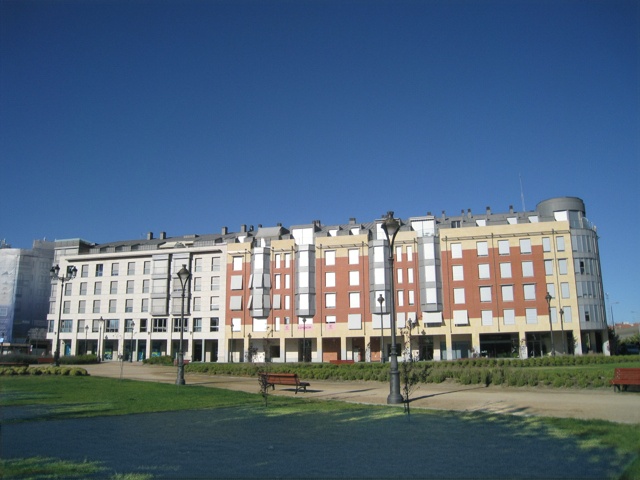}
    \includegraphics[width=.19\linewidth, height=.15\linewidth]{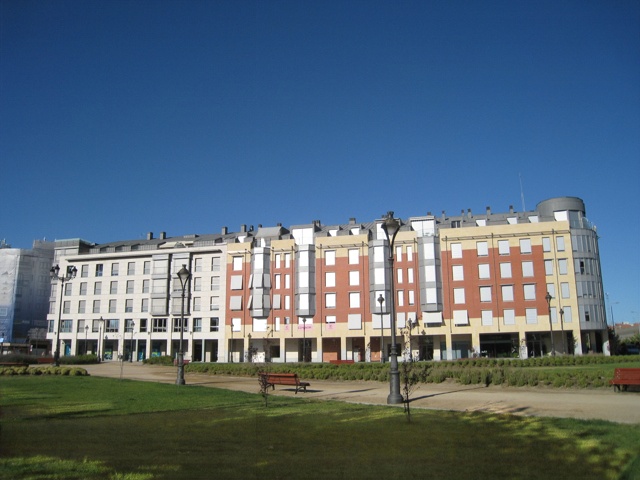}
    \includegraphics[width=.19\linewidth, height=.15\linewidth]{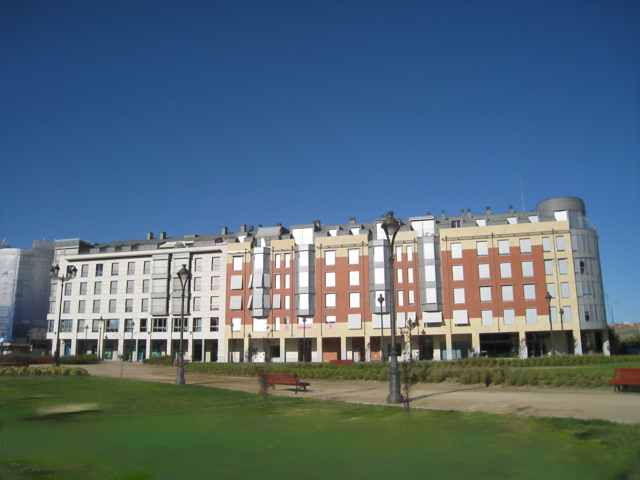}

    \includegraphics[width=.19\linewidth, height=.15\linewidth]{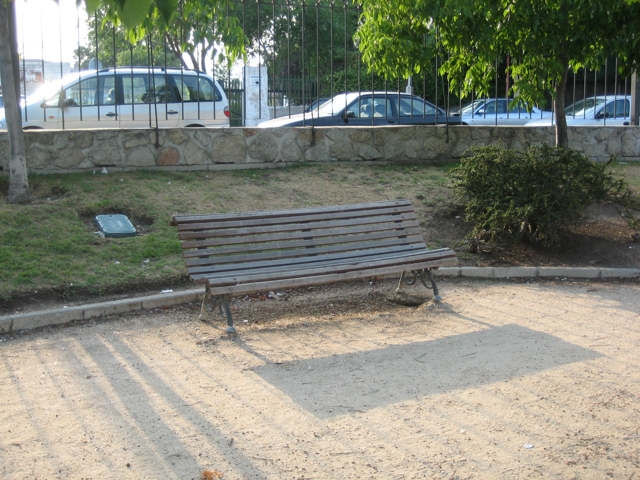}
    \includegraphics[width=.19\linewidth, height=.15\linewidth]{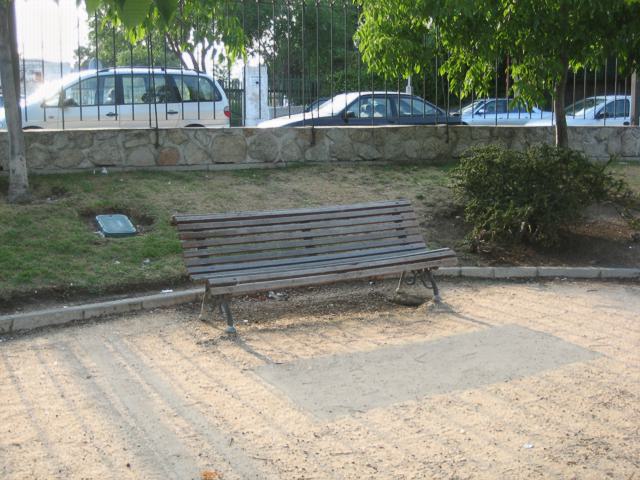}
    \includegraphics[width=.19\linewidth, height=.15\linewidth]{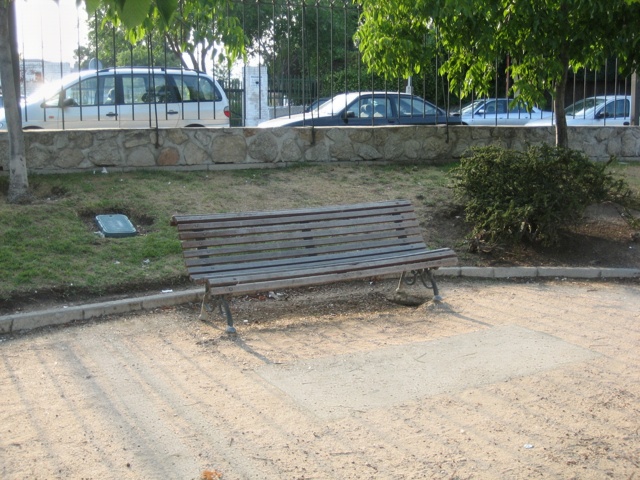}
    \includegraphics[width=.19\linewidth, height=.15\linewidth]{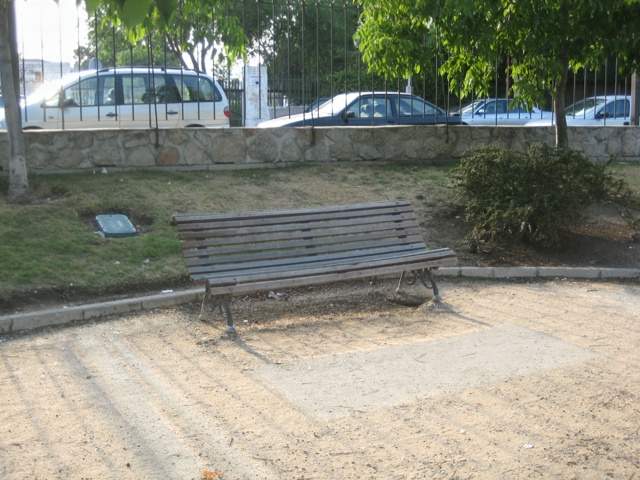}
    \includegraphics[width=.19\linewidth, height=.15\linewidth]{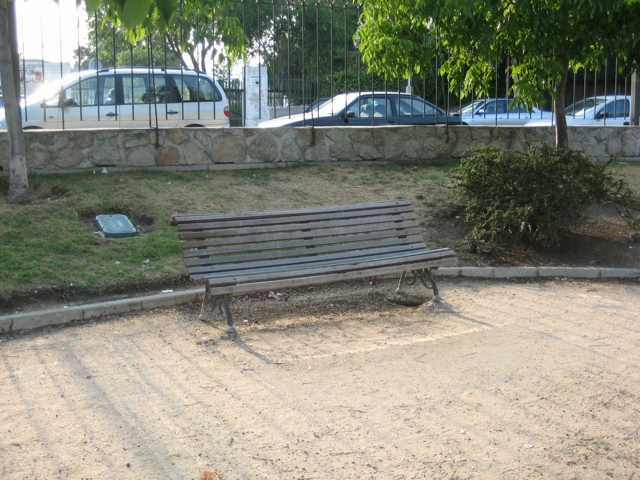}

    \includegraphics[width=.19\linewidth, height=.15\linewidth]{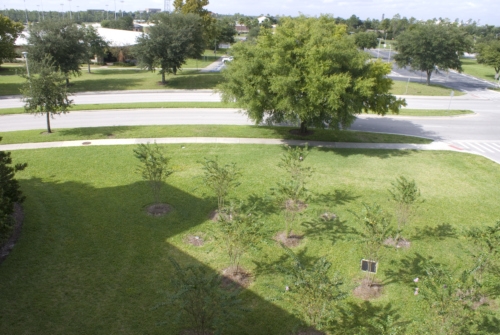}
    \includegraphics[width=.19\linewidth, height=.15\linewidth]{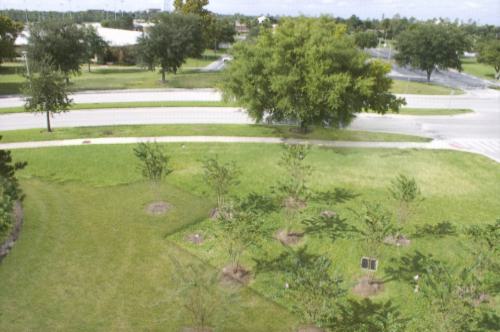}
    \includegraphics[width=.19\linewidth, height=.15\linewidth]{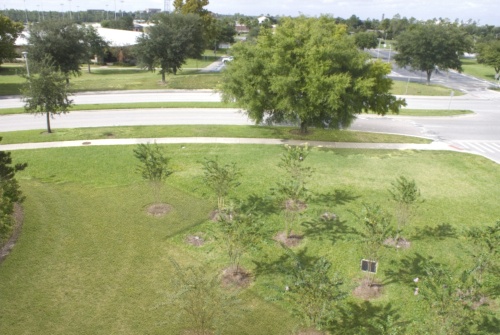}
    \includegraphics[width=.19\linewidth, height=.15\linewidth]{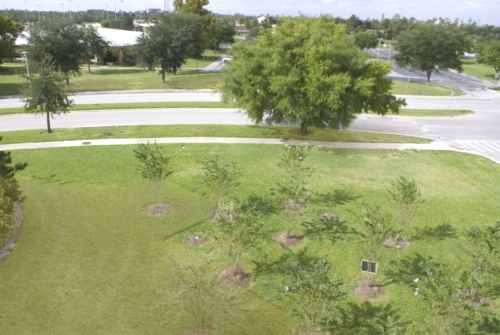}
    \includegraphics[width=.19\linewidth, height=.15\linewidth]{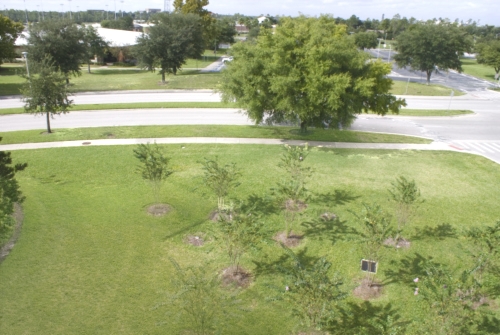}

    \includegraphics[width=.19\linewidth, height=.15\linewidth]{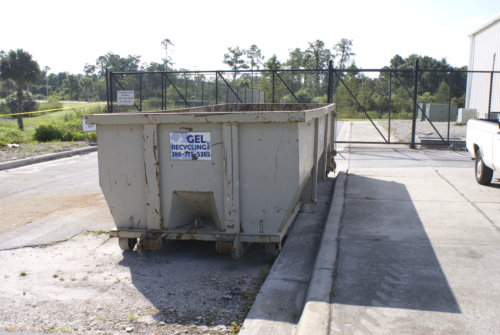}
    \includegraphics[width=.19\linewidth, height=.15\linewidth]{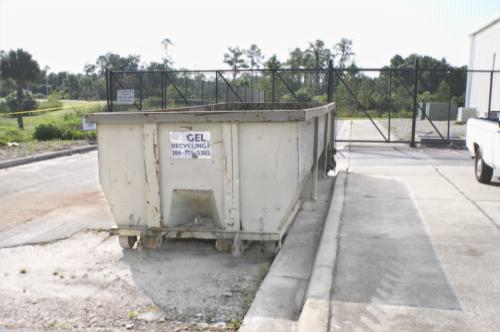}
    \includegraphics[width=.19\linewidth, height=.15\linewidth]{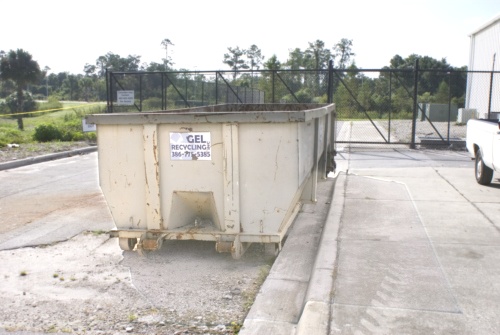}
    \includegraphics[width=.19\linewidth, height=.15\linewidth]{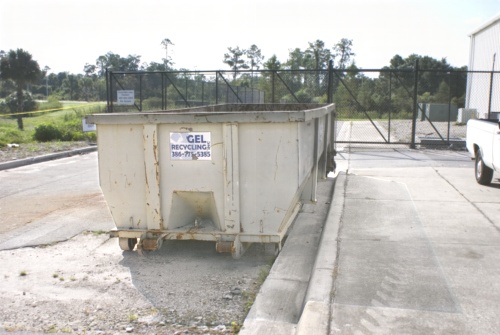}
    \includegraphics[width=.19\linewidth, height=.15\linewidth]{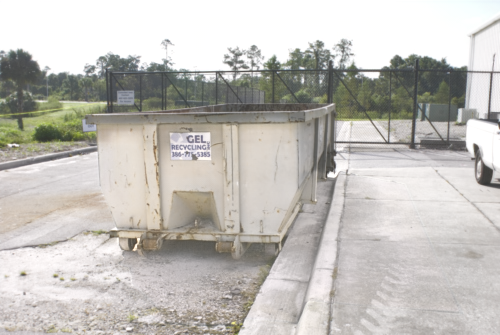}

    \begin{subfigure}{0.19\linewidth}
        \centering
        \subcaption{Input}
    \end{subfigure}
    \begin{subfigure}{0.19\linewidth}
        \centering
        \subcaption{SG~\cite{WanYWWLW22}}
    \end{subfigure}
    \begin{subfigure}{0.19\linewidth}
        \centering
        \subcaption{SF~\cite{GuoHLCW23}}
    \end{subfigure}
    \begin{subfigure}{0.19\linewidth}
        \centering
        \subcaption{HF~\cite{0002FZL00Z24}}
    \end{subfigure}
    \begin{subfigure}{0.19\linewidth}
        \centering
        \subcaption{Ours}
    \end{subfigure}
    \vspace{-5pt}\caption{Visual comparisons on the UIUC and UCF datasets.}
    \label{fig:exp.generalization}
    \vspace{-10pt}
\end{figure}

\subsection{Comparison with Prior Arts}
We compare our method with the state-of-the-art shadow removal approaches on benchmark datasets, including DHAN~\cite{CunPS20}, G2R~\cite{LiuYWWM021}, BMNet \cite{ZhuHFZSZ22}, SG-ShadowNet~\cite{WanYWWLW22}, ShadowFormer~\cite{GuoHLCW23}, ShadowDiffusion \cite{GuoWYHWPW23}, Li {\it et al.}~\cite{Li0A00T023}, Liu {\it et al.}~\cite{LiuKXLWL24}, DeS3~\cite{Jin0YYT24}, Homoformer~\cite{0002FZL00Z24} and RASM~\cite{liu2024regional}, as shown in Tab.~\ref{tab:aistd_res}. For a fair comparison, we use results published by the authors. Regarding missing data, we utilize images or checkpoints provided by the authors and evaluate them via the same metrics. 

Our proposed method achieves the highest scores in most metrics on the ISTD+ dataset compared to other state-of-the-art methods. Specifically, our model attains a PSNR of 36.31 for all image regions, surpassing the second-highest result of 36.16 by 0.15. The high PSNR highlights our model's ability to recover texture details effectively, demonstrating superior reconstruction quality. Additionally, the RMSE value of 2.48 is the lowest, which indicates that our recoloring module plays a crucial role in accurately restoring color information and reducing errors across the entire image. 
From the magnified areas in Figs.~\ref{fig:exp.istd} and ~\ref{fig:exp.srd}, it can be observed that our method not only effectively corrects color bias but also provides superior edge correction compared to other methods. The second part of Tab.~\ref{tab:aistd_res} presents a comparison of our results on the SRD dataset against other state-of-the-art methods. It can be observed that our method achieves a significant improvement in RMSE (-0.43), further demonstrating the substantial benefit of the decoupling approach for color recovery.

Additionally, we provide the total size of our models compared with state-of-the-art models in Tab~\ref{tab:complexity}, which demonstrates that while our model is only slightly larger than the leading state-of-the-art method, Homoformer~\cite{0002FZL00Z24}, it significantly outperforms it in both PSNR and RMSE.

\subsection{Ablation Study}

\noindent\textbf{Effect of the decoupling framework.}
To validate the effectiveness of our decoupling framework, we fine-tune our baseline model structure to adapt to RGB input/output for end-to-end shadow removal and compare its performance with our decoupling framework. As shown in Tab.~\ref{tab:Ablation_1}, our decoupling approach significantly improves both PSNR and RMSE. 
Moreover, based on Retinex decomposition, we separately recover L and R, and compare it with our decoupling method. In detail, we obtain illumination maps from a U-Net~\cite{RonnebergerFB15} and their reflectance maps via division as \cite{GuoH23} did. Subsequently, we relight the illumination and remove the color bias in reflectance with the same architecture as LRNet. 
This improvement indicates that this strategy benefits texture recovery and color correction.

\begin{table}[!t]
\centering
\resizebox{\linewidth}{!}{
    \begin{tabular}{l|ccc}
    \toprule
        \multirow{2}{*}{Method} & Shadow & Non-Shadow & All \\
         & PSNR $\uparrow$ & PSNR $\uparrow$ & PSNR $\uparrow$ \\
        \midrule
        w/o outreach & 39.17 & 40.77 & 35.93 \\
        w/o dilation & 39.74 & 41.09 & 36.42 \\
        w/o rectify & 39.96 & 41.27 & 36.60 \\
        $F_t \& F_t$ & 40.02 & 41.18 & 36.58 \\
        $F_c$ & 40.17 & 41.29 & 36.76 \\
        $[F_t;F_c]\& [F_t;F_c]$ & 40.35 & 41.37 & 36.87 \\
        $[F_t;F_c] \& [F_c]$ & \textbf{40.36} & \textbf{41.40} & \textbf{36.90} \\
    \bottomrule
    \end{tabular}}
\vspace{-5pt}\caption{Ablation study of the proposed rectified outreach attention module on the SRD dataset. 
We calculate PSNRs on the luminance.}
\label{tab:Ablation_2}
\vspace{-10pt}\end{table}

\begin{table}[t]
\centering
\resizebox{\linewidth}{!}{
    \begin{tabular}{l|ccc}
    \toprule
        \multirow{2}{*}{Method} & Shadow & Non-Shadow & All \\
         & RMSE $\downarrow$ & RMSE $\downarrow$ & RMSE $\downarrow$ \\
        \midrule
        w/o cross-att. & 5.78 & 2.64 & 3.39 \\
        w/o checkpt ensem. & 5.01 & 2.59 & 3.28 \\
        Ours & \textbf{4.18} & \textbf{2.38} & \textbf{2.90} \\
    \bottomrule
    \end{tabular}}
\vspace{-5pt}\caption{Ablation study of the CRNet on the SRD dataset.}
\label{tab:Ablation_3}\vspace{-10pt}
\end{table}

\begin{figure*}[t]
    \centering
    
    \includegraphics[width=.19\linewidth]{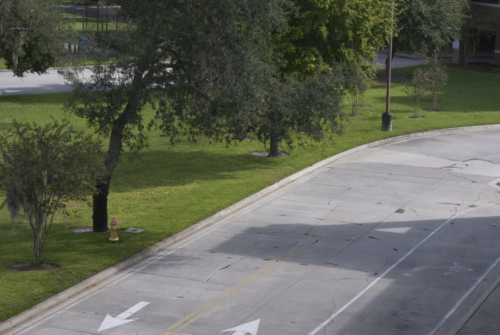}
    \includegraphics[width=.19\linewidth]{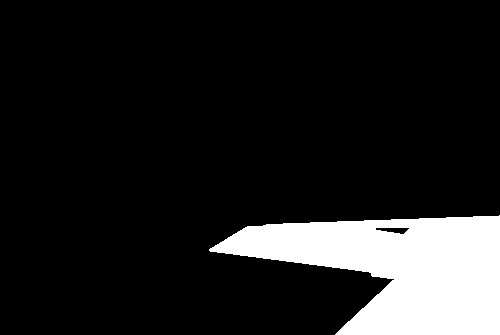}
    \includegraphics[width=.19\linewidth]{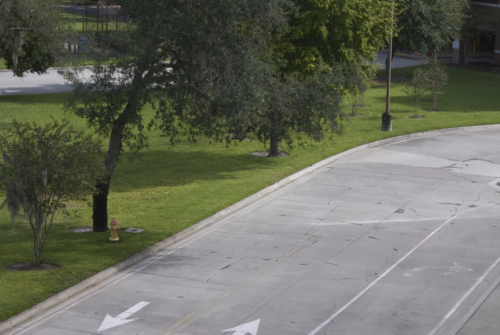}
    \includegraphics[width=.19\linewidth]{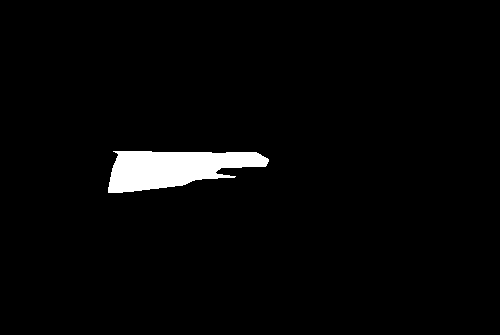}
    \includegraphics[width=.19\linewidth]{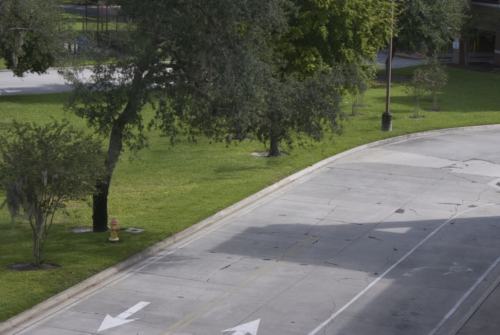}

    \includegraphics[width=.19\linewidth]{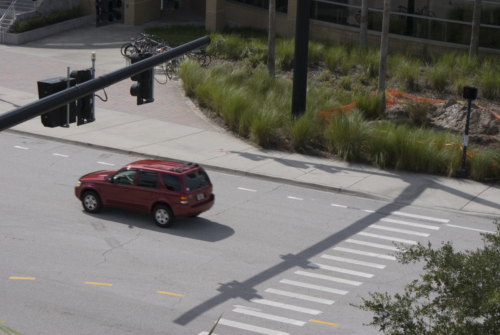}
    \includegraphics[width=.19\linewidth]{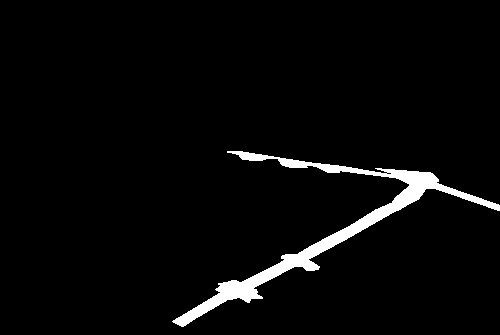}
    \includegraphics[width=.19\linewidth]{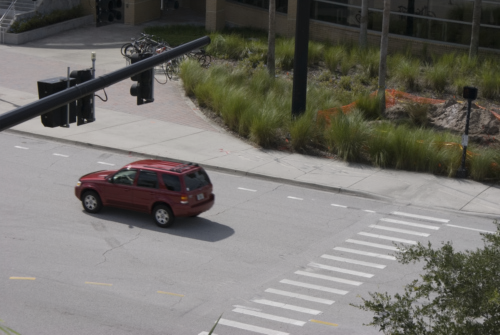}
    \includegraphics[width=.19\linewidth]{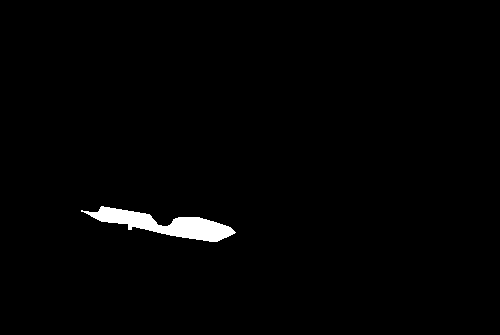}
    \includegraphics[width=.19\linewidth]{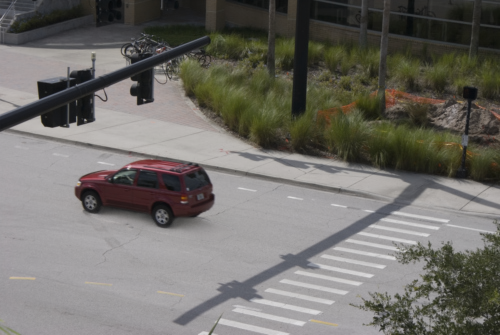}

    \includegraphics[width=.19\linewidth]{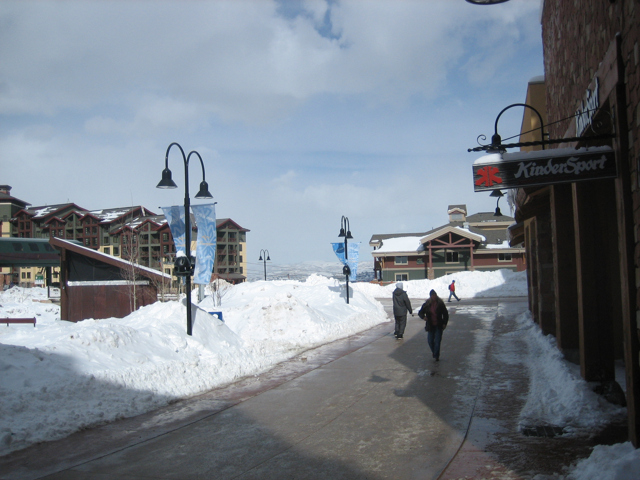}
    \includegraphics[width=.19\linewidth]{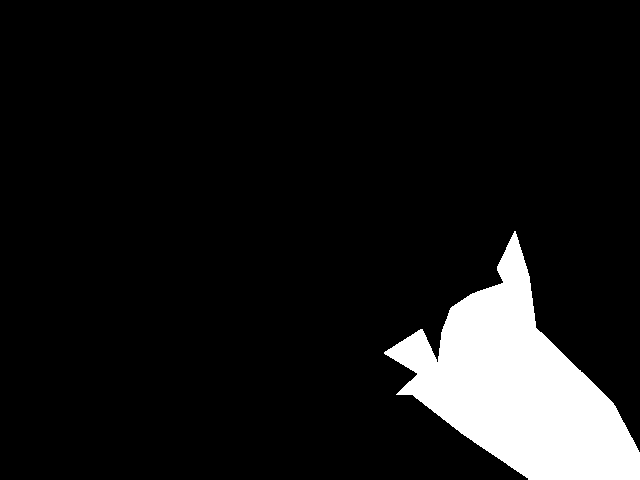}
    \includegraphics[width=.19\linewidth]{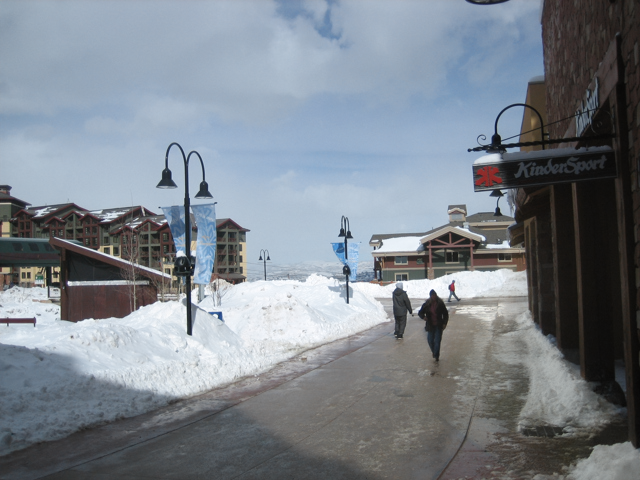}
    \includegraphics[width=.19\linewidth]{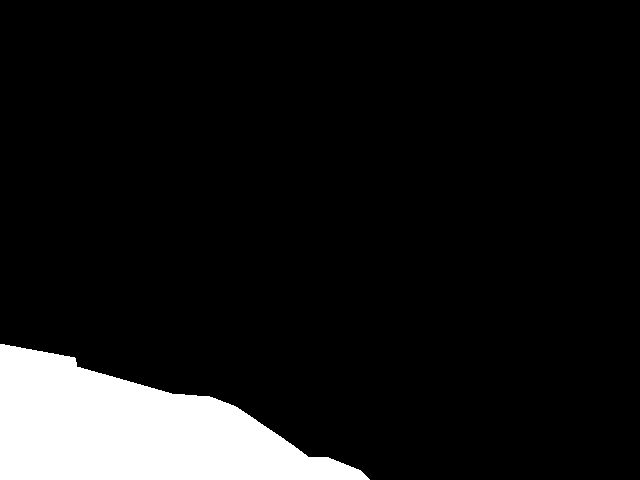}
    \includegraphics[width=.19\linewidth]{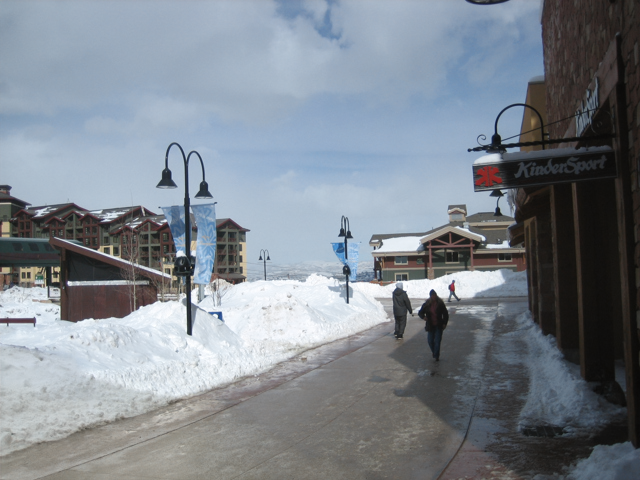}

    \begin{subfigure}{0.19\linewidth}
        \centering
        \subcaption{Input}
    \end{subfigure}
    \begin{subfigure}{0.19\linewidth}
        \centering
        \subcaption{Mask1}
    \end{subfigure}
    \begin{subfigure}{0.19\linewidth}
        \centering
        \subcaption{Result1}
    \end{subfigure}
    \begin{subfigure}{0.19\linewidth}
        \centering
        \subcaption{Mask2}
    \end{subfigure}
    \begin{subfigure}{0.19\linewidth}
        \centering
        \subcaption{Result2}
    \end{subfigure}
    
    \vspace{-5pt}
    \caption{Visual results of selective shadow masks on the UCF dataset. }
    \label{fig:suppl.flexible}
        \vspace{-10pt}
\end{figure*}

\begin{figure}[!t]
    \centering
    \includegraphics[width=0.99\linewidth]{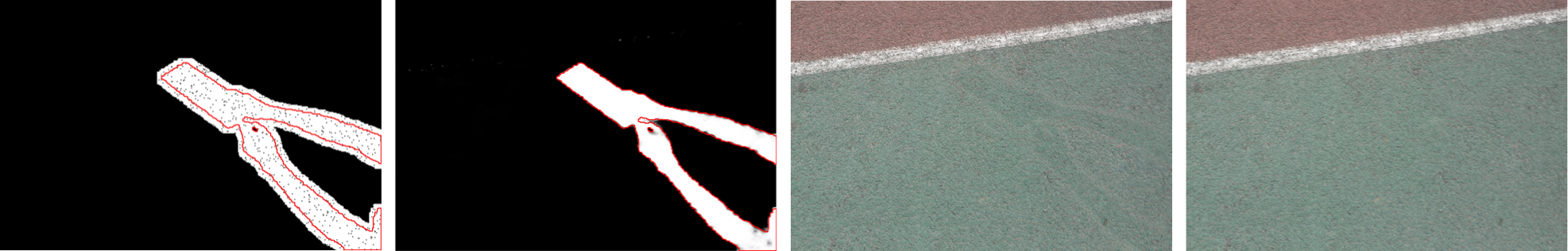}

    \begin{subfigure}{0.24\linewidth}
        \centering
        \subcaption{dirty mask}
    \end{subfigure}
    \begin{subfigure}{0.24\linewidth}
        \centering
        \subcaption{refined mask}
    \end{subfigure}
    \begin{subfigure}{0.24\linewidth}
        \centering
        \subcaption{w/o refinement}
    \end{subfigure}
    \begin{subfigure}{0.24\linewidth}
        \centering
        \subcaption{w refinement}
    \end{subfigure}

    \vspace{-5pt}\caption{Robustness analysis under corrupted mask. The accurate mask is highlighted in red.}
    \label{fig:mask_refine}
        \vspace{-15pt}
    \end{figure}

\noindent\textbf{Effect of rectified outreach attention module.}
We demonstrate the effectiveness of the rectified outreach attention layer in Tab.~\ref{tab:Ablation_2}, by conducting experiments on outreach windows and rectification mechanisms. Furthermore, extra experiments on different feature combinations of rectified approaches are conducted to explore more suitable configurations. $[X] \And [Y]$ in Tab.~\ref{tab:Ablation_2} means performing a differential calculation for rectification between attention $X$ and $Y$. 
Clearly, using outreach windows allows shadow regions to better reference surrounding areas, resulting in significant performance improvements. Incorporating color information enhances luminance recovery, while rectification refines the attention map for greater precision. Additionally, we find that $[F_t;F_c] \& [F_c]$ slightly outperforms $[F_t;F_c] \& [F_t;F_c]$, and thus we ultimately choose $[F_t;F_c] \& [F_c]$. Fig.~\ref{fig:ablation_ROA} visualizes the effectiveness of rectification, which has a more precise result with less edge remaining.

\noindent\textbf{Effect of the color restoration network.} To validate the effectiveness of the key component and design in the CRNet, we replace the cross-attention-based color feature injection with concatenation along the feature channel. As demonstrated in Tab.~\ref{tab:Ablation_3}, if cross-attention is replaced, the accuracy has dropped by a large margin, indicating the effectiveness of the cross-attention. Furthermore, the model suffers from a misalignment between training and testing luminance results if the checkpoint ensemble is removed.

\subsection{Robustness Analysis}
\textbf{Generalization to other datasets.}
To demonstrate the generalization capability of ShadowHack, we evaluate it on the UIUC~\cite{GuoDH13} and UCF~\cite{ZhuSMT10} datasets with our ISTD+ pretrained model. 
The results are presented in Fig.~\ref{fig:exp.generalization}, where ShadowHack achieving the cleanest shadow removal with minimal color bias.

\noindent\textbf{Robustness to inaccurate masks.} 
To get rid of the performance degradation brought by inaccurate masks, we propose a mask refine network, which corrects the mask for better generalization. 
The mask refine network simply adopts U-Net~\cite{RonnebergerFB15}. Considering the complexity of shadow characteristics, we inject ConvNext-v2 atto model~\cite{0003MWFDX22} as our CRNet does. The extracted features of RGB inputs are fed into the U-Net decoder for a more precise mask.
After our remediation, the result is recovered from inaccurate enlightening as the example in Fig.~\ref{fig:mask_refine}. 

\noindent\textbf{Robustness to user-specified shadow masks.}
There are situations where it is necessary to remove specific regions of a shadow selectively. To illustrate the capability of our model in addressing this requirement, we evaluate its performance using flexible shadow masks. As shown in Fig.~\ref{fig:suppl.flexible}, the results highlight the model's high flexibility in removing shadows according to user-defined mask specifications.

%% file: sec/5_conclusion.tex
\section{Conclusion}
This work proposed a novel framework, \textit{ShadowHack}, which decouples images into luminance and color components, enabling targeted processing for more effective shadow removal. This approach can well address limitations in existing end-to-end models by enhancing the use of color information to improve both texture and luminance recovery. To further boost shadow removal, a rectified outreach attention mechanism has been designed to reduce noise commonly introduced by self-attention. Additionally, the color regeneration network has demonstrated its critical role in accurately correcting and restoring image colors. Extensive experimental results have underscored the superior performance and robustness of our model.